\documentclass[nonacm, sigconf]{acmart}

\definecolor{ginger}{rgb}{0.9, 0.45, 0.0}
\definecolor{darkorange}{rgb}{1.0, 0.55, 0.0}

\acmSubmissionID{384}

\usepackage{booktabs} 

\usepackage[ruled]{algorithm2e} 

\SetAlFnt{\small}
\SetAlCapFnt{\small}
\SetAlCapNameFnt{\small}
\SetAlCapHSkip{0pt}

\usepackage{tikz}
\usetikzlibrary{arrows}

\usepackage{subfig}
\usepackage{comment}
\usepackage{multirow}
\newcommand{\tabincell}[2]{\begin{tabular}{@{}#1@{}}#2\end{tabular}}

\setcopyright{none}
\renewcommand\footnotetextcopyrightpermission[1]{}

\begin{document}
\title{AgileAvatar: Stylized 3D Avatar Creation via Cascaded Domain Bridging}

\settopmatter{authorsperrow=5}

\thanks{Authors’ email addresses: Shen Sang: shen.sang@bytedance.com; Tiancheng Zhi: tiancheng.zhi@bytedance.com; Guoxian Song: guoxiansong@bytedance.com; Minghao Liu: mliu40@ucsc.edu; Chunpong Lai: chunpong.lai@bytedance.com; Jing Liu: jing.liu@bytedance.com; Xiang Wen: sidao@bytedance.com; James Davis: davis@cs.ucsc.edu; Linjie Luo: linjie.luo@bytedance.com}


\author{Shen Sang}
\affiliation{
 \institution{ByteDance}
 \city{Mountain View}
 \country{USA}
}

\author{Tiancheng Zhi}
\affiliation{
 \institution{ByteDance}
 \city{Mountain View}
 \country{USA}
}

\author{Guoxian Song}
\affiliation{
 \institution{ByteDance}
 \city{Mountain View}
 \country{USA}
}

\author{Minghao Liu}
\affiliation{
 \institution{UC Santa Cruz}
 \city{Santa Cruz}
 \country{USA}
}

\author{Chunpong Lai}
\affiliation{
 \institution{ByteDance}
 \city{Mountain View}
 \country{USA}
}

\author{Jing Liu}
\affiliation{
 \institution{ByteDance}
 \city{Mountain View}
 \country{USA}
}

\author{Xiang Wen}
\affiliation{
 \institution{ByteDance}
 \city{Hangzhou}
 \country{China}
}

\author{James Davis}
\affiliation{
 \institution{UC Santa Cruz}
 \city{Santa Cruz}
 \country{USA}
}

\author{Linjie Luo}
\affiliation{
 \institution{ByteDance}
 \city{Mountain View}
 \country{USA}
}

\renewcommand\shortauthors{Sang, S. et al}
\begin{teaserfigure}
\centering
\captionsetup{position=top}
\subfloat[]{
    \includegraphics[width=0.178\linewidth]{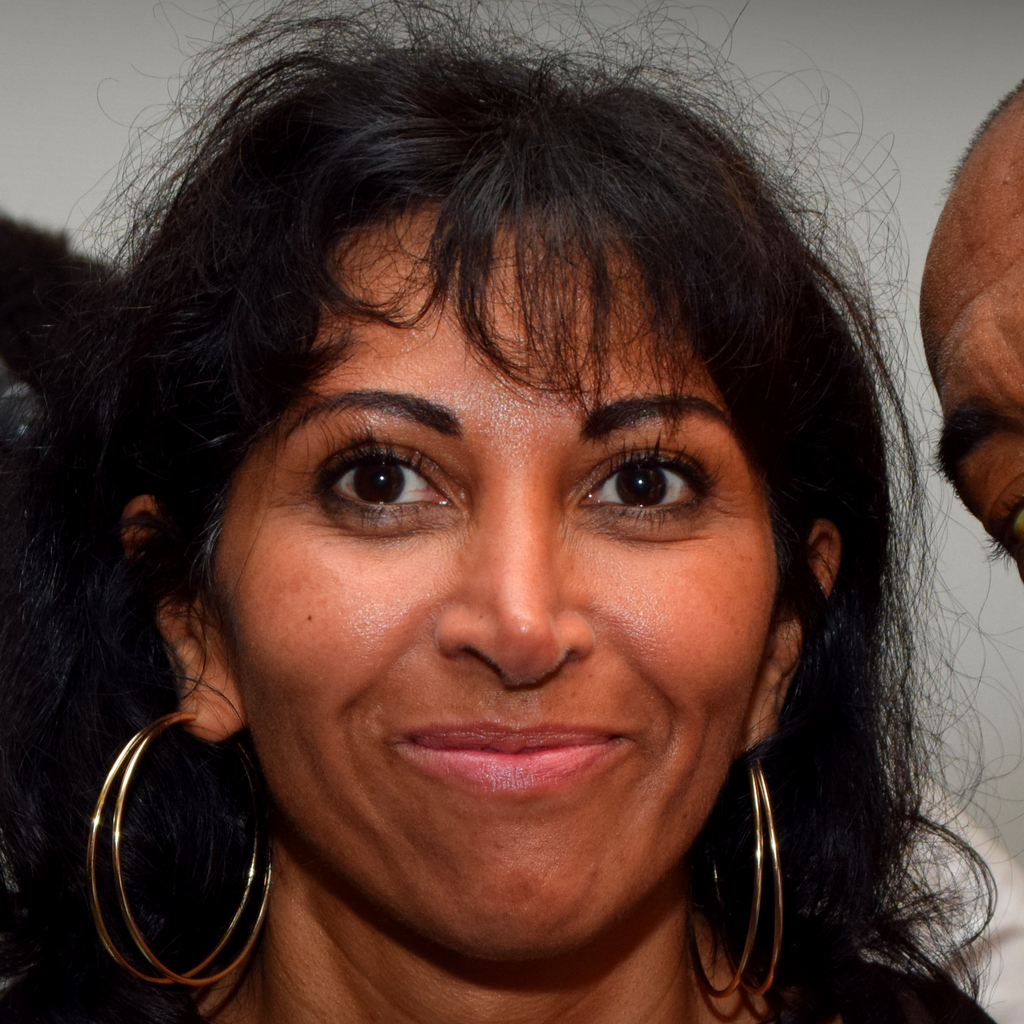}}
\hspace{8pt}
\subfloat[(b.1) Stylization]{
    \includegraphics[width=0.178\linewidth]{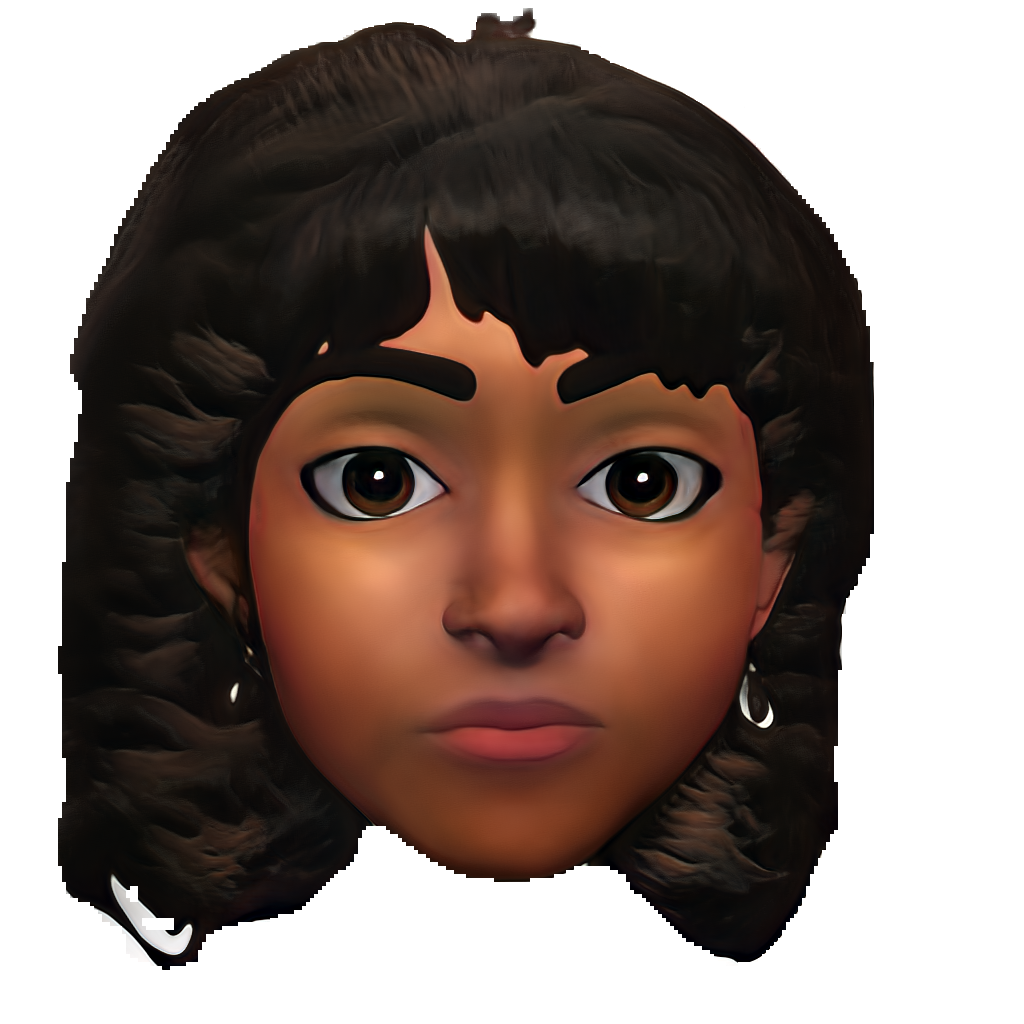}}  
\hspace{-2.5pt}
\subfloat[(b.2) Parameterization]{
    \includegraphics[width=0.178\linewidth]{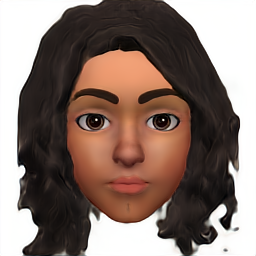}}
    \hspace{1pt}
\subfloat[(b.3) Conversion]{
    \includegraphics[width=0.178\linewidth]{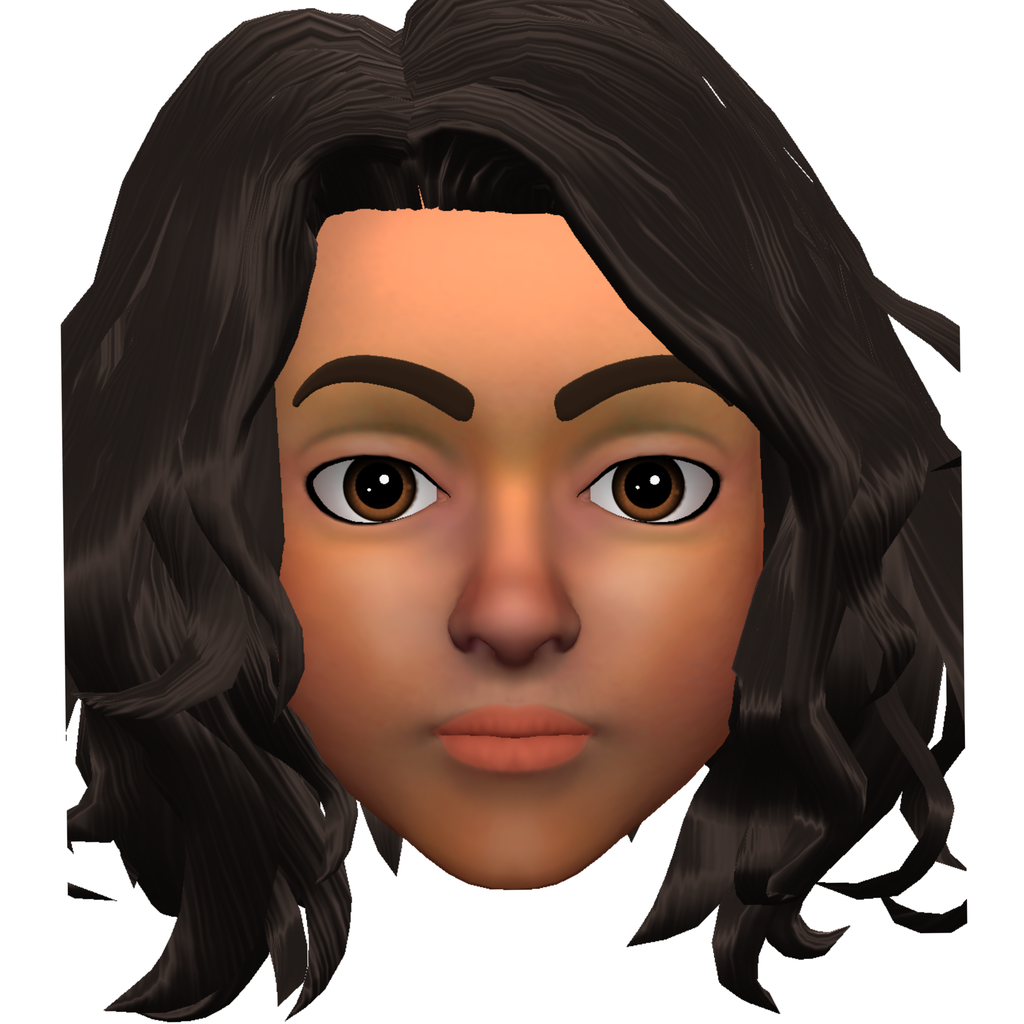}}
    \hspace{1.5pt}
\subfloat[]{
    \includegraphics[width=0.178\linewidth]{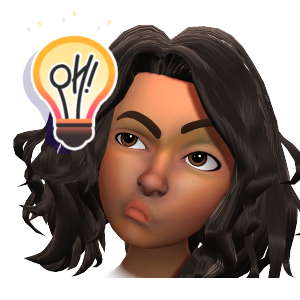}}
\vspace{-20pt}
\captionsetup{position=bottom}
\subfloat[(a) Input]{
    \includegraphics[width=0.178\linewidth]{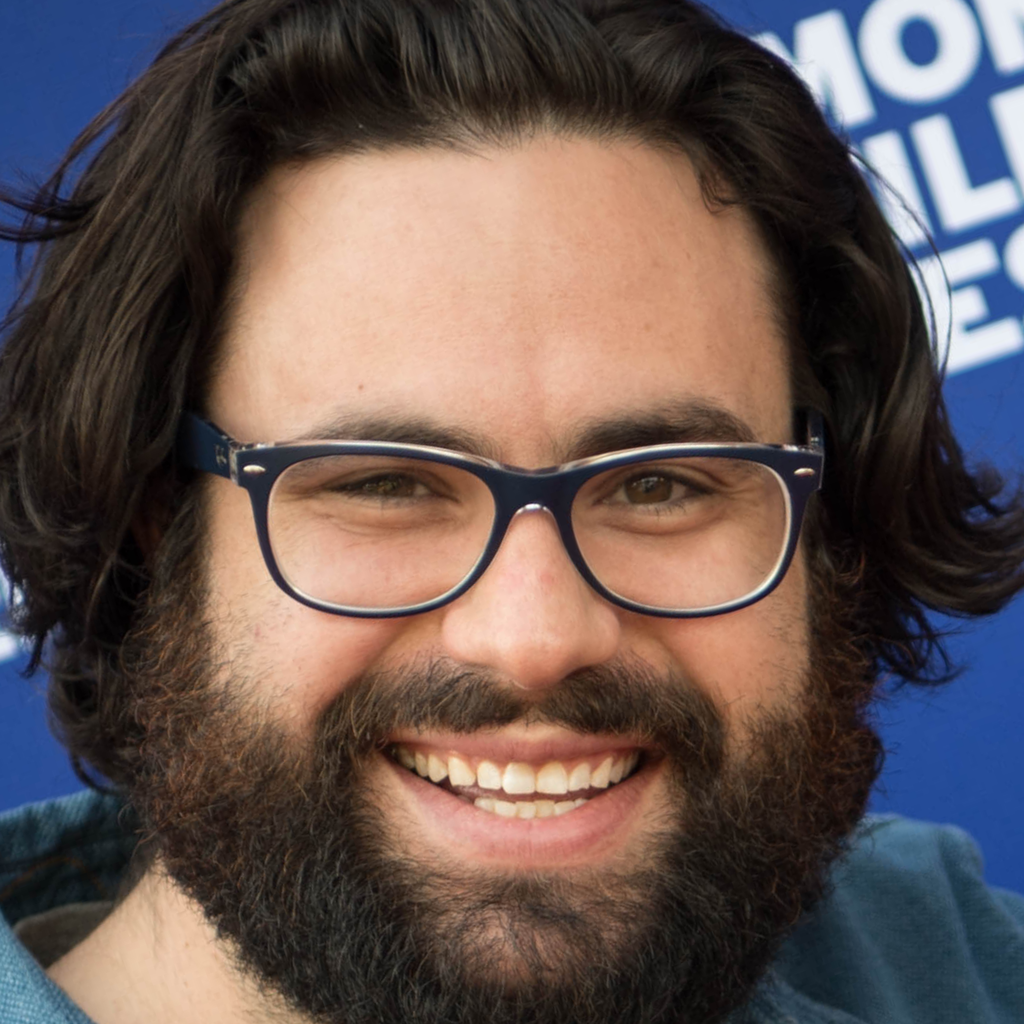}}
\hspace{8pt}
\subfloat[(b) Our method: cascaded domain bridging (from left to right)]{
    \includegraphics[width=0.178\linewidth]{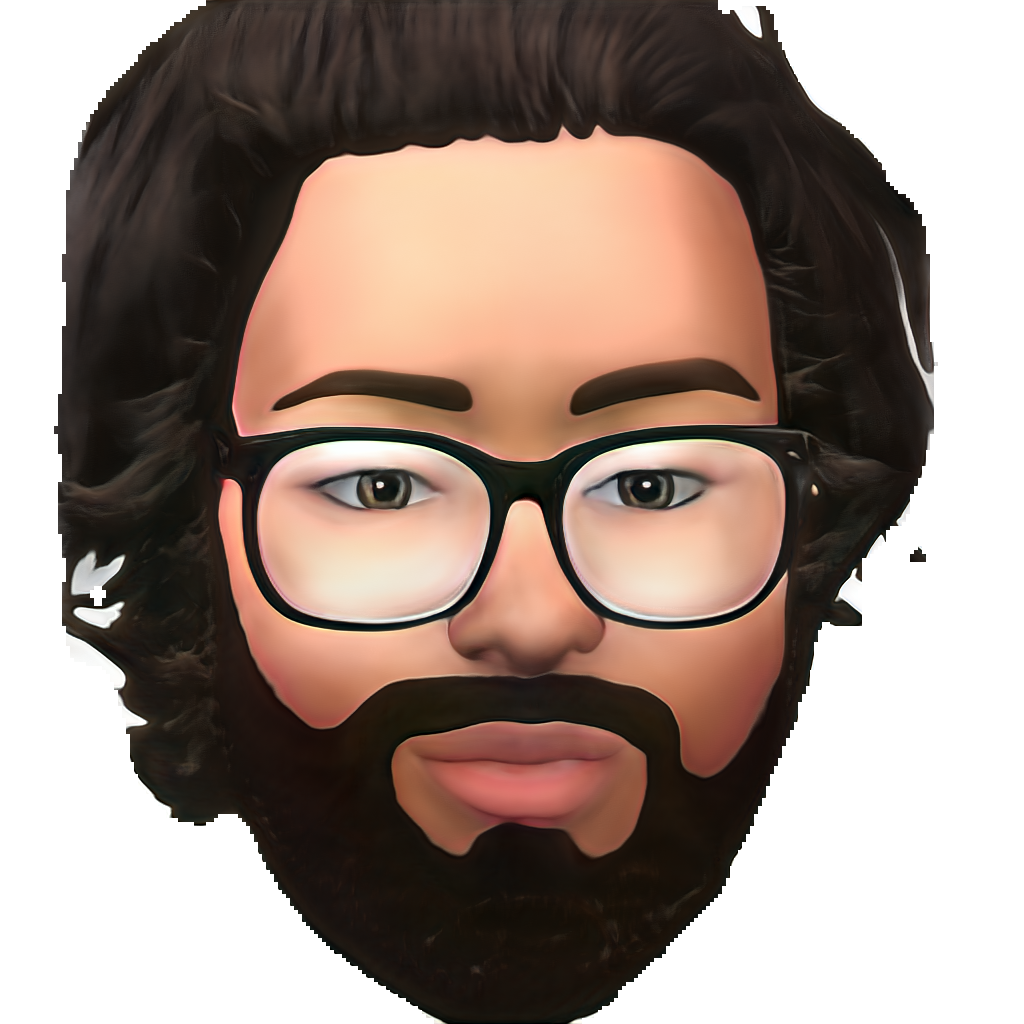}
    \hspace{-2.5pt}
    \includegraphics[width=0.178\linewidth]{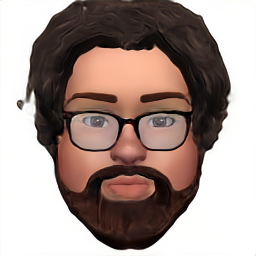}
    \hspace{1pt}
    \includegraphics[width=0.178\linewidth]{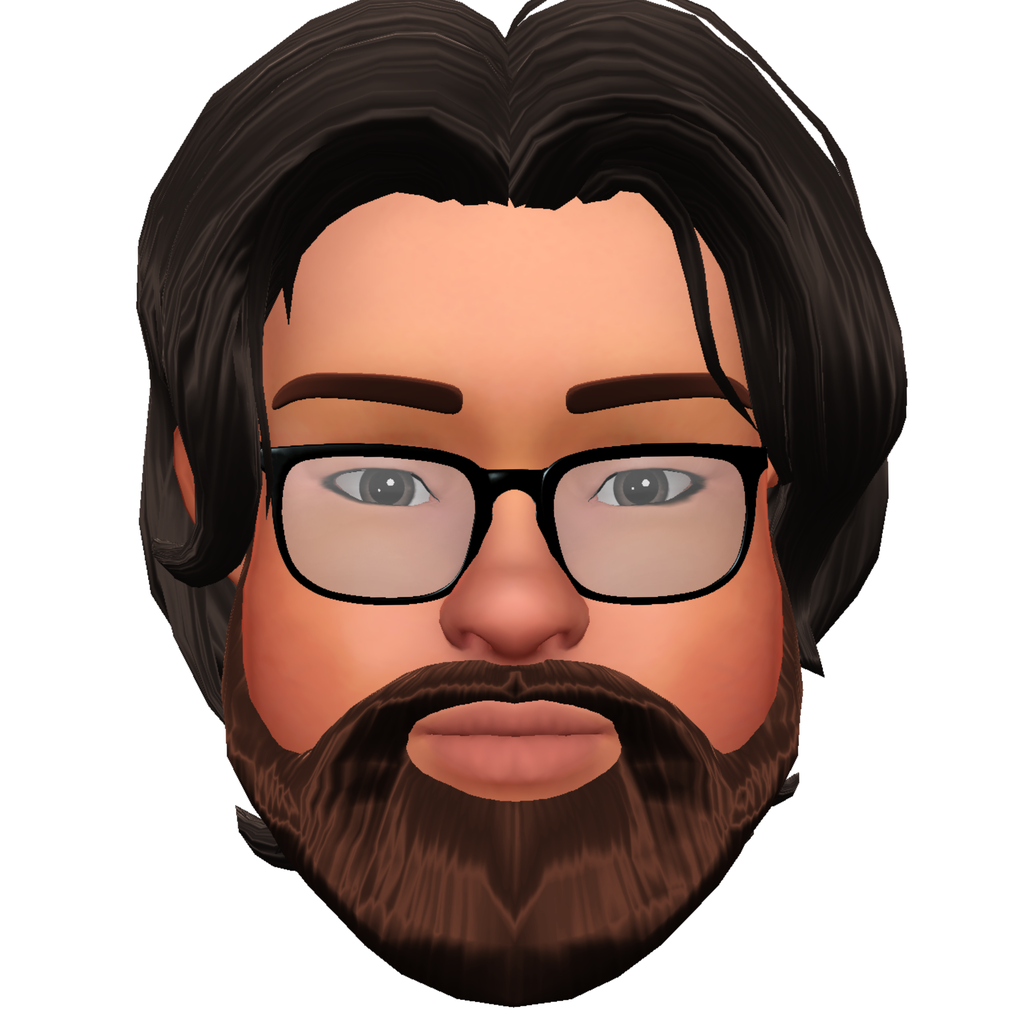}}
    \hspace{1.5pt}
\subfloat[(c) Application]{
    \includegraphics[width=0.178\linewidth]{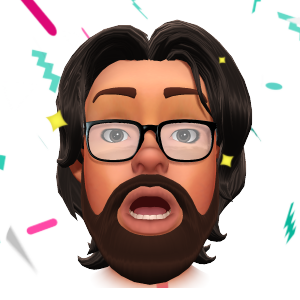}}
  \caption{(a) Given a front-facing user image as input, (b) our method progressively bridges the domain gap between real faces and 3D avatars through three stages: (b.1) The stylization stage performs an image space translation to generate a stylized portrait while normalizing expressions. (b.2) The parameterization stage uses a learned model to find avatar parameters which match the results of stylization. (b.3) The conversion stage searches for a valid avatar vector matching the parameterization that can be rendered by the graphics engine. (c) The output is a user editable 3D model which can be animated and applied to various applications, for example personalized emoji. \textcopyright H JACQUOT and Montclair Film.} 
  \label{fig:teaser}
  \vspace{6pt}
\end{teaserfigure}

\begin{abstract}
Stylized 3D avatars have become increasingly prominent in our modern life. Creating these avatars manually usually involves laborious selection and adjustment of continuous and discrete parameters and is time-consuming for average users. Self-supervised approaches to automatically create 3D avatars from user selfies promise high quality with little annotation cost but fall short in application to stylized avatars due to a large style domain gap. We propose a novel self-supervised learning framework to create high-quality stylized 3D avatars with a mix of continuous and discrete parameters. Our cascaded domain bridging framework first leverages a modified portrait stylization approach to translate input selfies into stylized avatar renderings as the targets for desired 3D avatars. Next, we find the best parameters of the avatars to match the stylized avatar renderings through a differentiable imitator we train to mimic the avatar graphics engine. To ensure we can effectively optimize the discrete parameters, we adopt a cascaded relaxation-and-search pipeline. We use a human preference study to evaluate how well our method preserves user identity compared to previous work as well as manual creation. Our results achieve much higher preference scores than previous work and close to those of manual creation. We also provide an ablation study to justify the design choices in our pipeline.

\end{abstract}

%
%


%
%


\maketitle

\section{Introduction}


An attractive and animatable 3D avatar is an important entry point to the digital world that has become increasingly prominent in modern life for socialization, shopping and gaming etc. A good avatar should be both personalized (reflecting the person's unique appearance) and good-looking. Many popular avatar systems adopt cartoonized and stylized designs for their playfulness and appealingness to the users such as Zepeto\footnote{https://zepeto.me/} and ReadyPlayer\footnote{https://readyplayer.me/}. However, creating an avatar manually usually involves laborious selections and adjustments from a swarm of art assets which is both time-consuming and difficult for average users with no prior experience.

In this paper, we study automatic creation of stylized 3D avatars from a single front-facing selfie image. To be specific, given a selfie image, our algorithm predicts an \emph{avatar vector} as the complete configuration for a \emph{graphics engine} to generate a 3D avatar and render avatar images from predefined 3D assets.
The avatar vector consists of parameters specific to the predefined assets which can be either continuous (e.g. head length) or discrete (e.g. hair types).

A naive solution is to annotate a set of selfie images and train a model to predict the avatar vector via supervised learning. However, large scale annotations are needed to handle a large range of assets (usually in the hundreds). 
To alleviate the annotation cost, self-supervised methods \cite{shi2019face,shi2020fast} are proposed to train a differentiable \emph{imitator} that mimics the renderings of the graphics engine to automatically match the rendered avatar image with the selfie image using various losses of identity and semantic segmentation. While these methods proved effective to create semi-realistic avatars close to user's identity, they fall short in application to stylized avatars since the style domain gap between selfie images and stylized avatars are too large (see Fig.~\ref{fig:compare_methods}).


Our main technical challenges are two folds: (1) the large domain gap between user selfie images and stylized avatars and (2) the complex optimization of a mix of continuous and discrete parameters in the avatar vector.
To address these challenges, we formulate a cascaded framework which progressively bridge the domain gap while ensuring optimization convergence on both continuous and discrete parameters.  Our novel framework consists of three stages: Portrait Stylization, Self-supervised Avatar Parameterization, and Avatar Vector Conversion. Fig.~\ref{fig:teaser} shows the domain gap gradually bridged across the three stages, while the identity information (hair style, skin tone, glasses, etc.) is maintained throughout the pipeline. 

First, the Portrait Stylization stage focuses on 2D real-to-stylized visual appearance domain crossing. This stage translates input selfie image to a stylized avatar rendering and remains in image space. Naively applying existing stylization methods~\cite{song2021agilegan, pinkney2020resolution} for translation will retain factors such as expression, which would unnecessarily complicate later stages of our pipeline. Thus, we create a modified variant from AgileGAN~\cite{song2021agilegan} to ensure uniformity in expression while preserving user identity.

Next, the Self-Supervised Avatar Parameterization stage focuses on crossing from image pixel domain to avatar vector domain.
We observed that strictly enforcing parameter discreteness causes optimization to fail to converge. To address this, we use a \emph{relaxed} formulation called a \emph{relaxed avatar vector} in which discrete parameters are encoded as continuous one-hot vectors. To enable differentiability in training, we trained an imitator in similar spirit to F2P~\cite{shi2019face} to mimic the behavior of the non-differentiable 
engine.

Finally, the Avatar Vector Conversion stage focuses on domain crossing from the relaxed avatar vector space to the \emph{strict avatar vector space} where all the discrete parameters are one-hot vectors. The strict avatar vector can then be used by the graphics engine to create final avatars and for rendering. We employ a novel search process that leads to better results than direct quantization.


To evaluate our results, we use a human preference study to evaluate how well our method preserves personal identity relative to baseline methods including F2P~\shortcite{shi2019face} as well as manual creation. Our results achieve much higher scores than baseline methods and close to those of manual creation. We also provide an ablation study to justify the design choices in our pipeline.

In summary, our technical contributions are:

\begin{itemize}
    \item A novel self-supervised learning framework to create high-quality stylized 3D avatars with a mix of continuous and discrete parameters;
    \item A novel approach to cross the large style domain gap in stylized 3D avatar creation using portrait stylization;
    \item A cascaded relaxation and search pipeline that solves the convergence issue in discrete avatar parameter optimization.
\end{itemize}

\begin{figure*}[]
  \includegraphics[width=0.98\linewidth]{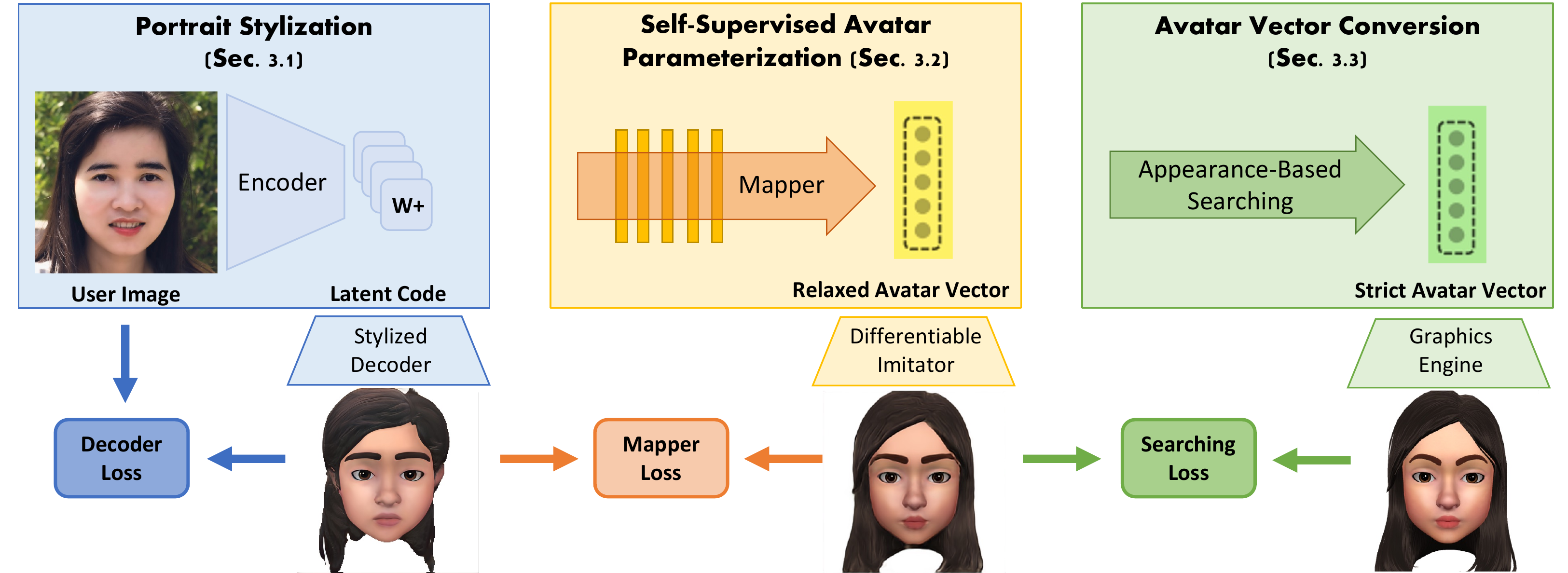}
  \caption{Pipeline. Our framework consists of three modules: Portrait Stylization for image-space real-to-stylized domain crossing, Self-supervised Avatar Parametrization for recovering relaxed avatar vector from the stylization latent code, and Avatar Vector Conversion for discretizing the predicted relaxed avatar vector into a strict avatar vector that can be taken by the graphics engine directly.  \textcopyright NGÁO STUDIO.}
  \label{fig:pipeline}
\end{figure*}
\section{Related Work}

\paragraph{3D Face Reconstruction:}
Photorealistic 3D face reconstruction from images has been studied extensively for many years. Extremely high quality models can be obtained using gantries with multiple cameras followed by a stereo or photogrammetry reconstruction~\cite{beeler2010high, Yang_2020_CVPR}. When only a single image is available, researchers leverage a parameterized 3D morphable model to reconstruct realistic 3D faces~\cite{blanz1999morphable,peng2017parametric, deng2019accurate, xu2020deep, chen2021learning}. Excellent surveys~\cite{egger20203d, zollhofer2018state} exist providing great insights in this direction. These methods focus on an accurate reconstruction of the real human, and the model parameters often lack physical meaning. In contrast our work focuses on cross domain creation of a stylized avatar which has parameters with direct meaning to casual users.





\paragraph{3D Caricature:} Non-photorealistic 3D face reconstruction has also received interest recently, a popular style being caricature. Qiu et al.~\shortcite{qiu20213dcaricshop} created a dataset of 3D caricature models for reconstructing meshes from caricature images. Some works generate caricature meshes by exaggerating or deforming real face meshes, with~\cite{wu2018alive,cai2021landmark} or without~\cite{lewiner2011interactive,vieira2013three} caricature image input. Sketches can be used to guide the creation~\cite{han2017deepsketch2face,han2018caricatureshop}. Recent works~\cite{li2021deep,ye20213d} use GANs to generate 3D caricatures given real images. However, these methods are designed for reconstructing caricature meshes and/or textures while we focus on cartoonish avatars constrained by parameters with semantic meaning.

\paragraph{Game Avatars:}
Commercial products such as Zepeto and ReadyPlayer use a graphics engine to render cartoon avatars from user selfies. While no detailed description of their methods exists, we suspect these commercial methods are supervised with a large amount of manual annotations, something this paper seeks to avoid. 

Creating semi-realistic 3D avatars has also been explored~\cite{hu2017avatar, ichim2015dynamic, luo2021normalized, cao2016real}. Most relevant to our framework, Shi et al. \shortcite{shi2019face} proposed an algorithm to search for the optimal avatar parameters by comparing the input image directly to the rendered avatar. Follow-up work improves efficiency~\cite{shi2020fast}, and seeks to use the photograph's texture to make the avatar match more closely~\cite{lin2021meingame}. These efforts seek to create a similar looking avatar, while this paper seeks to create a highly stylized avatar with a large domain gap.


\paragraph{Portrait Stylization:}
Many methods for non-photorealistic stylization of 2D images exist. Gatys et al.~\shortcite{gatys2016image} proposed neural style transfer, matching features at different levels of CNNs. Image-to-image models focus on the translation of images from a source to target domain, either with paired data supervision~\cite{isola2017image} or without~\cite{zhu2017unpaired, park2020contrastive}. Recent development in GAN inversion~\cite{richardson2021encoding, tov2021designing} and interpolation~\cite{pinkney2020resolution} methods make it possible to achieve high quality cross-domain stylization~\cite{song2021agilegan, cao2018cari, zhu2021mind}. The end result of these methods are in 2D pixels space and directly inspire the first stage of our pipeline.




\section{Proposed Approach}

 Our cascaded avatar creation framework consists of three stages: Portrait Stylization (Sec.~\ref{ssec:agilegan}), Self-supervised Avatar Parameterization (Sec.~\ref{ssec:imitator}), and Avatar Vector Conversion (Sec.~\ref{ssec:post}). A diagram of their relationship is shown in Fig.~\ref{fig:pipeline}.  Portrait Stylization transforms a real user image into a stylized avatar image, keeping as much personal identity (glasses, hairs, colors, etc.) as possible, while simultaneously normalizing the face to look closer to an avatar rendering. Next, the Self-supervised Avatar Parameterization module regresses a relaxed avatar vector from the stylization latent code via a MLP based Mapper. 
 Finally, the Avatar Vector Conversion module discretizes part of the relaxed avatar vector to meet the requirement of the graphics engine using an appearance-based search.

\subsection{Portrait Stylization}
\label{ssec:agilegan}


Portrait Stylization transforms user images into stylized  images close to our target domain. This stage of our pipeline occurs entirely within the 2D image domain. We adopt an encoder-decoder framework for the stylization task. A novel transfer learning approach is applied to a StyleGAN model~\cite{karras2020analyzing}, including W+ space transfer learning, using a normalized style exemplar set, and a loss function that supports these modifications.

\begin{figure}

\subfloat[]{
\includegraphics[width=0.25\linewidth]{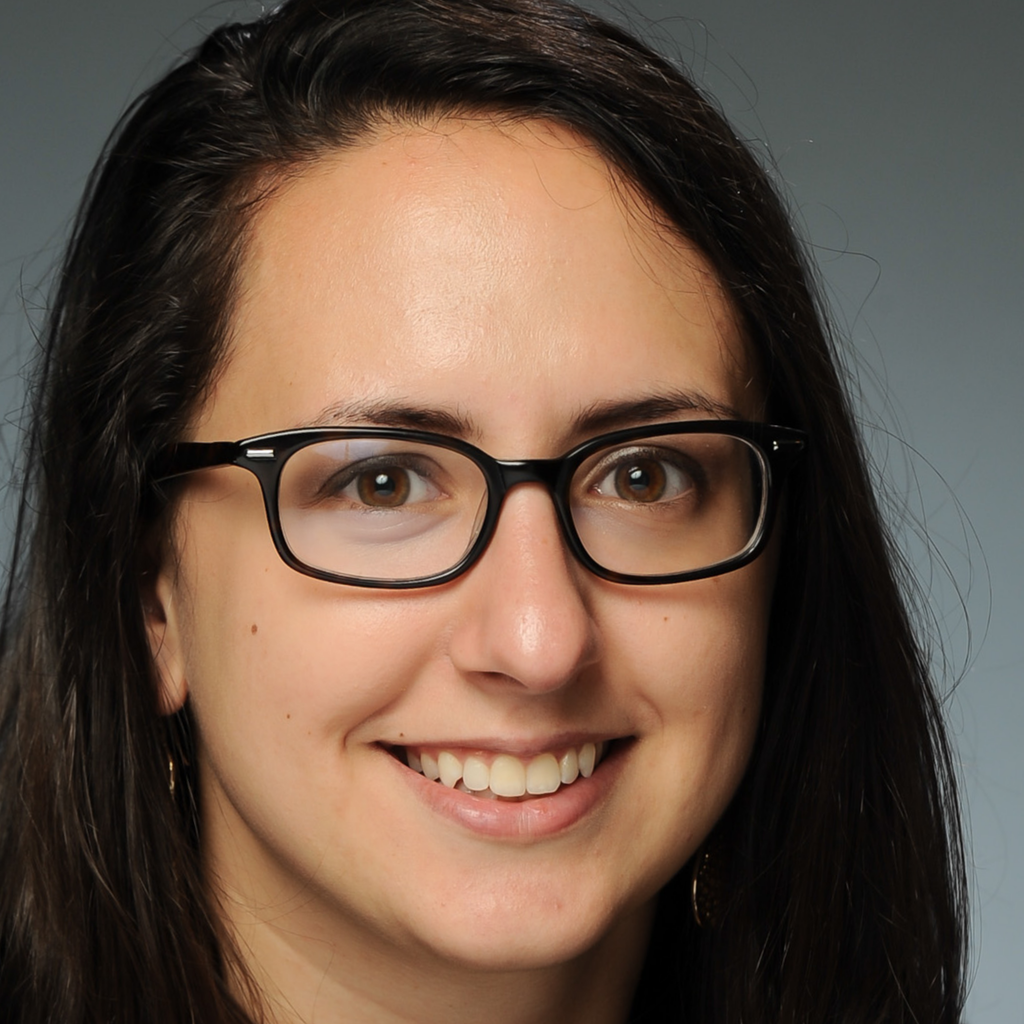}}
\hspace{0.05\linewidth}
\subfloat[]{
\includegraphics[width=0.25\linewidth]{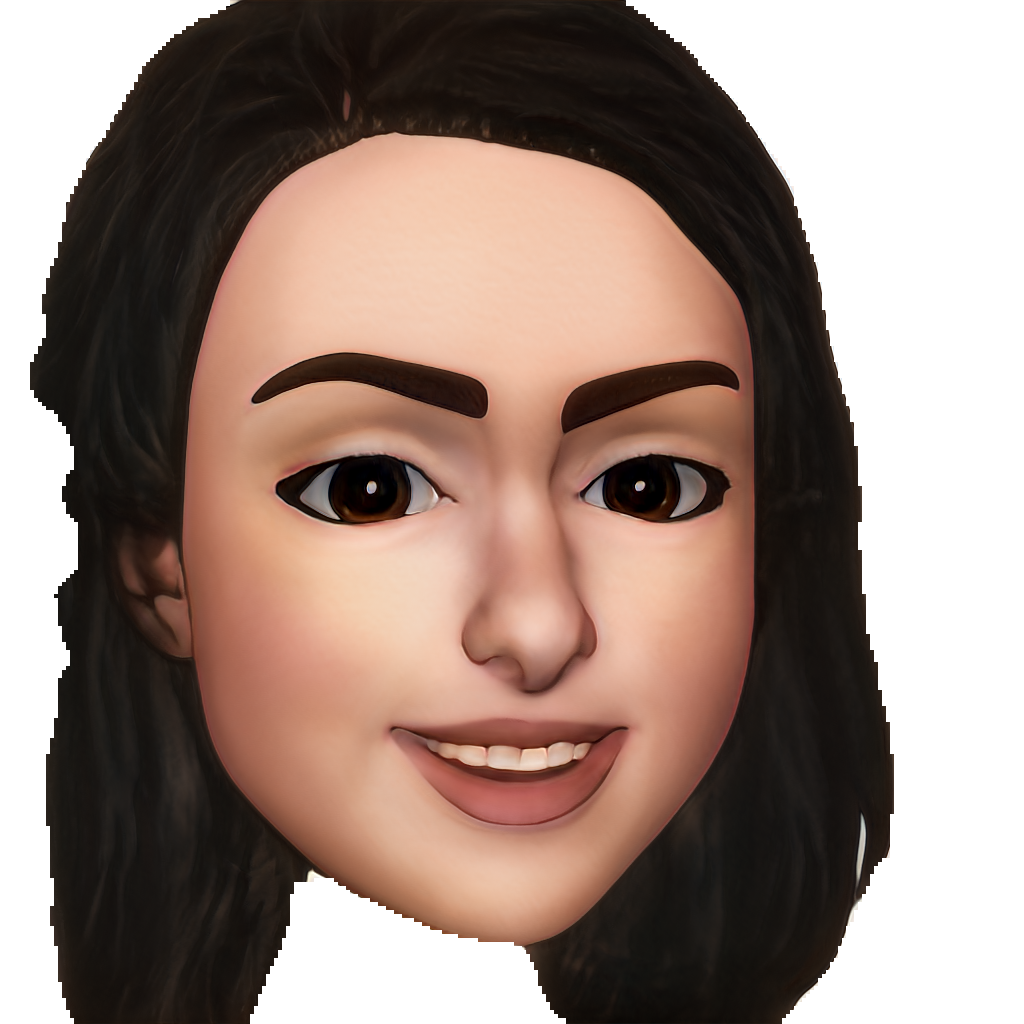}}
\hspace{0.05\linewidth}
\subfloat[]{
\includegraphics[width=0.25\linewidth]{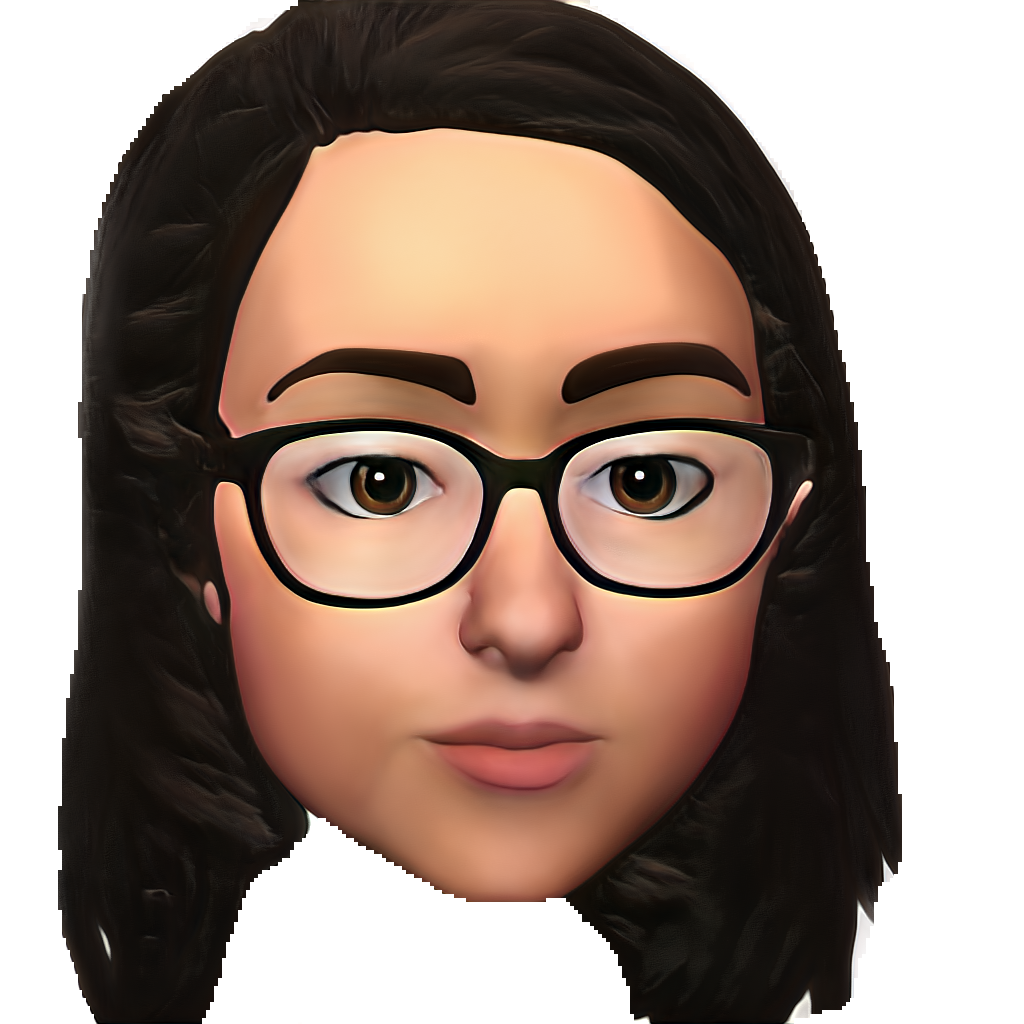}}
\vspace{-22pt}

\subfloat[(a) Input]{
\includegraphics[width=0.25\linewidth]{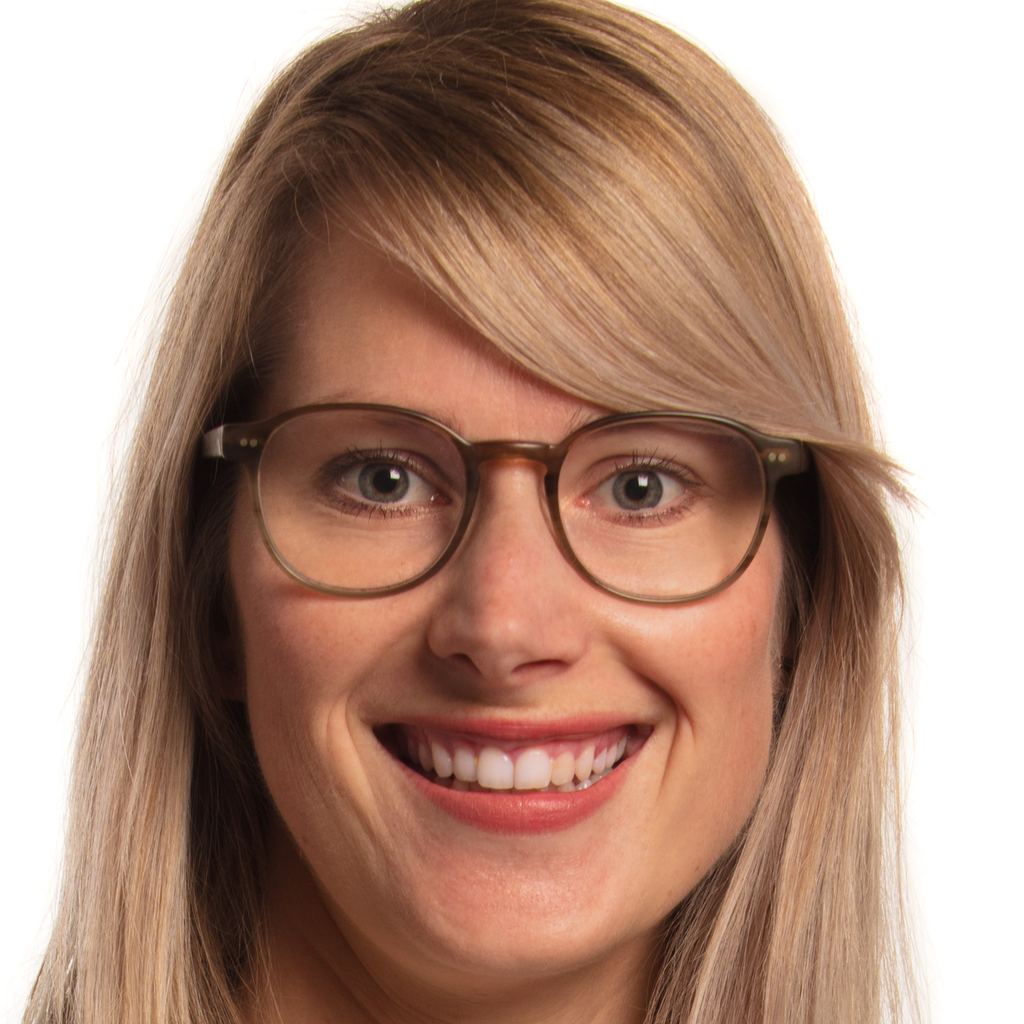}}
\hspace{0.05\linewidth}
\subfloat[(b) AgileGAN]{
\includegraphics[width=0.25\linewidth]{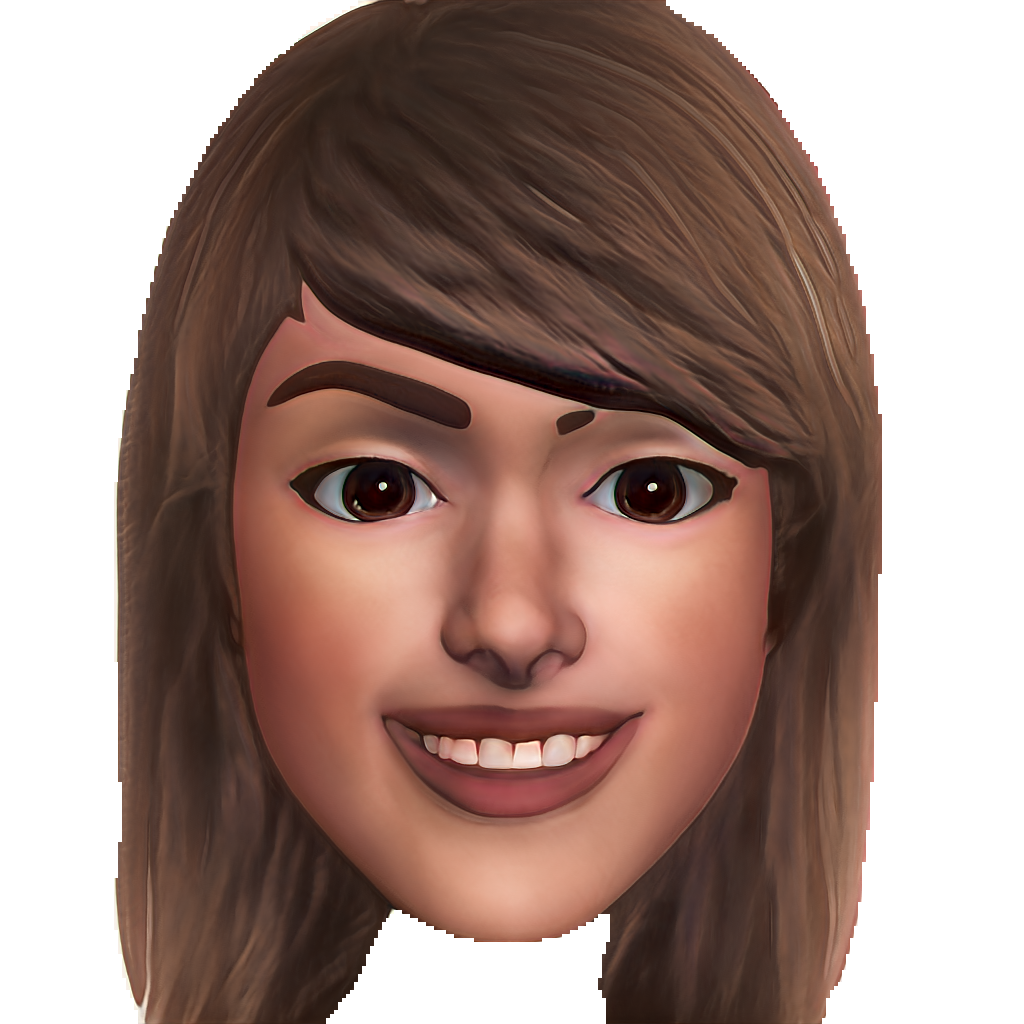}}
\hspace{0.05\linewidth}
\subfloat[(c) Our Stylization]{
\includegraphics[width=0.25\linewidth]{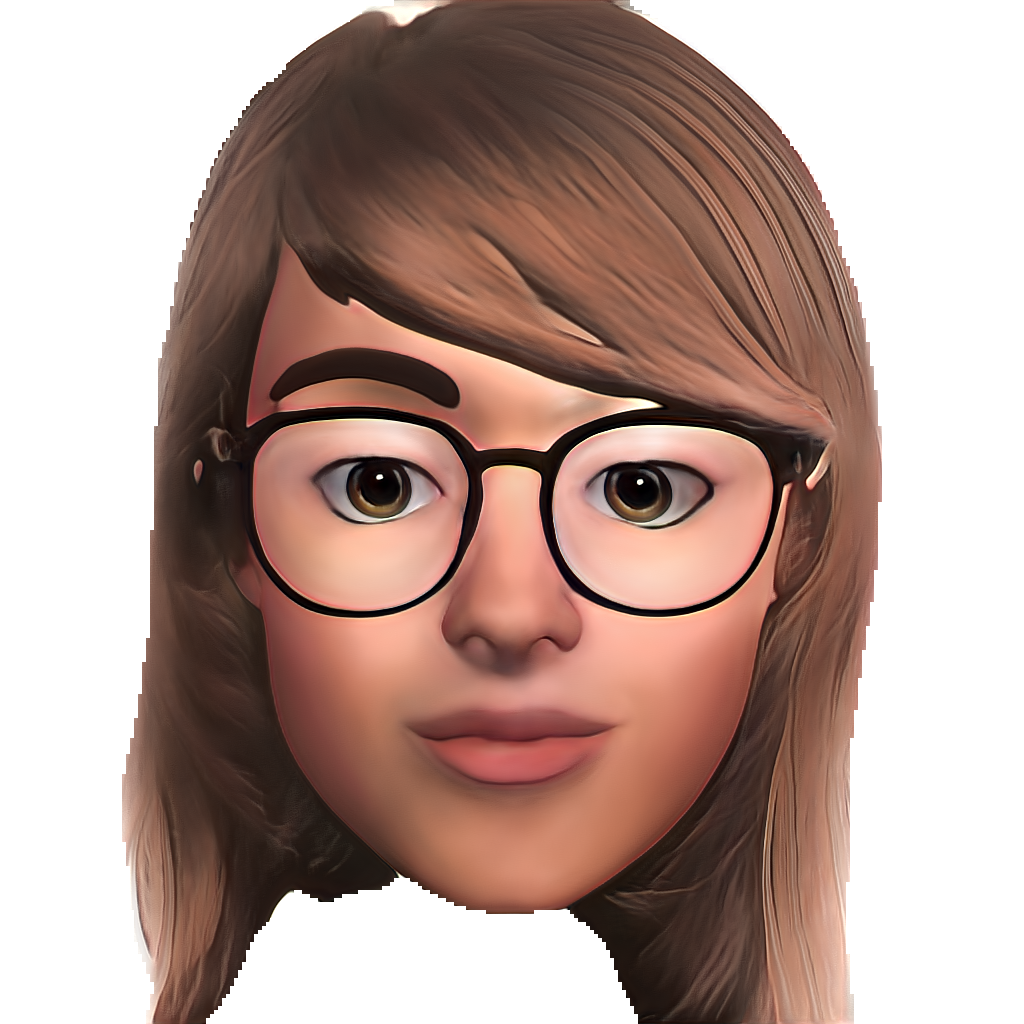}}
  \caption{Portrait stylization results. Compared with a state-of-the-art stylization method, AgileGAN~\cite{song2021agilegan}, our stylization does a better job at preserving the user's personal identity (e.g. glasses are preserved), and simultaneously normalizing the expressions (e.g. mouth is closed) for easier fitting in the downstream pipeline. \textcopyright Greg Mooney and Sebastiaan ter Burg.}
  \label{fig:agilegan++}
\end{figure}

\paragraph{W+ space transfer learning:} We perform transfer learning directly from the W+ space, unlike previous methods ~\cite{song2021agilegan,gal2021stylegannada} where stylization transfer learning is done in the more entangled Z/Z+ space. The W+ space is more disentangled and can preserve more personal identity features. However, this design change introduces a challenge. We need to model a distribution prior $\mathcal{W}$ of the W+ space, as it is a highly irregular space~\cite{Wulff}, and cannot be directly sampled like the Z/Z+ space (standard Gaussian distribution). We achieve this by inverting a large dataset of real face images into a W+ embeddings via a pre-trained image encoder ~\cite{tov2021designing}, and then sample the latent codes from that prior.  Fig.~\ref{fig:agilegan++} provides one example of better preserved personalization. Notice that our method preserves glasses which are lost in the comparison method.

\paragraph{Normalized Style Exemplar Set:} Our stylization method seeks to ignore pose and expression and produce a normalized image. In contrast, existing methods are optimized to preserve source to target similarities literally, transferring specific facial expressions, head poses, and lighting conditions directly from user photos into target stylized images. This is not desirable for our later avatar parameterization stage as we are trying to extract the core personal identity features only. In order to produce normalized stylizations we limit the rendered exemplars provided during transfer learning to contain only neutral poses, expressions and illumination to ensure a good normalization. Fig.~\ref{fig:agilegan++} provides an example of a smiling face. The comparison method preserves the smile, while our method successfully provides only the normalized core identity.


\paragraph{Loss:} 
Our loss contains non-standard terms to support the needs of our pipeline. The target output stylization is not exactly aligned with the input due to pose normalization. Therefore, commonly used perceptual loss~\cite{zhang2018perceptual} cannot be applied directly in decoder training. We instead use a novel segmented color loss.

The full objective comprises three loss terms to fine-tune the generator $\mathcal{G}_\phi$. Let $\mathcal{G}_{\phi_o}$ and $\mathcal{G}_{\phi_t}$ be the model before and after fine-tuning.
We introduce a color matching loss at a semantic level.
Specifically, we leverage two face segmentation models from BiSeNet~\cite{yu2018bisenet} pre-trained on real and stylized data separately to match the color of semantic regions.
Let $\mathbb{S}=\{hair, skin\}$ be the classes taken into consideration, and $\mathcal{B}^k(I)$ ($k\in\mathbb{S}$) be the mean color of pixels belonging to class $k$ in image $I$. $\mathcal{B}^k_{real}$ and $\mathcal{B}^k_{style}$ represent real and stylized models separately. The semantic color matching loss is:

\begin{equation}\label{eq:loss_decoder_per}
\mathcal{L}_{sem} = \mathbb{E}_{w \sim \mathcal{W}}[ \sum_{k \in \mathbb{S}}\left( \left \| \mathcal{B}^k_{real}(\mathcal{G}_{\phi_o}(w) ) -  \mathcal{B}^k_{style}(\mathcal{G}_{\phi_t}(w) )\right \|^2  \right)]
\end{equation}

An adversarial loss is used to match the distribution of the translated images to the target stylized set distribution $\mathcal{Y}$, where $D$ is the StyleGAN2 discriminator~\cite{karras2020analyzing}: 
\begin{equation}\label{eq:loss_decoder_adv}
\mathcal{L}_{adv} = \mathbb{E}_{y\sim \mathcal{Y}}[\min(0, -1+D(y))] + \mathbb{E}_{w \sim \mathcal{W}} [\min(0, -1-D(\mathcal{G}_{\phi_t}(w) ))]
\end{equation}

Also, to improve training stability and prevent artifacts, we use R1 regularization~\cite{mescheder2018training} for the discriminator: $\mathcal{L}_{R1} =  \frac{\gamma }{2}\mathbb{E}_{y\sim \mathcal{Y}}[\left \|\nabla D(y)\right \|^2]$, where we set $\gamma=10$ empirically. 

Finally, the generator and discriminators are jointly trained to optimize the combined objective ${\mathrm{min}}_{\phi}  {\mathrm{max}}_{D}   \mathcal{L}_{stylize}$, where 
\begin{equation}\label{eq:loss_decoder_all}
\mathcal{L}_{stylize}=\lambda_{adv}\mathcal{L}_{adv}+  \lambda_{sem}  \mathcal{L}_{sem} + \lambda_{R1}  \mathcal{L}_{R1}
\end{equation}
$\lambda_{adv}=1,\lambda_{sem}=12,\lambda_{R1}=5$ are constant weights set empirically. Please see the appendix \ref{appendix:agilegan} for more details.



\subsection{Self-supervised Avatar Parameterization}
\label{ssec:imitator}

\begin{figure}

\subfloat[]{
\includegraphics[width=0.25\linewidth]{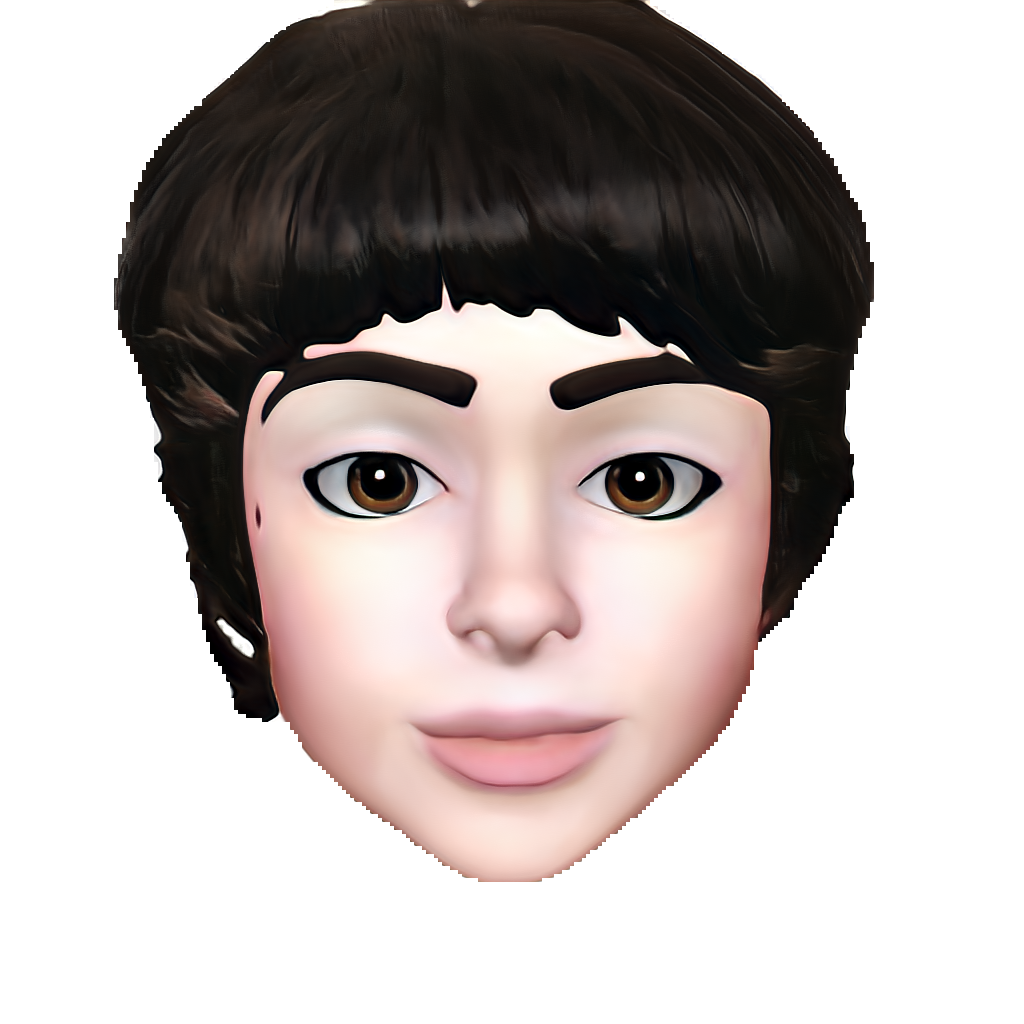}}
\hspace{0.05\linewidth}
\subfloat[]{
\includegraphics[width=0.25\linewidth]{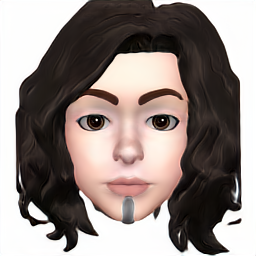}}
\hspace{0.05\linewidth}
\subfloat[]{
\includegraphics[width=0.25\linewidth]{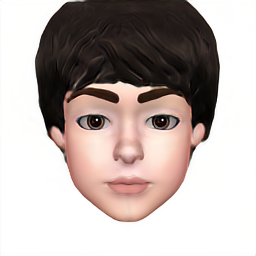}}
\vspace{-22pt}

\subfloat[(a) Stylization]{
\includegraphics[width=0.25\linewidth]{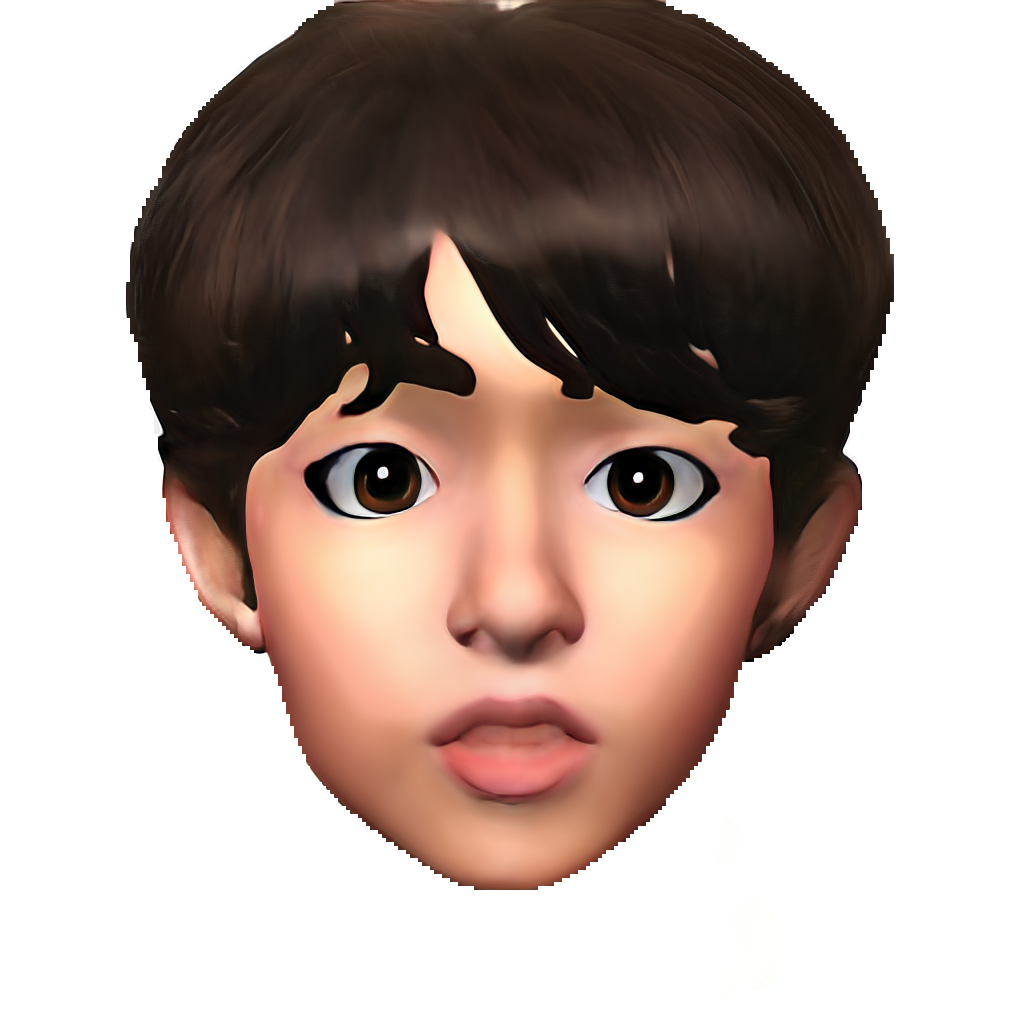}}
\hspace{0.05\linewidth}
\subfloat[(b) Strict]{
\includegraphics[width=0.25\linewidth]{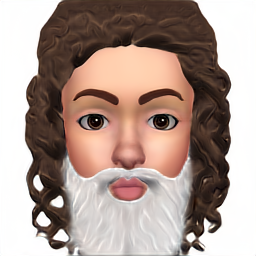}}
\hspace{0.05\linewidth}
\subfloat[(c) Relaxed]{
\includegraphics[width=0.25\linewidth]{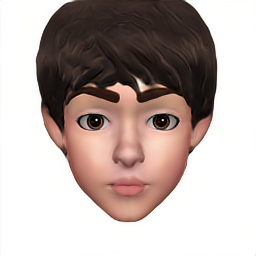}}
\caption{Avatar Parameterization produces errors in final predictions if discrete types are enforced during training, such as hair and beard types in this example. Relaxing the discrete constraint allows easier optimization and thus better predictions which match the stylization target more closely.}
\label{fig:quantized_training}

\end{figure}


Avatar Parameterization finds a set of parameters for the rendering engine which produces an avatar matching the stylized portrait as closely as possible. We call the module which finds parameters the \textit{mapper}. To facilitate training the mapper, we use a differentiable neural rendering engine we call the \textit{imitator}.

A particular avatar is defined by an avatar vector with both continuous and discrete parameters. Continuous parameters are used to control primarily placement and size, for example eye size, eye rotation, mouth position, and head width. Discrete parameters are used to set individual assets and textures such as hair types, beard types, and skin tone textures. All parameters are concatenated into a vector with discrete parameters represented as one-hot vectors. 

\paragraph{Mapper Training:}
The Mapper takes the results of portrait stylization as input and outputs an avatar vector which defines a similar looking avatar.  Rather than using the stylized image itself as input, we use the latent code $w+$ derived from the stylization encoder, since it is a more compact representation and contains facial semantic styles from coarse to fine \cite{karras2019style}.

The Mapper is built as an MLP, and trained using a Mapper Loss which measures the similarity between the stylized image, $\mathcal{I}_{style}$,  and the imitator output, $\mathcal{I}_{imitate}$. This loss function contains several terms to measure the global and local similarity. 

To preserve global appearance, we incorporate identity loss $\mathcal{L}_{id}$ measuring the cosine similarity between two faces built upon a pretrained ArcFace \cite{Deng_2019_CVPR} face recognition network $\mathcal{R}$:
$\mathcal{L}_{id} = 1 - cos(\mathcal{R}(I_{style}), \mathcal{R}(I_{imitate}))$.
For a more fine-grained similarity measurement, LPIPS loss~\cite{zhang2018perceptual} is adopted: $\mathcal{L}_{lpips} = \| \mathcal{F}(I_{style}) - \mathcal{F}(I_{imitate}) \|_{2}$, where $\mathcal{F}$ denotes the perceptual feature extractor.
Additionally, we use a color matching loss to obtain more faithful colors for the skin and hair region:
\begin{equation}\label{eq:loss_decoder_per}
\mathcal{L}_{color} = \sum_{k \in \mathbb{S}}\left( \left \| \mathcal{B}^k_{style}(I_{style}) -  \mathcal{B}^k_{style}(I_{imitate})\right \|^2  \right)
\end{equation}

The final loss function is:
\begin{equation}\label{eq:loss_mapper}
\begin{split}
\mathcal{L}_{mapper} = \lambda_{id} \mathcal{L}_{id} + \lambda_{lpips} \mathcal{L}_{lpips}  + \lambda_{color} \mathcal{L}_{color}
\end{split}
\end{equation}
where $\lambda_{id}=0.4$, $\lambda_{lpips}=0.8$, $\lambda_{color}=0.8$ are set empirically. 

 We empirically choose the best loss terms to provide good results. An ablation study of these terms is provided in the results section.

\paragraph{Differentiable Imitator:}
 The imitator is a neural renderer trained to replicate the output of the graphics engine as closely as possible given an input avatar vector.  The imitator has the important property of differentiablity, making it suitable for inclusion in an optimization framework. We leverage an existing neural model \cite{karras2019style} as the backbone generator, which is capable of generating high quality avatar renderings. We train it with synthetic avatar data supervisedly. See the appendix \ref{appendix:imitator} for details.

\begin{figure}

\centering

\subfloat[]{
\includegraphics[width=0.25\linewidth]{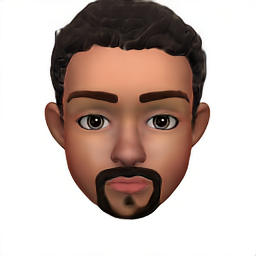}}
\hspace{0.05\linewidth}
\subfloat[]{
\includegraphics[width=0.25\linewidth]{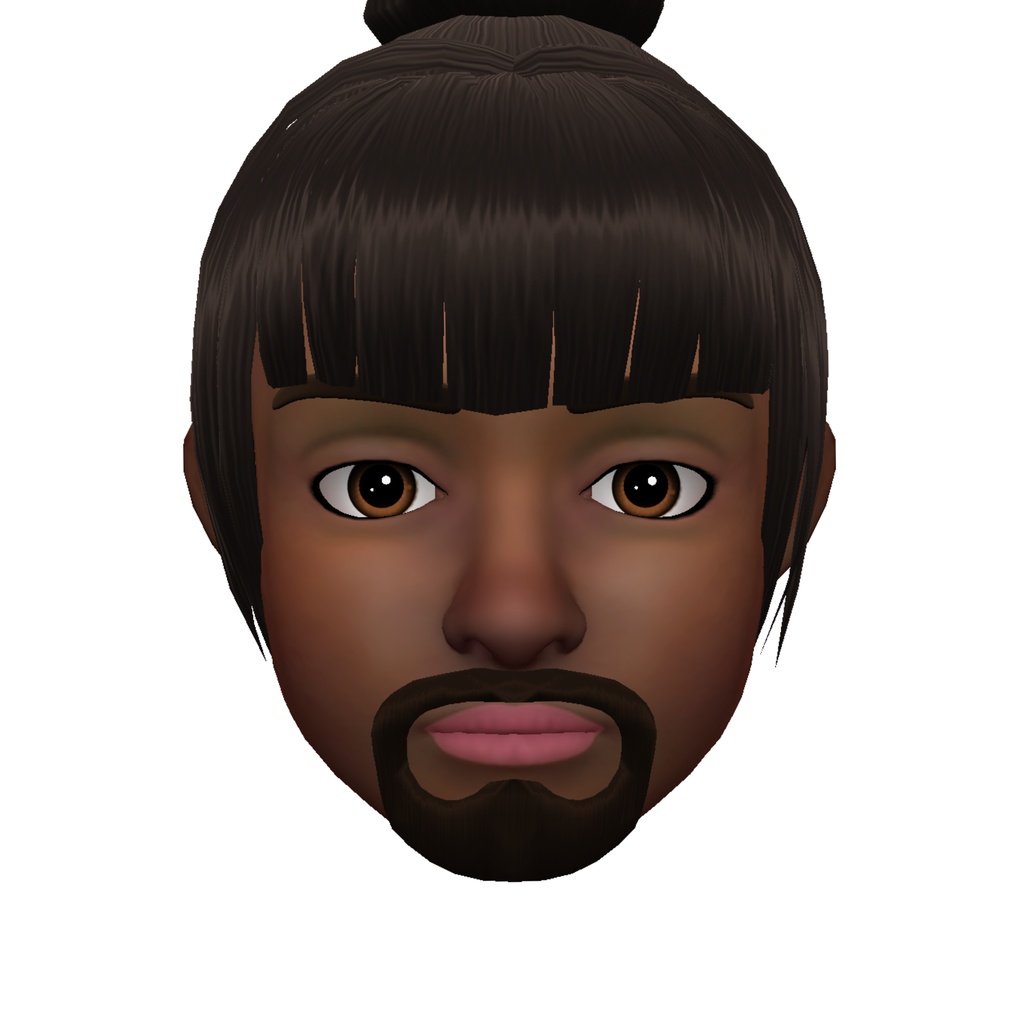}}
\hspace{0.05\linewidth}
\subfloat[]{
\includegraphics[width=0.25\linewidth]{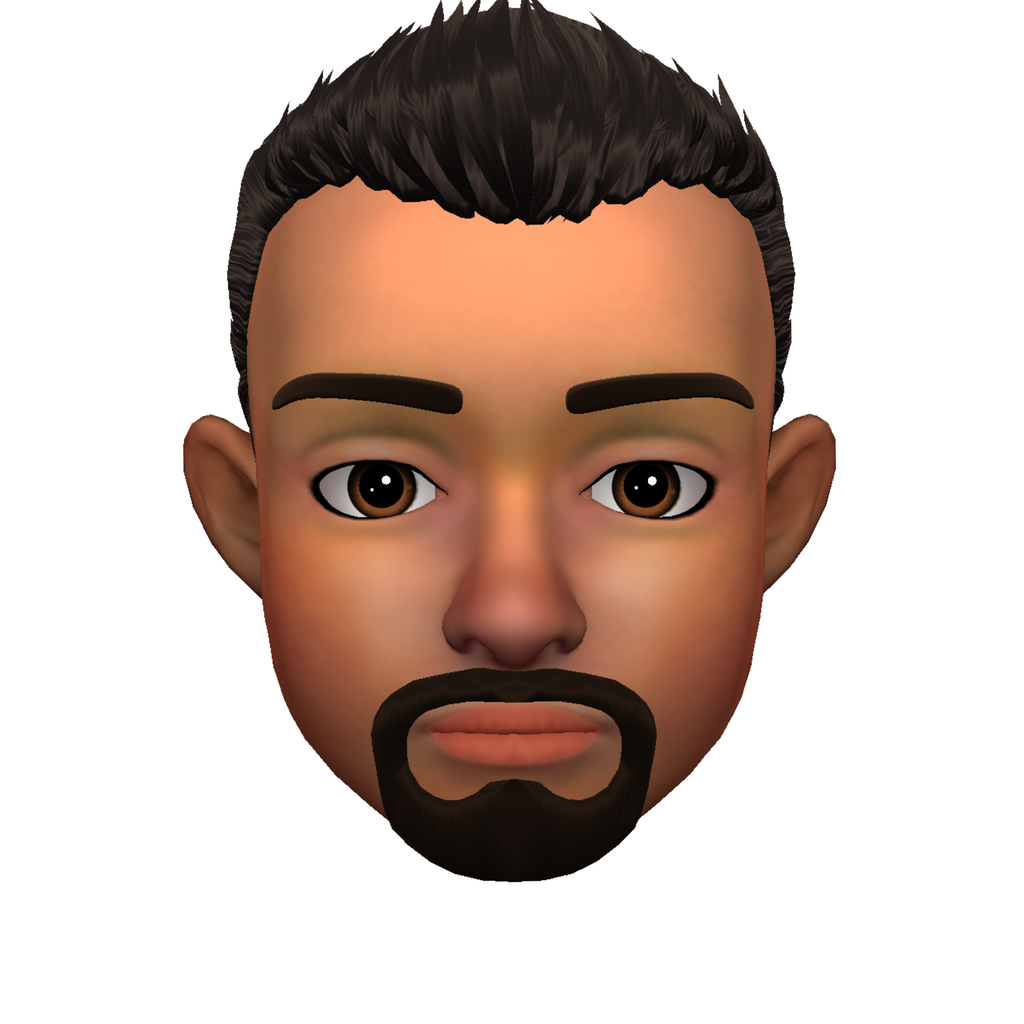}}
\vspace{-28pt}
\subfloat[(a) Relaxed Result]{
\includegraphics[width=0.25\linewidth]{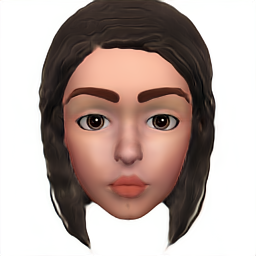}}
\hspace{0.05\linewidth}
\subfloat[(b) Directly Classify]{
\includegraphics[width=0.25\linewidth]{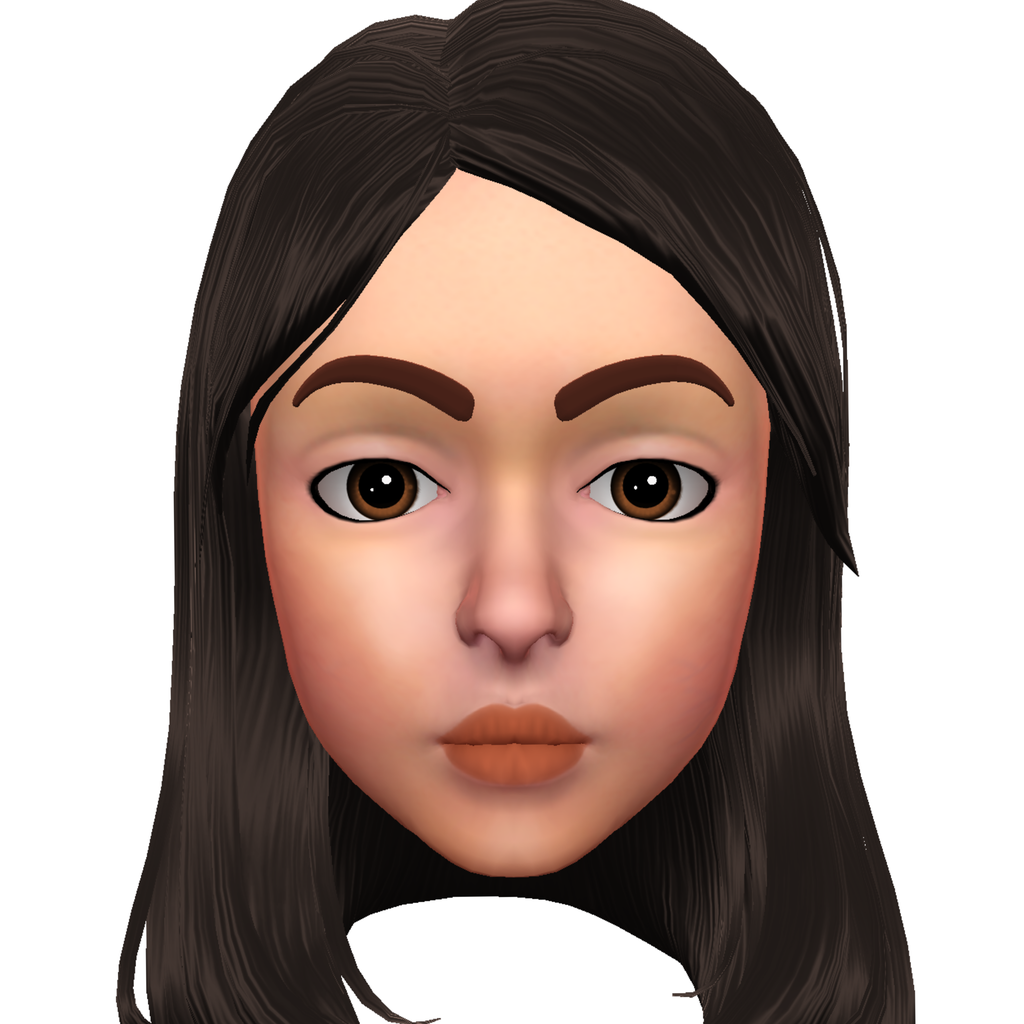}}
\hspace{0.05\linewidth}
\subfloat[(c) Our Conversion]{
\includegraphics[width=0.25\linewidth]{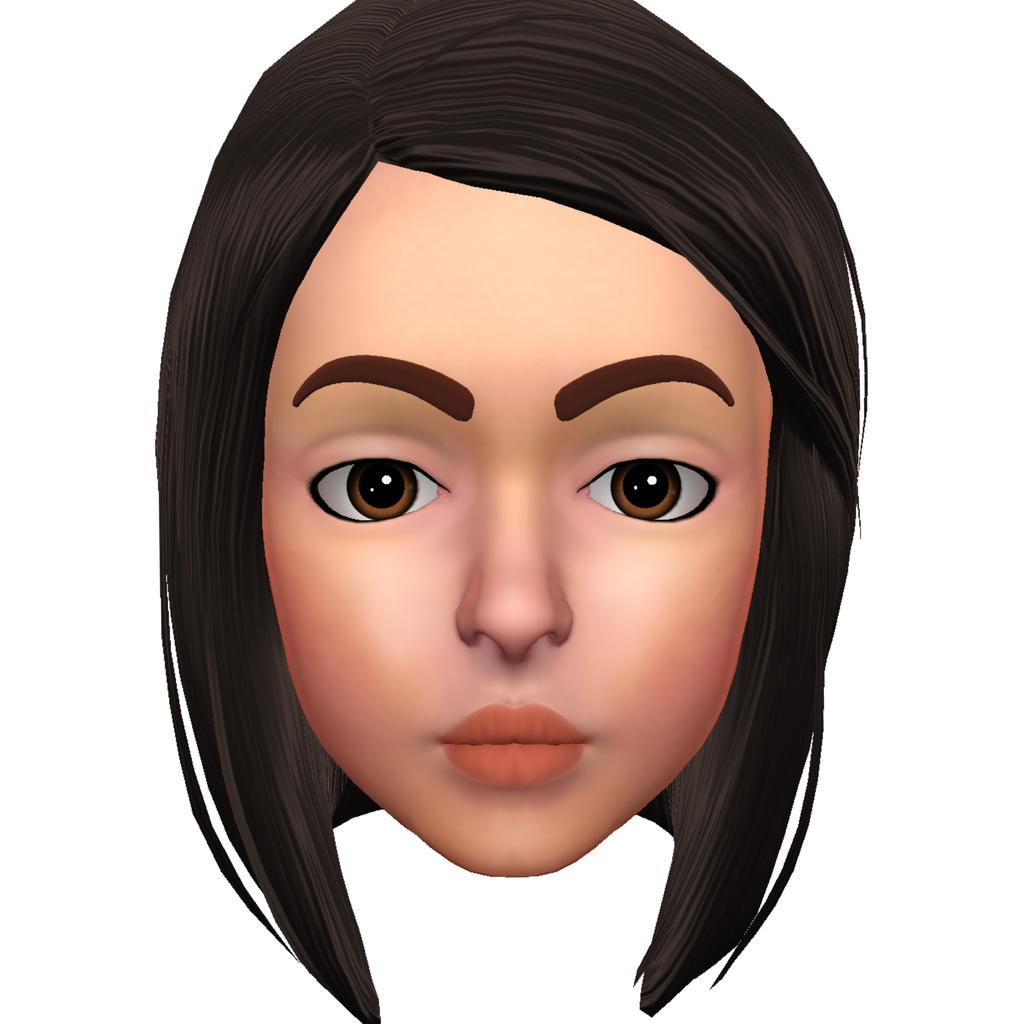}}
\vspace{-4pt}
\caption{Avatar Vector Conversion is necessary to convert the relaxed result produced during parameterization into discrete types suitable for the 
graphics engine. Direct classification often fails to select the best type. Our conversion selects the best match to the relaxed type by searching through all available discrete types. Notice in this example that the skin tone and hair type are much closer using our method.}
\label{fig:post_fitting}
\end{figure}

\paragraph{Discrete Parameters:}
\label{sssec:discrete_params}
Solving for discrete parameters is challenging because of unstable convergence. Some methods handle this via quantization during optimization~\cite{bengio2013estimating,cheng2019straight,van2017neural,jang2016categorical}. However, we found that quantization after optimization, which relaxes the discrete constraint during training and re-apply it as postprocessing, is more effective for our task. Below we describe the relaxed optimization and in Sec.~\ref{ssec:post} we present the quantization method.


Our solution to training discrete parameters in the mapper makes use of the imitator's interpolation property. 
 When mixing two avatar vectors, the imitator still produces a valid rendering. That is, given the one-hot encoding $\mathbf{v}_{1}$ and $\mathbf{v}_{2}$ of two hair or beard types, their linear interpolation $\mathbf{v}_{mix} = (1 - \alpha) \cdot \mathbf{v}_{1} + \alpha \cdot \mathbf{v}_{2}$ ($\alpha\in[0,1]$) produces a valid result. Please see the appendix \ref{appendix:imitator} for details.

Thus, when training the mapper we do \textit{not} strictly enforce discrete parameters, and instead apply a softmax function to the final activation of the mapper to allow a continuous optimization space while still discouraging mixtures of too many asset types.

We compare our relaxed training with a strict training method performing quantization during optimization. In the forward pass, it quantizes the softmax result by picking the entry with maximum probability. In the backward pass, it back-propagates unaltered gradients in a straight-through way~\cite{bengio2013estimating}.
In Fig.~\ref{fig:quantized_training}, our method produces a much closer match to the stylization results.




\subsection{Avatar Vector Conversion}
\label{ssec:post}

The graphics engine requires discrete inputs for attributes such as hair and glasses. However the mapper module in Avatar Parameterization produces continuous values. One straightforward approach for discretization is to pick the type with the highest probability given the softmax result. However, we observe that this approach does not achieve good results, especially when dealing with multi-class attributes (e.g. 45 hair types). The challenge is that the solution space is under-constrained. Medium length hair can be achieved by selecting the medium length hair type, or by mixing between short and long hair types. In the latter case, simply selecting the highest probability of short or long hair is clearly not optimal.


We discretize the relaxed avatar vector via searching over all candidates from the asset list for each attribute, while fixing all other parameters. 
Using the image result from the imitator $\mathcal{I}_{imitate}$ as target, we use the loss function from Eq.~\ref{eq:loss_mapper} as an objective to measure the similarity between $\mathcal{I}_{imitate}$ and the candidate result $\mathcal{I}_{cand}$. By minimizing the objective, we can find the best solution for each attribute. The selections for each attribute are combined to create the avatar vector used for graphics rendering and animation. Fig.~\ref{fig:post_fitting} provides a comparison of direct classification and our method. Note that direct classification makes incorrect choices for hair type and skin color while ours closely matches the reference image. 
 

\begin{figure}[t]
\hspace{-5pt} FID = 236.8 \hspace{14pt}  FID = 38.7  \hspace{18pt}   FID = 17.9  \hspace{20pt}   In domain
\subfloat[]{
\includegraphics[width=0.24\linewidth]{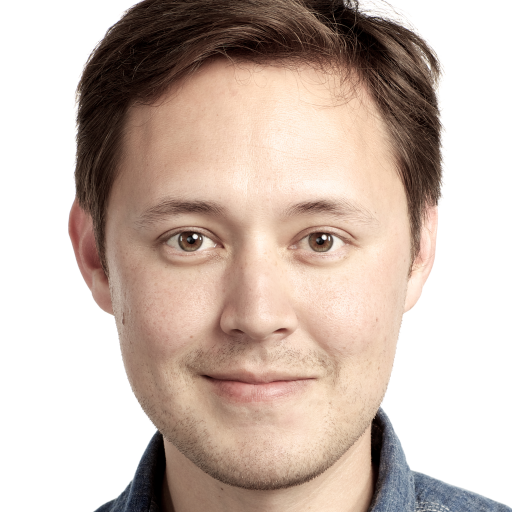}}
\subfloat[]{
\includegraphics[width=0.24\linewidth]{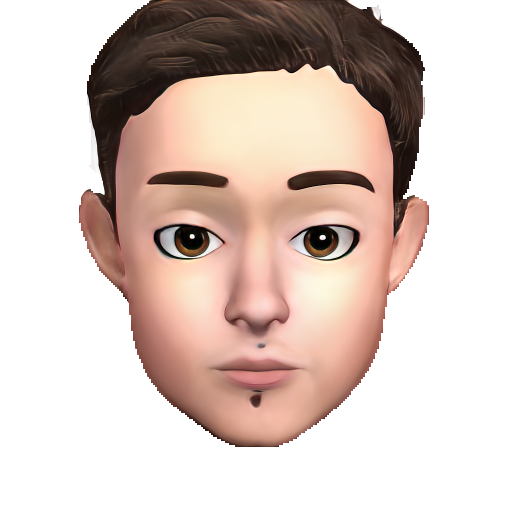}}
\subfloat[]{
\includegraphics[width=0.24\linewidth]{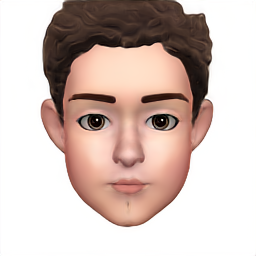}}
\subfloat[]{
\includegraphics[width=0.24\linewidth]{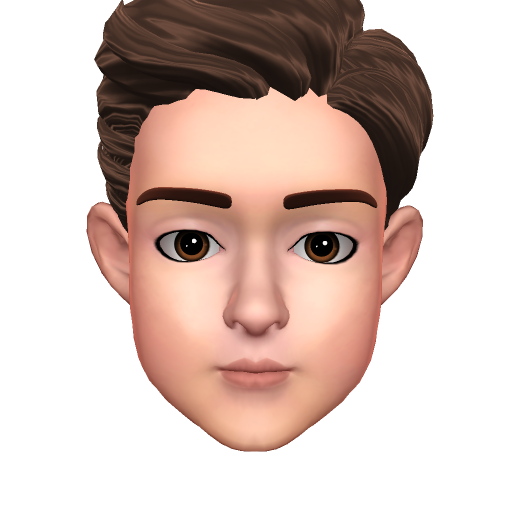}}
\vspace{-22pt}
\subfloat[]{
\includegraphics[width=0.24\linewidth]{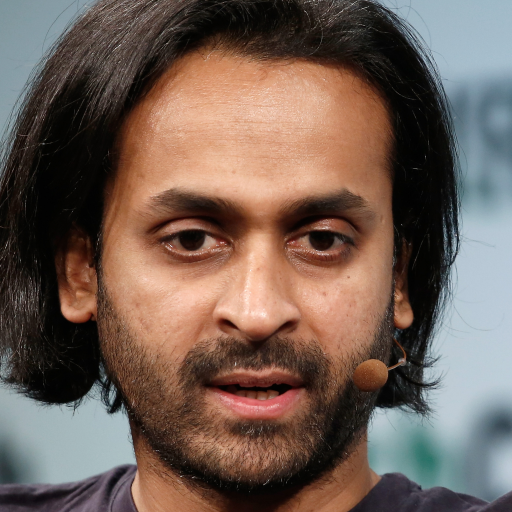}}
\subfloat[]{
\includegraphics[width=0.24\linewidth]{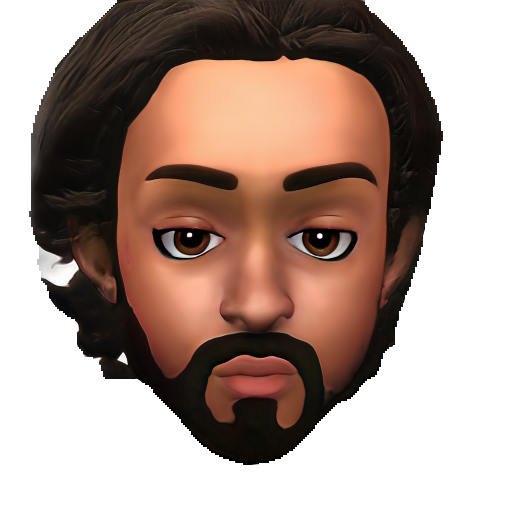}}
\subfloat[]{
\includegraphics[width=0.24\linewidth]{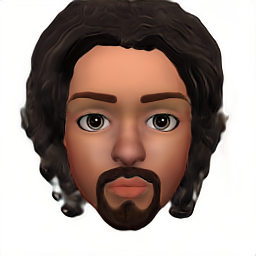}}
\subfloat[]{
\includegraphics[width=0.24\linewidth]{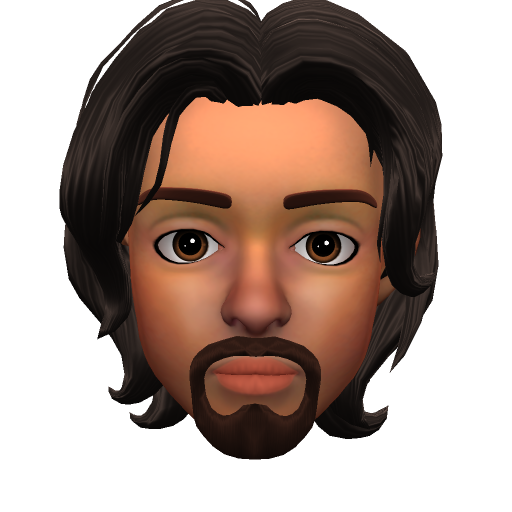}}
\vspace{-22pt}
\subfloat[(a) Input]{
\includegraphics[width=0.24\linewidth]{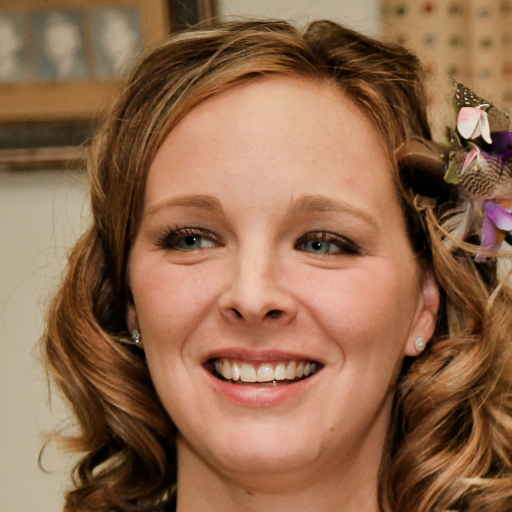}}
\subfloat[(b) Stylized]{
\includegraphics[width=0.24\linewidth]{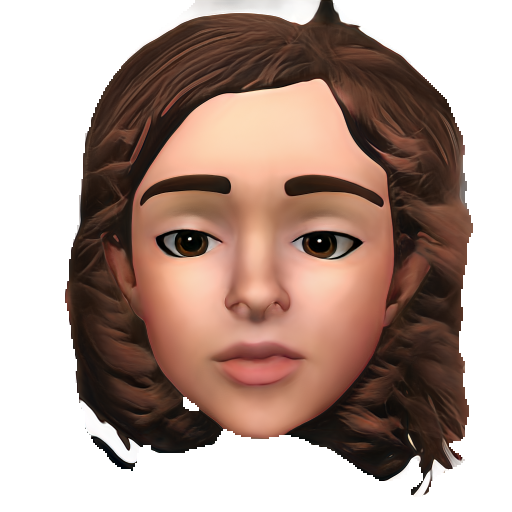}}
\subfloat[(c) Parameterized]{
\includegraphics[width=0.24\linewidth]{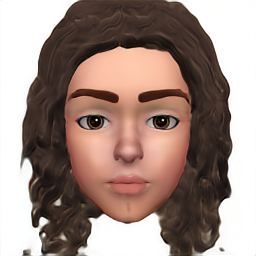}}
\subfloat[(d) Converted]{
\includegraphics[width=0.24\linewidth]{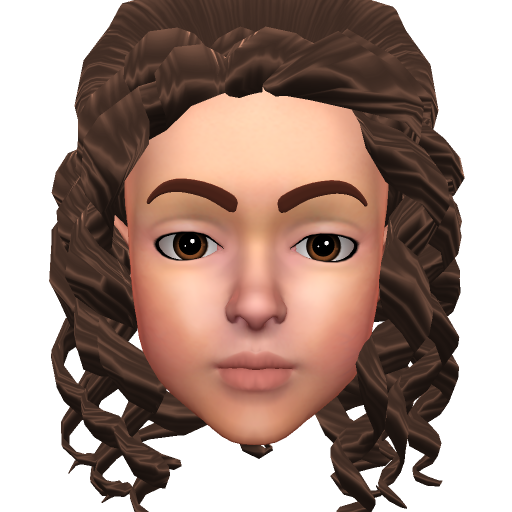}}
  \caption{Progressive domain crossing. 
  (b) At the portrait stylization stage, the images may still contain characteristics outside the domain of a graphics avatar, such as hair shape and non-frontal pose. (c) At the parameterization stage, the images are within the target domain, but may contain mixtures of components. (d) Finally, after vector conversion the output is a strict avatar vector which can be rendered by the graphics engine. Using FID as a measure of image distribution similarity, notice that each step brings us closer to the final target avatar domain.  \textcopyright Marcin Wichary, TechCrunch and Vanity Productions.
}
  \label{fig:domain}
\end{figure}

\begin{figure*}
\centering
\subfloat[]{
\includegraphics[width=0.155\linewidth]{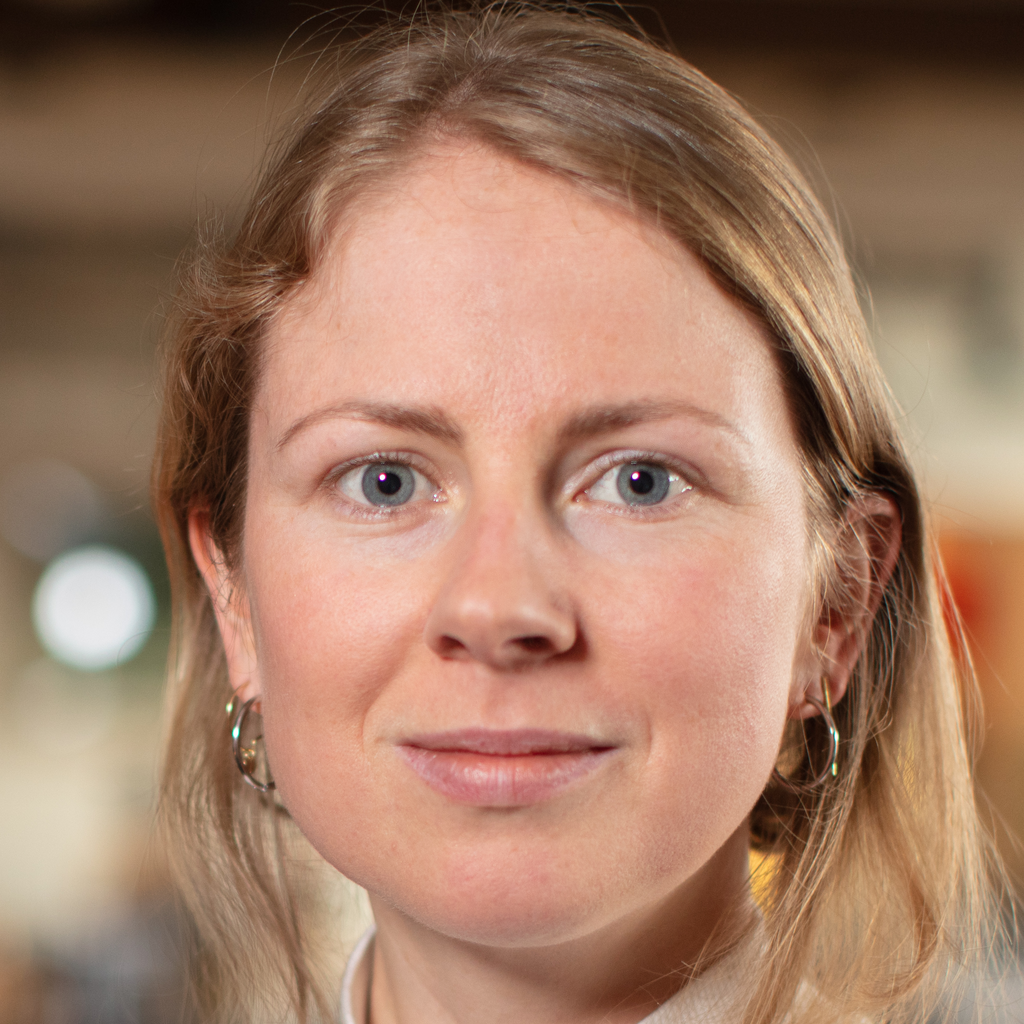}}
\subfloat[]{
\includegraphics[width=0.155\linewidth]{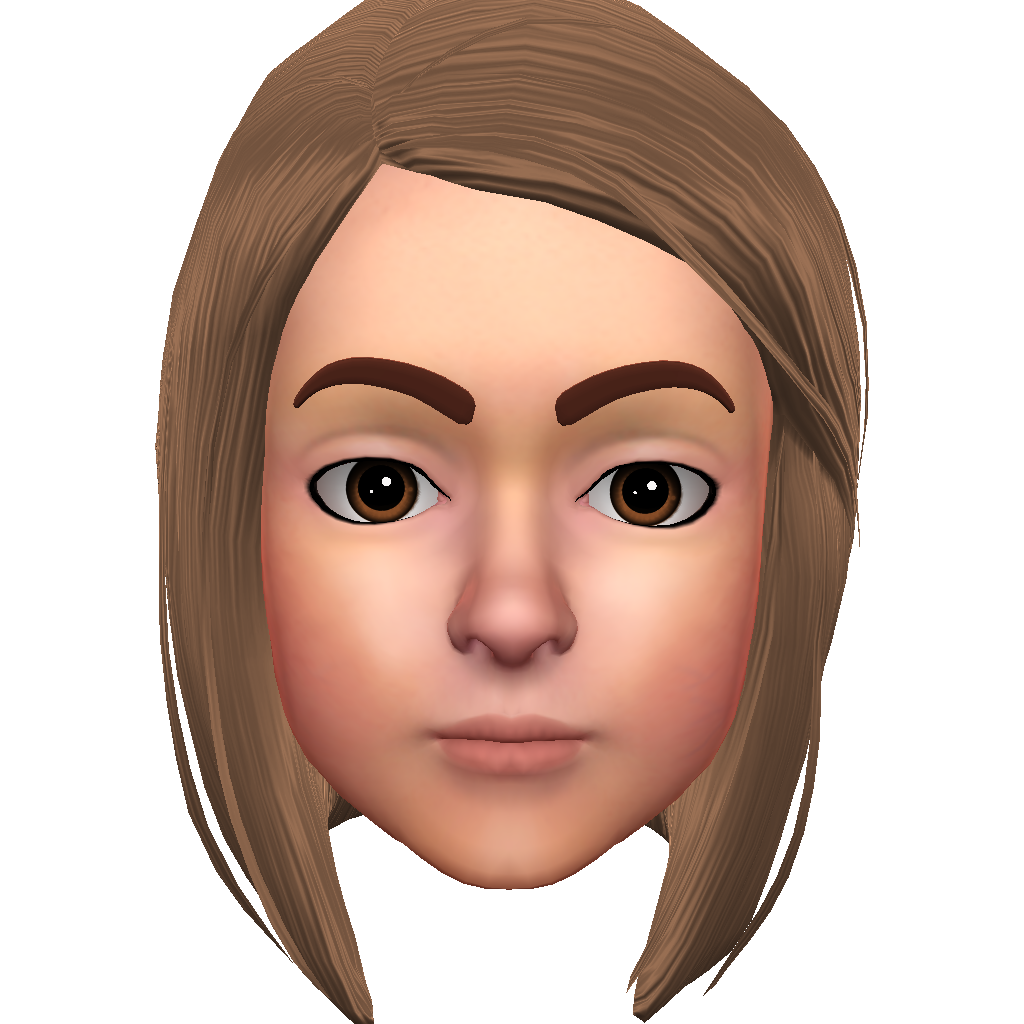}}
\subfloat[]{
\includegraphics[width=0.155\linewidth]{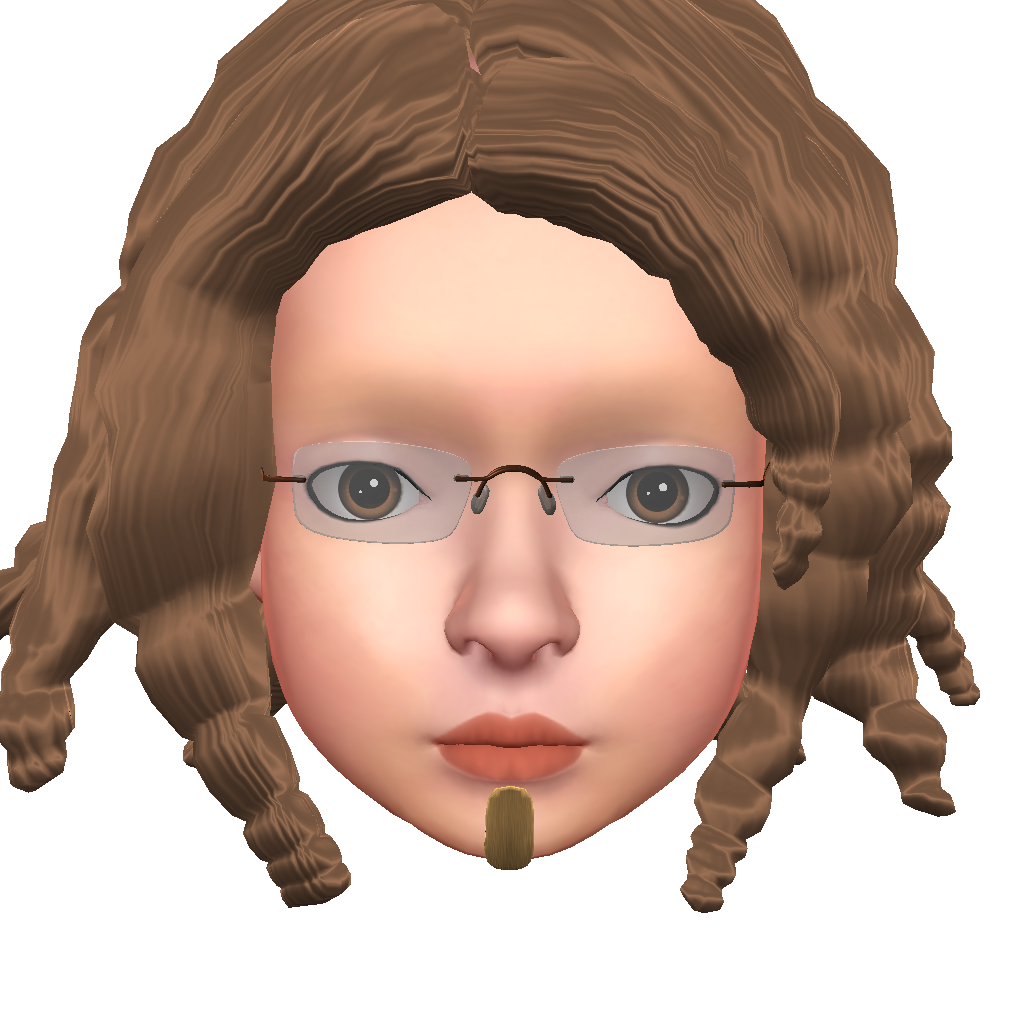}}
\subfloat[]{
\includegraphics[width=0.155\linewidth]{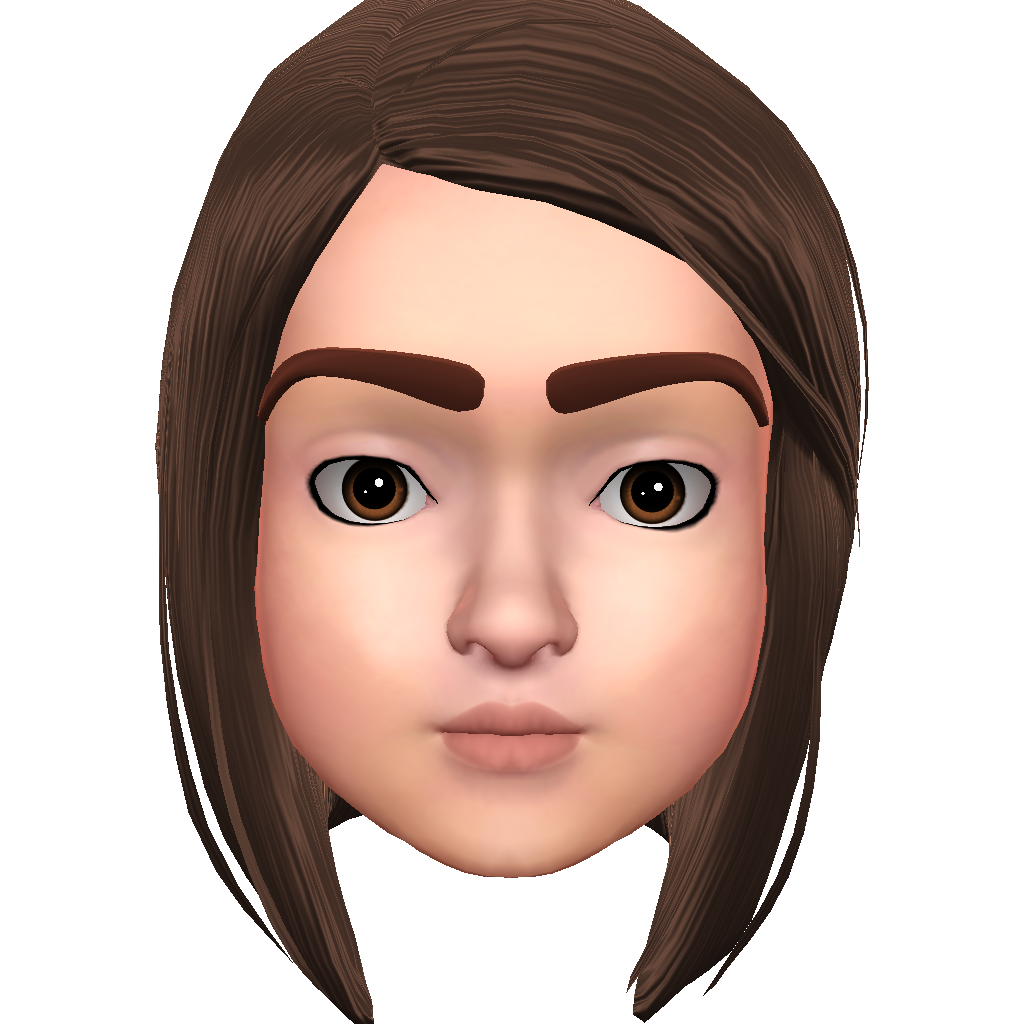}}
\subfloat[]{
\includegraphics[width=0.155\linewidth]{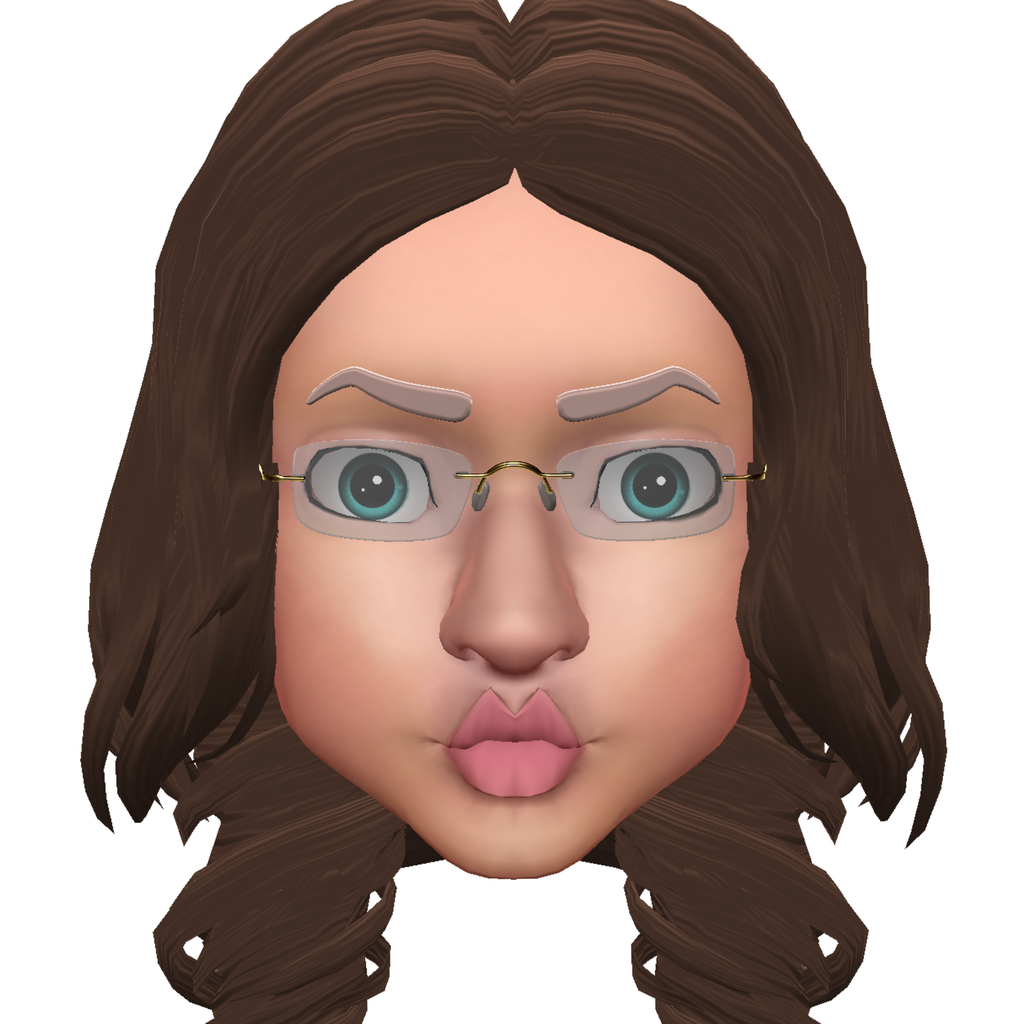}}
\subfloat[]{
\includegraphics[width=0.155\linewidth]{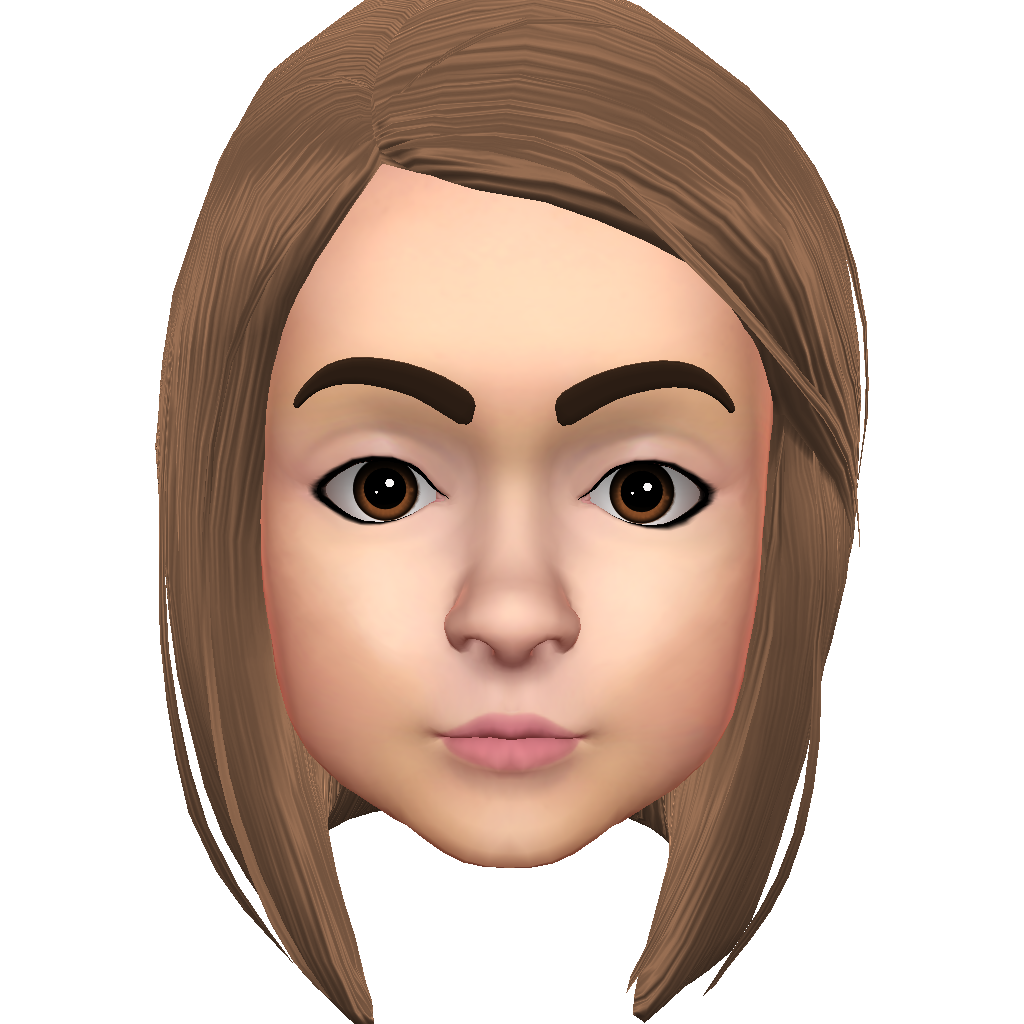}}
\vspace{-22pt}

\subfloat[]{
\includegraphics[width=0.155\linewidth]{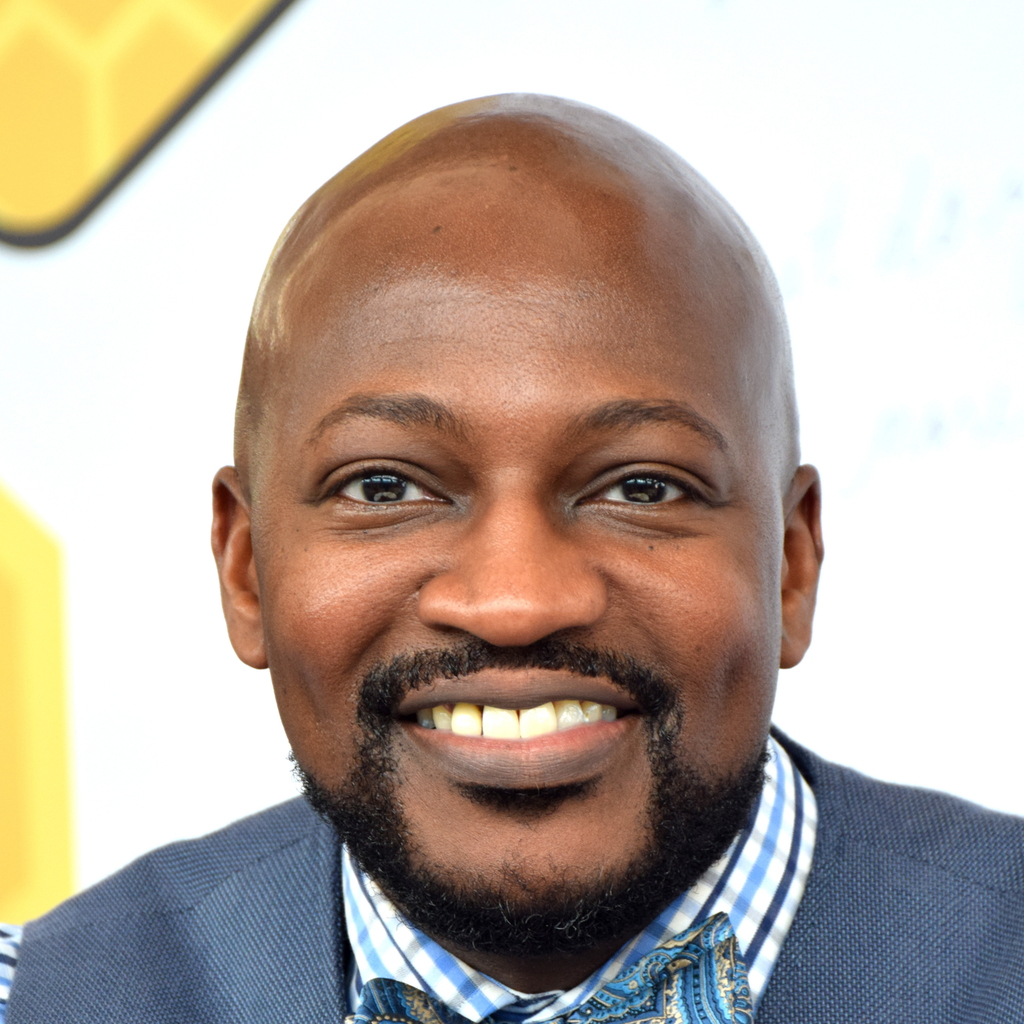}}
\subfloat[]{
\includegraphics[width=0.155\linewidth]{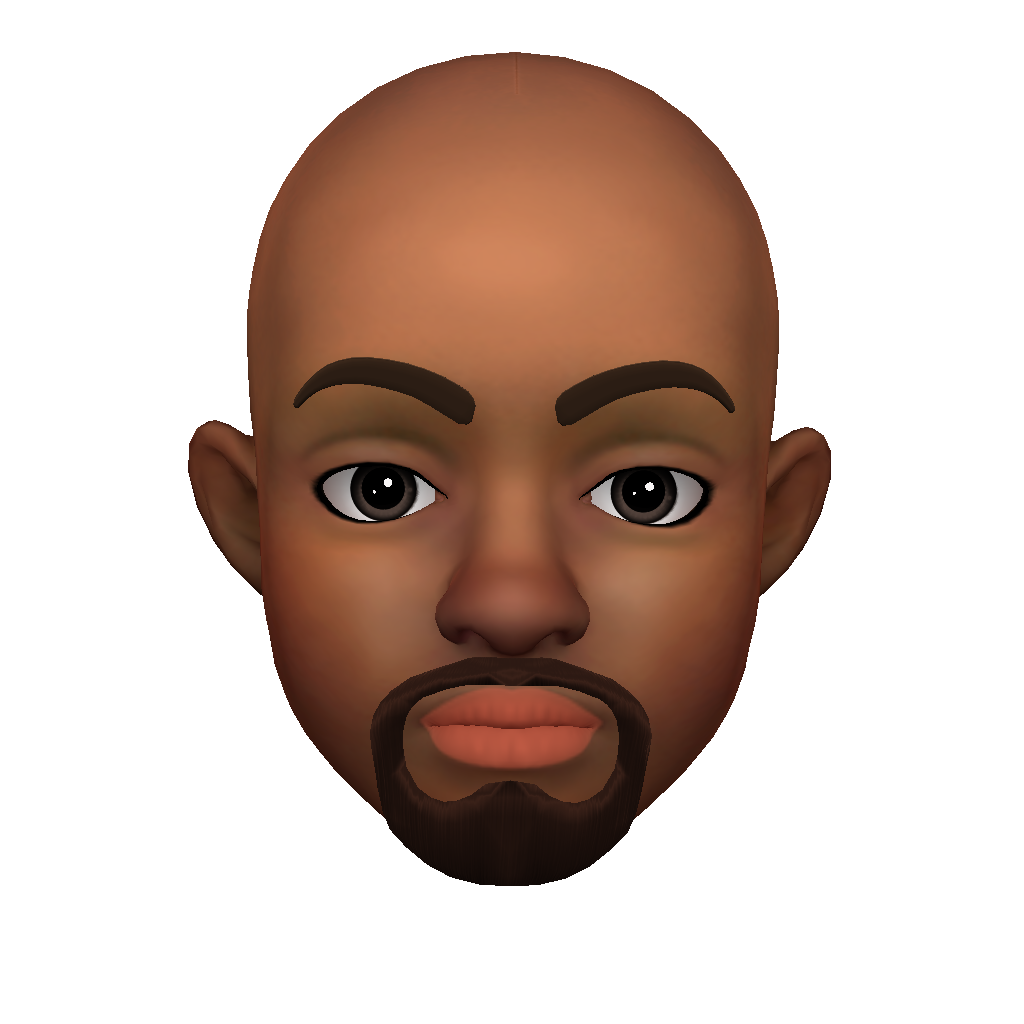}}
\subfloat[]{
\includegraphics[width=0.155\linewidth]{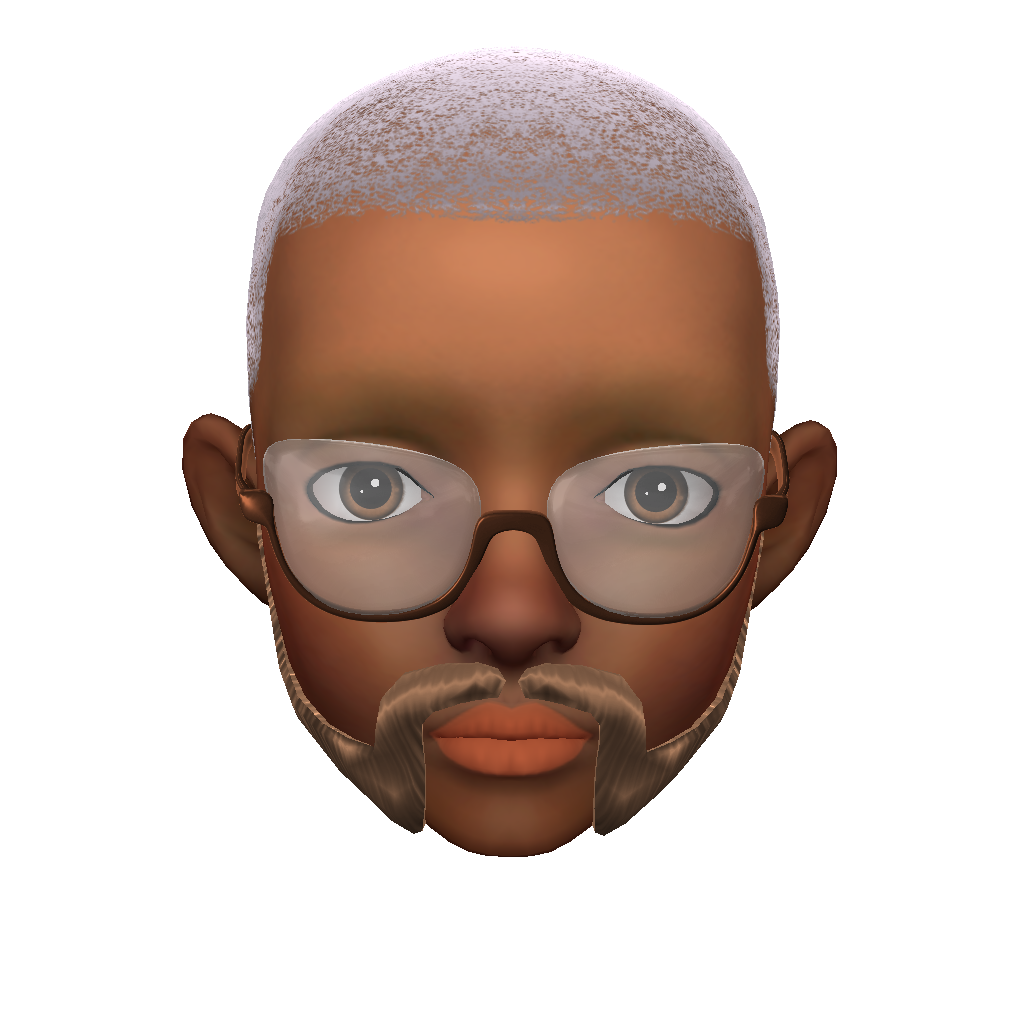}}
\subfloat[]{
\includegraphics[width=0.155\linewidth]{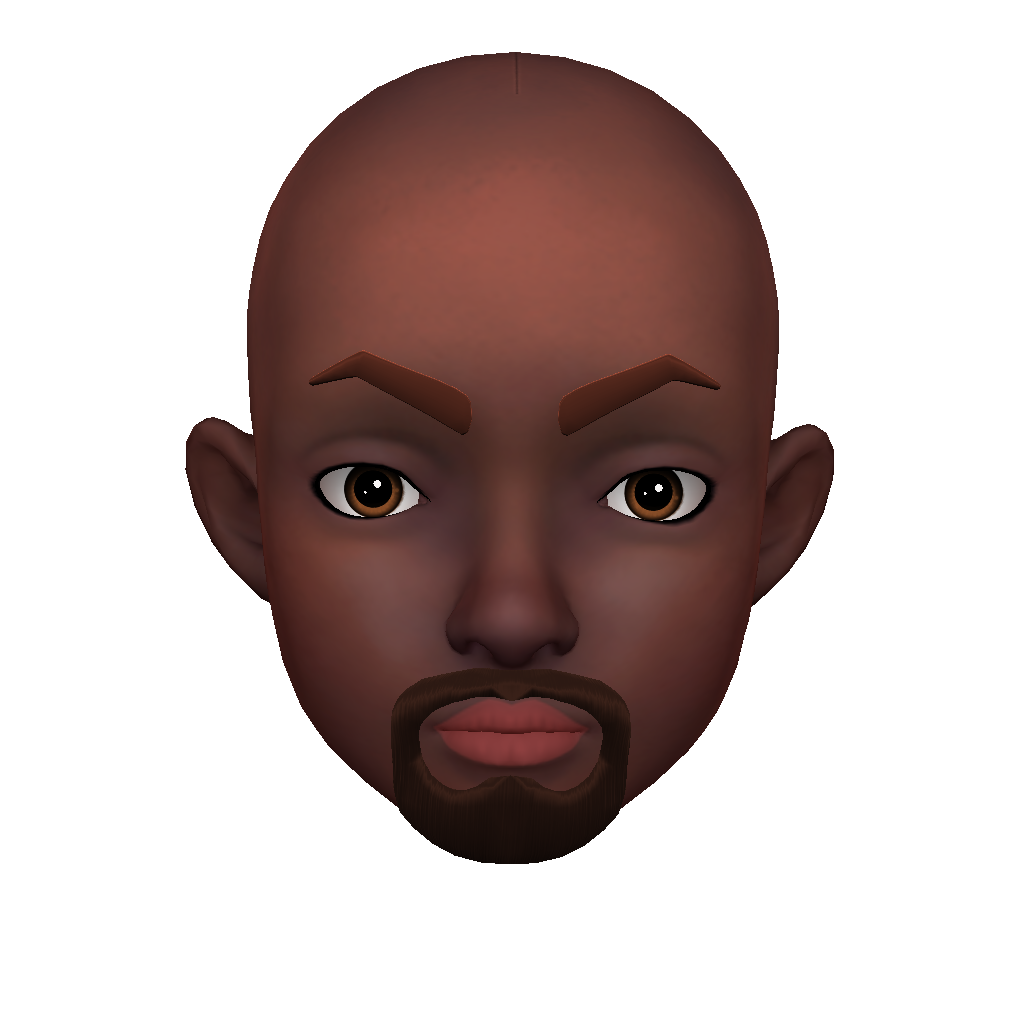}}
\subfloat[]{
\includegraphics[width=0.155\linewidth]{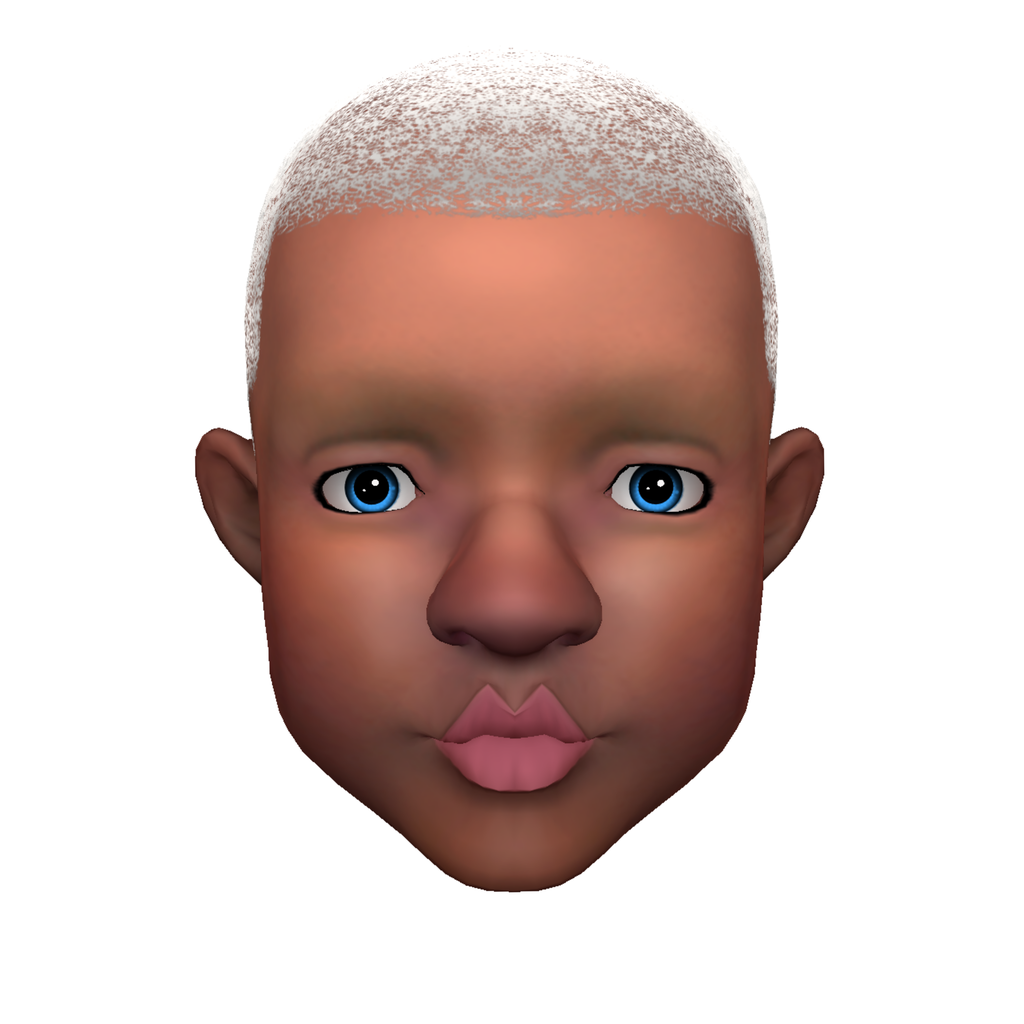}}
\subfloat[]{
\includegraphics[width=0.155\linewidth]{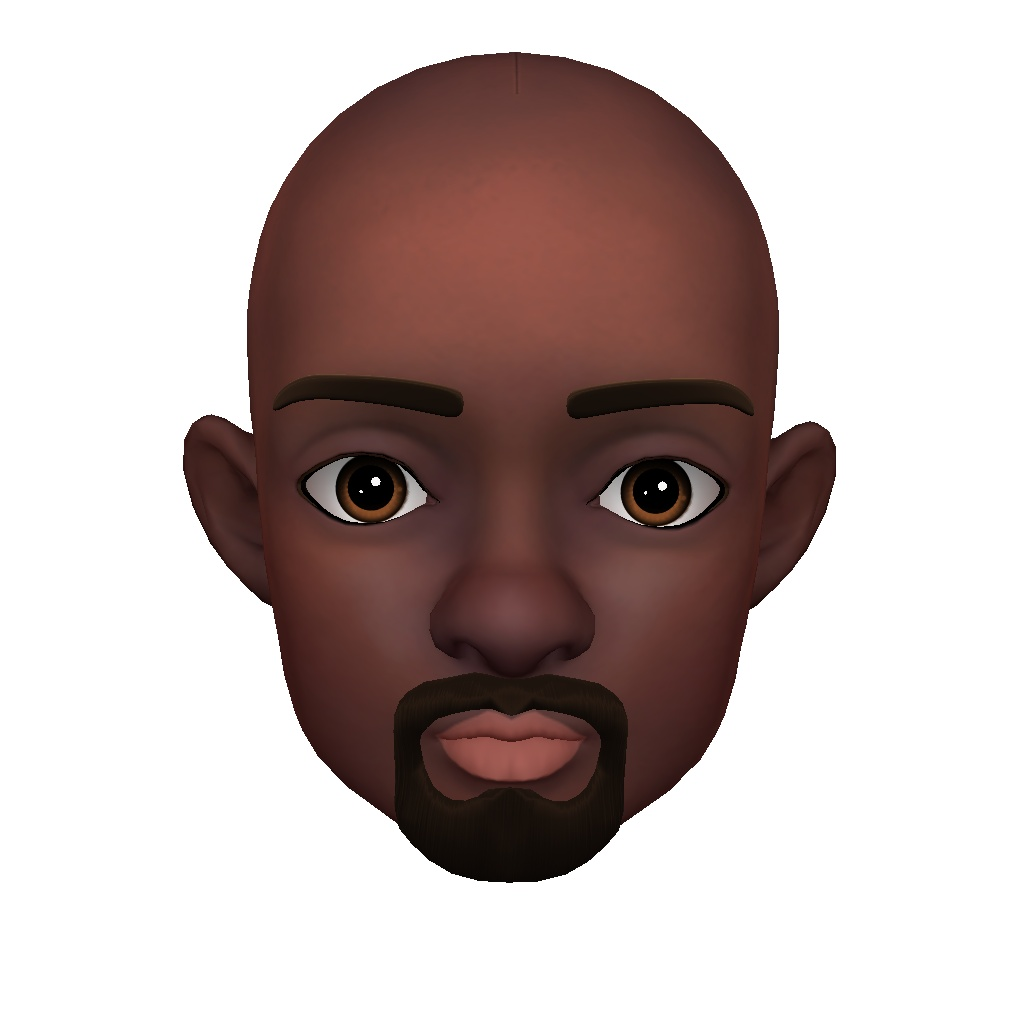}}
\vspace{-22pt}

\subfloat[]{
\includegraphics[width=0.155\linewidth]{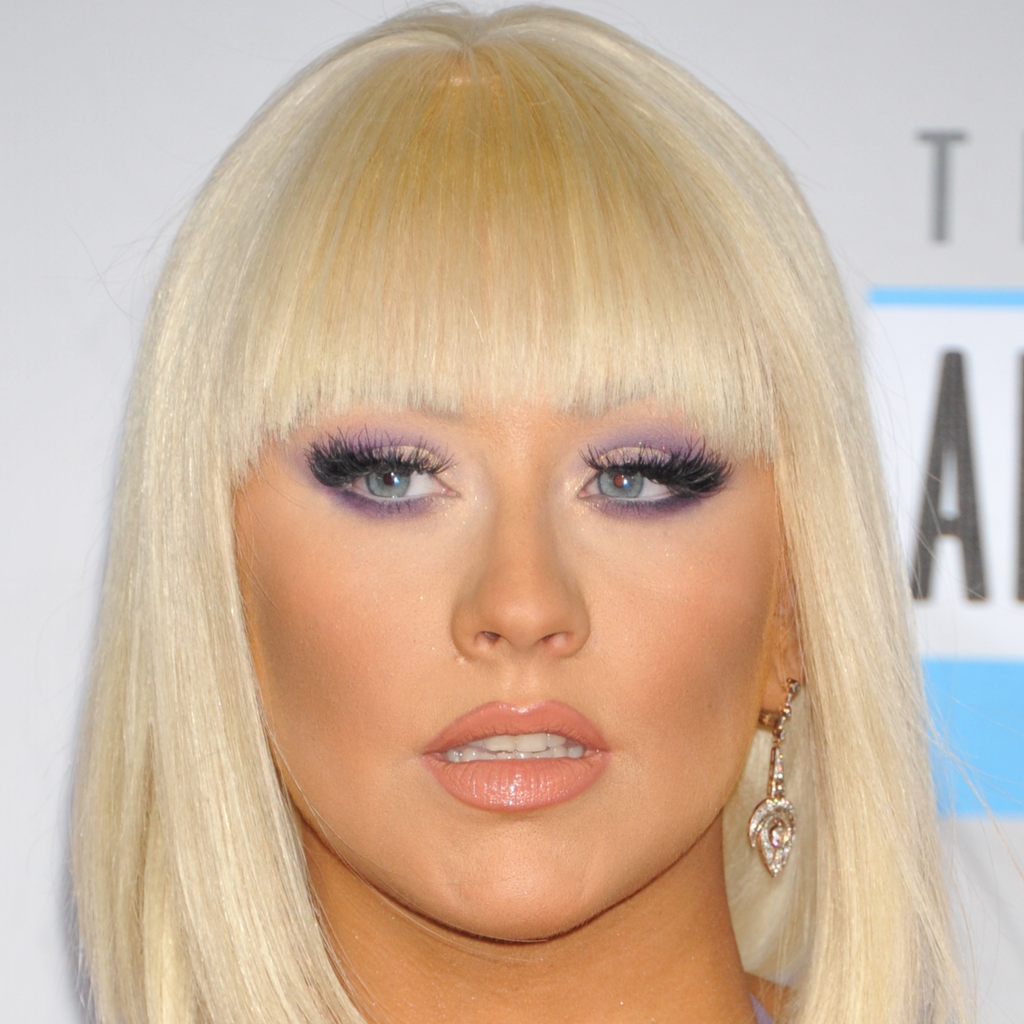}}
\subfloat[]{
\includegraphics[width=0.155\linewidth]{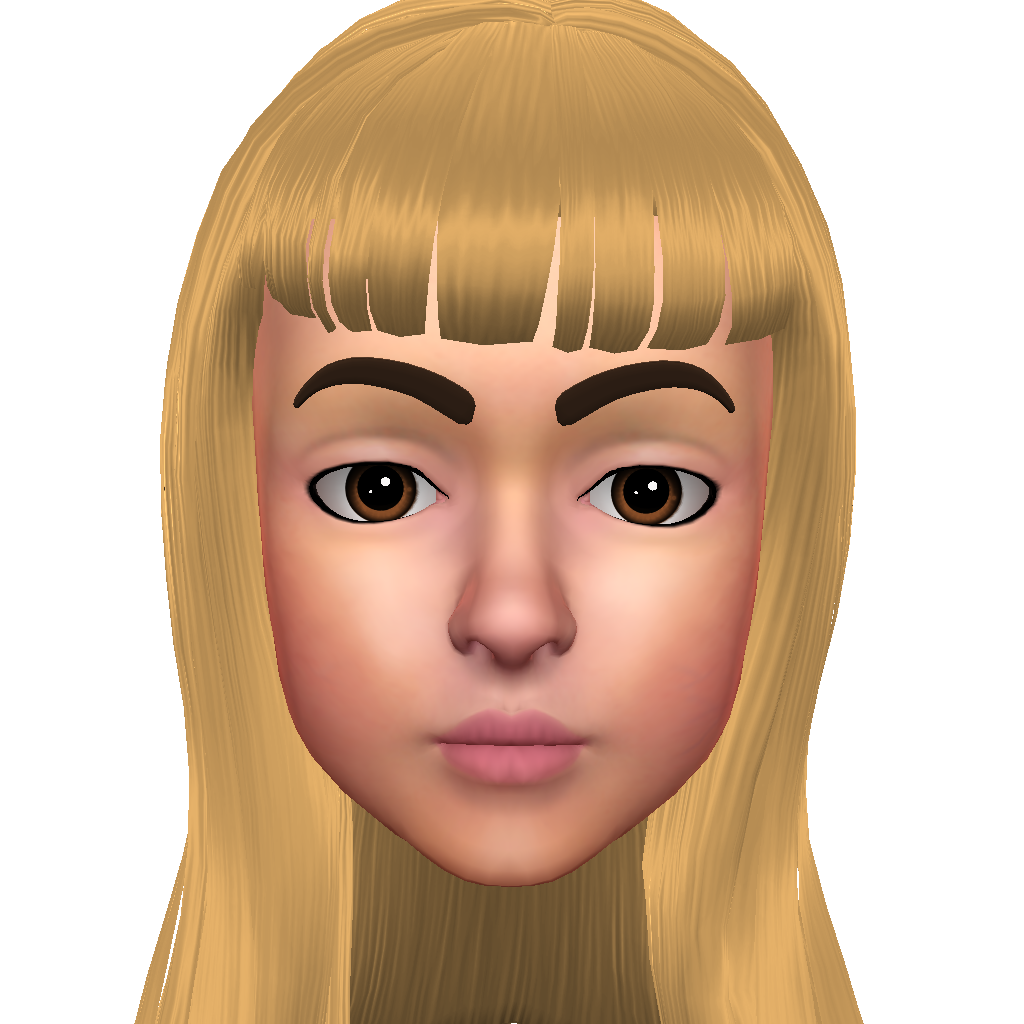}}
\subfloat[]{
\includegraphics[width=0.155\linewidth]{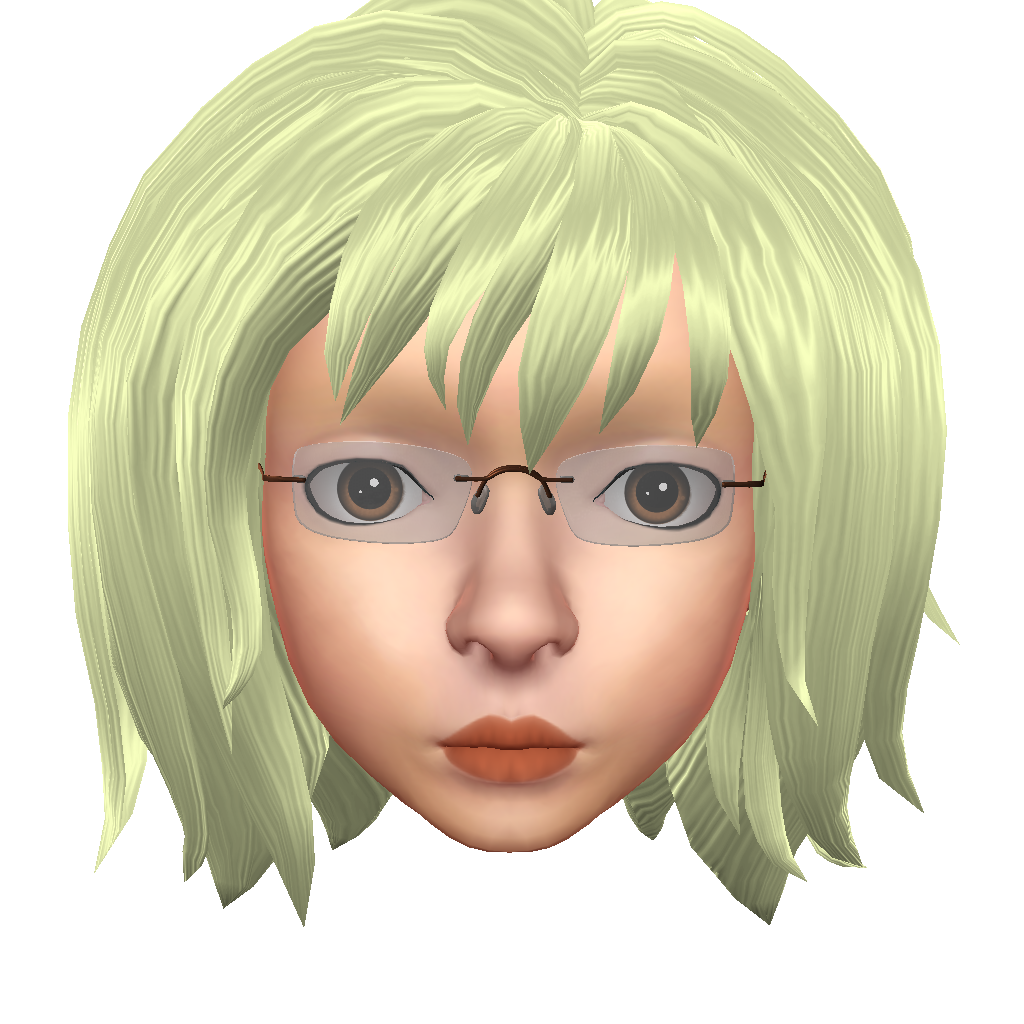}}
\subfloat[]{
\includegraphics[width=0.155\linewidth]{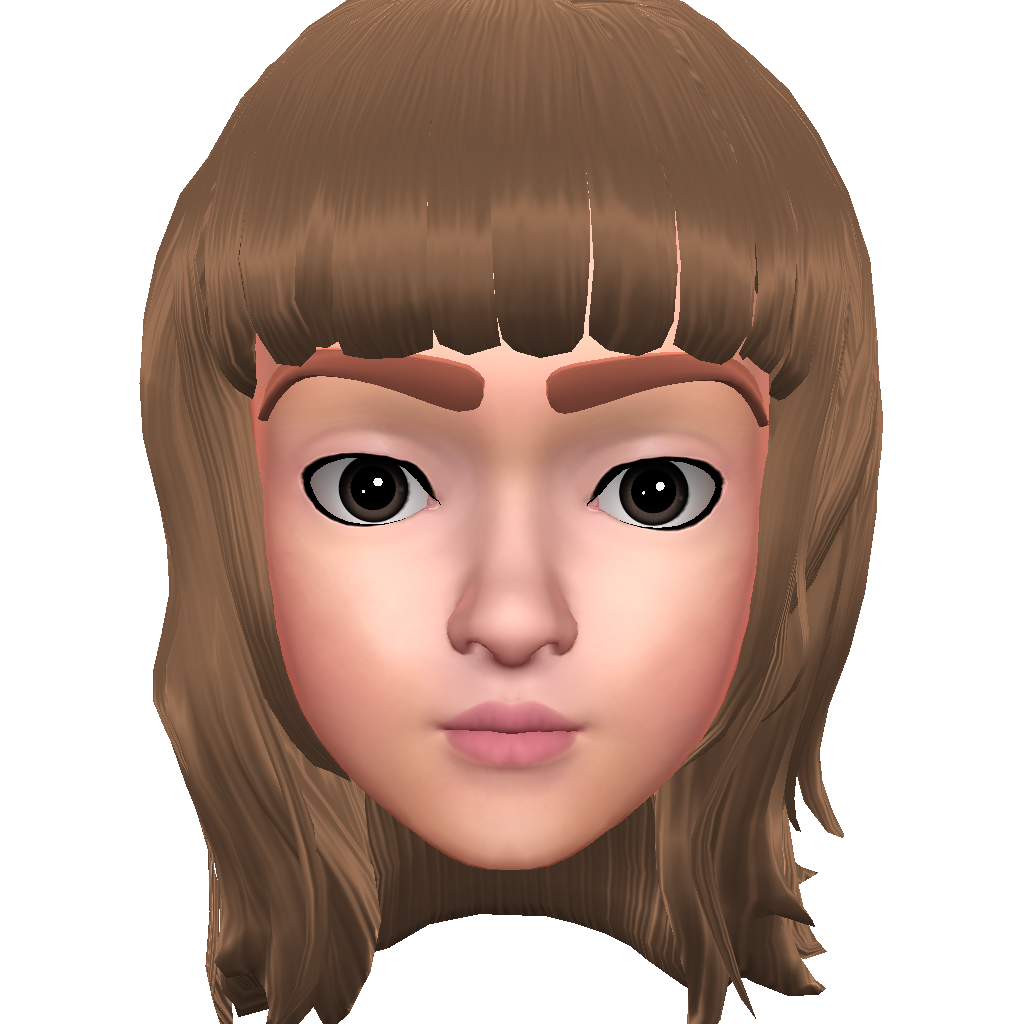}}
\subfloat[]{
\includegraphics[width=0.155\linewidth]{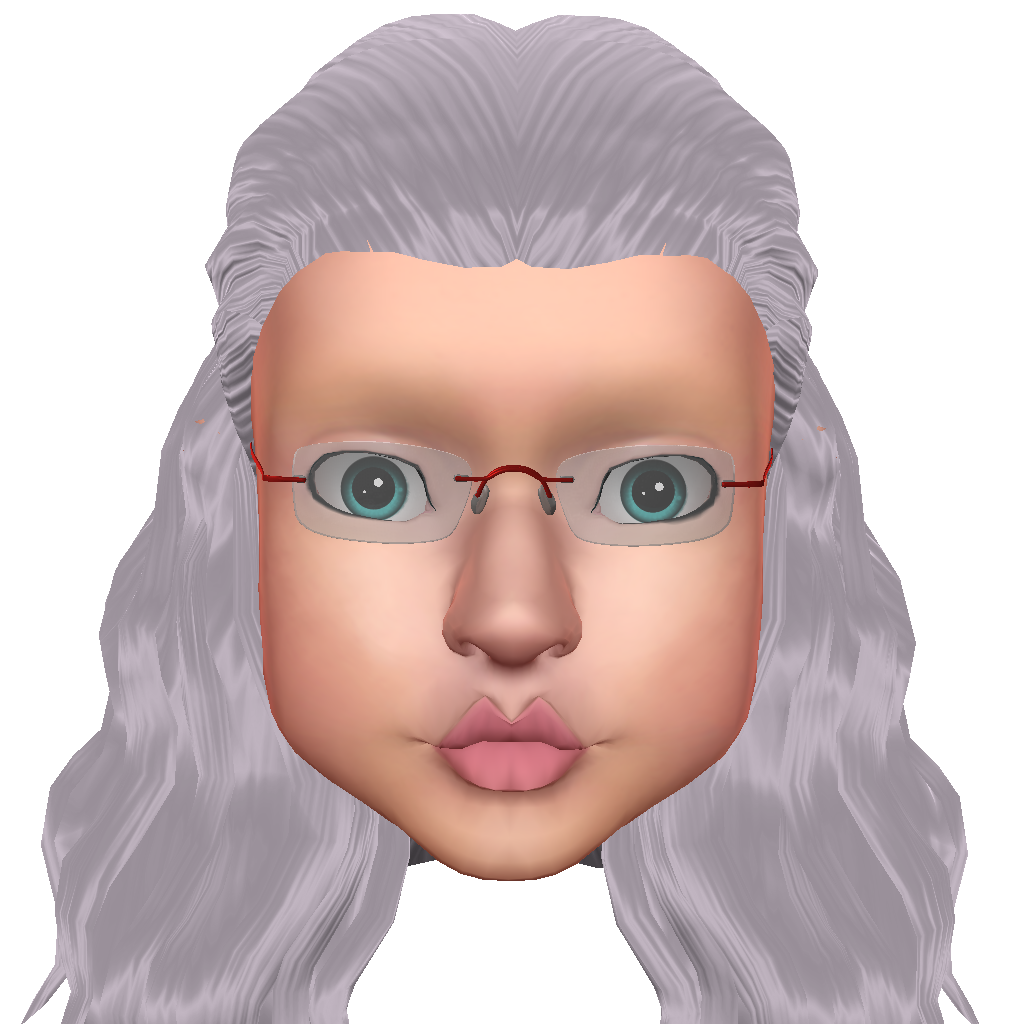}}
\subfloat[]{
\includegraphics[width=0.155\linewidth]{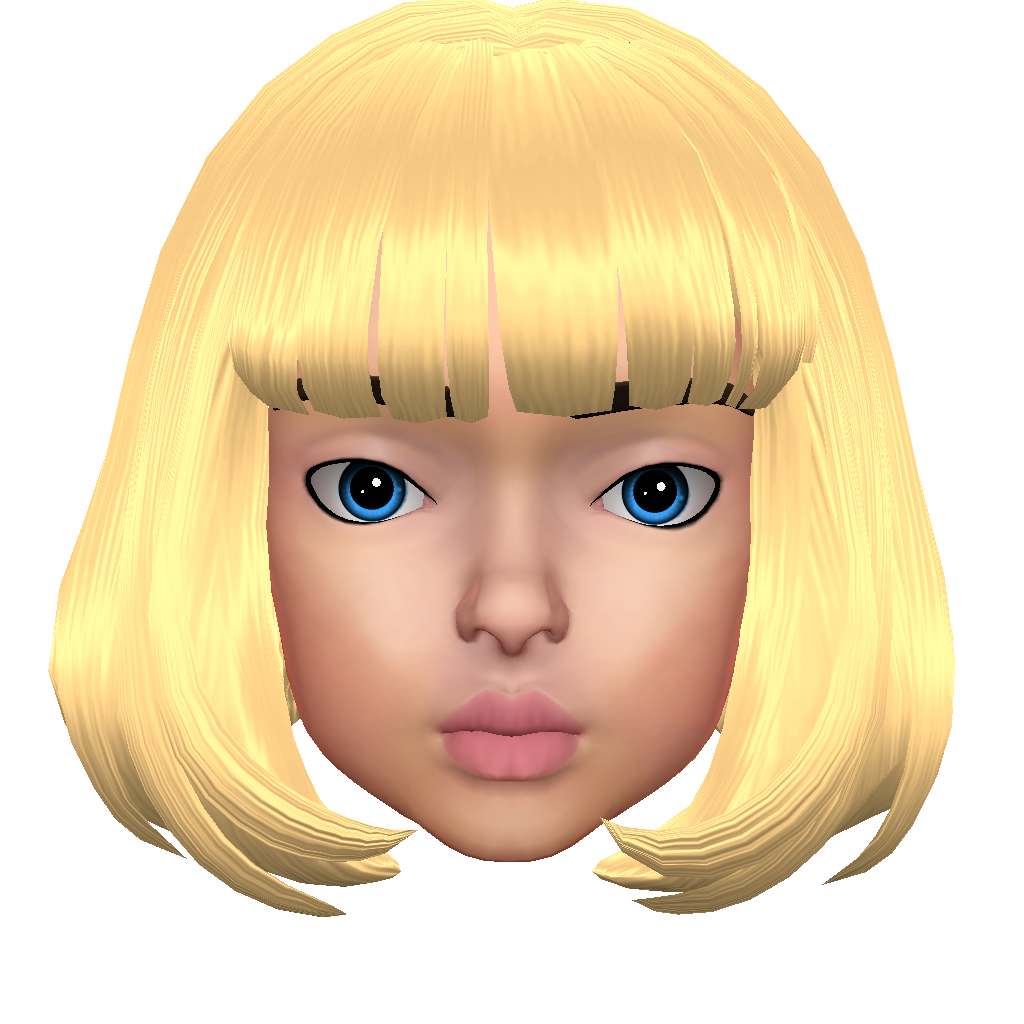}}
\vspace{-22pt}

\subfloat[(a) Input]{
\includegraphics[width=0.155\linewidth]{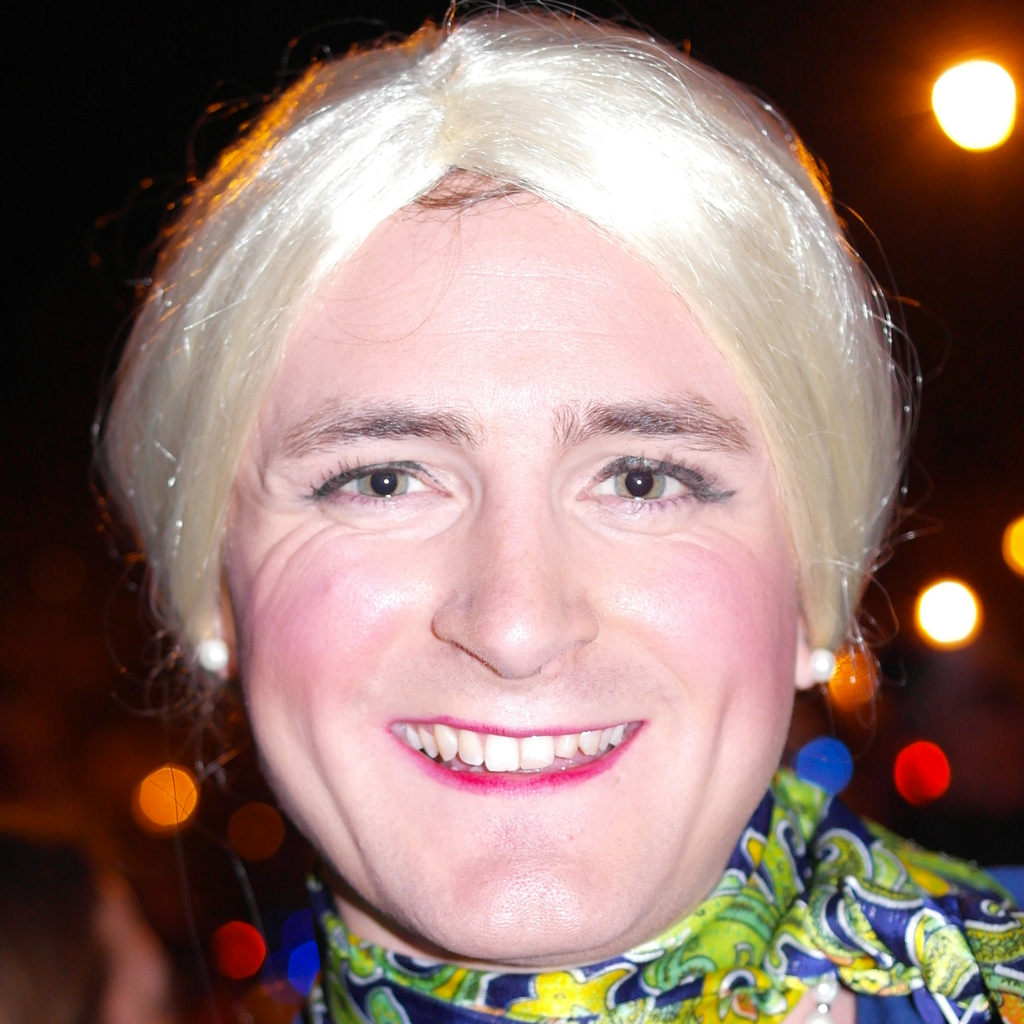}}
\subfloat[(b) Ours]{
\includegraphics[width=0.155\linewidth]{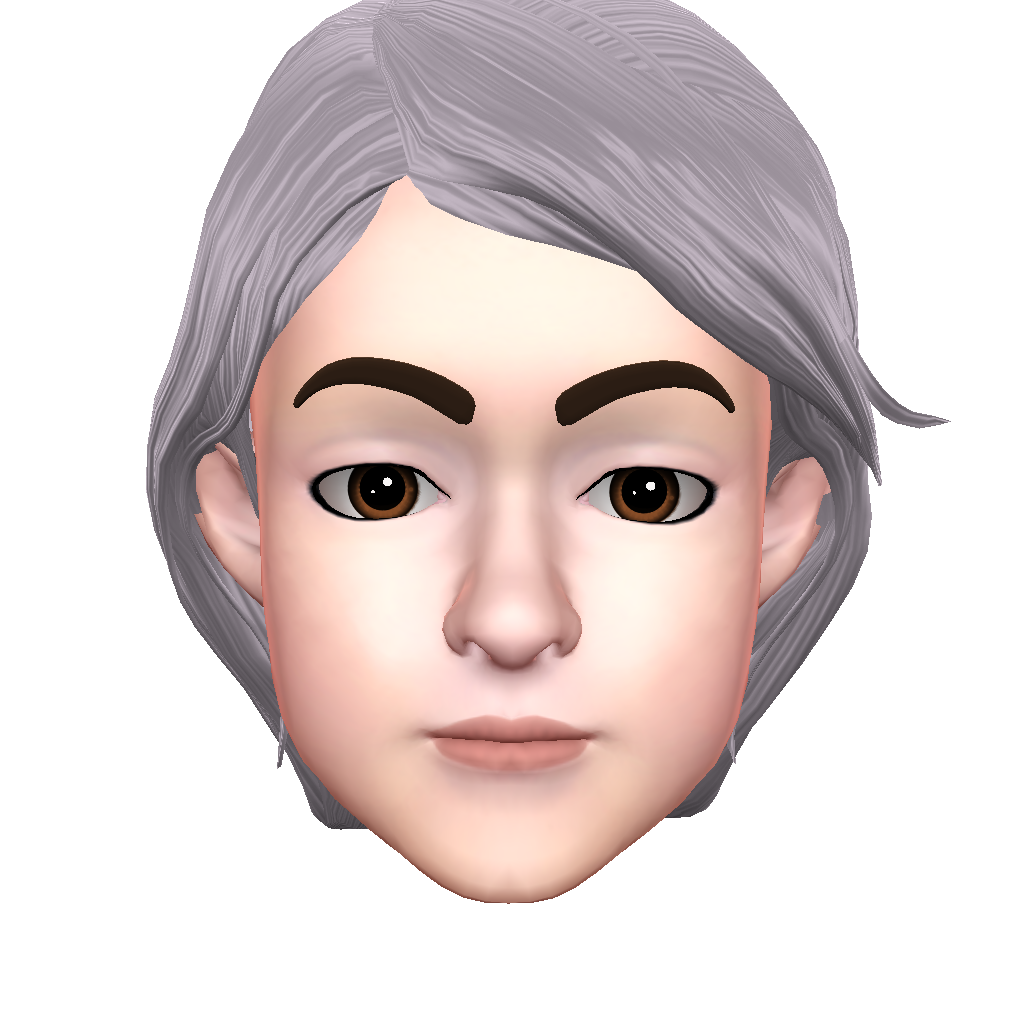}}
\subfloat[(c) CNN]{
\includegraphics[width=0.155\linewidth]{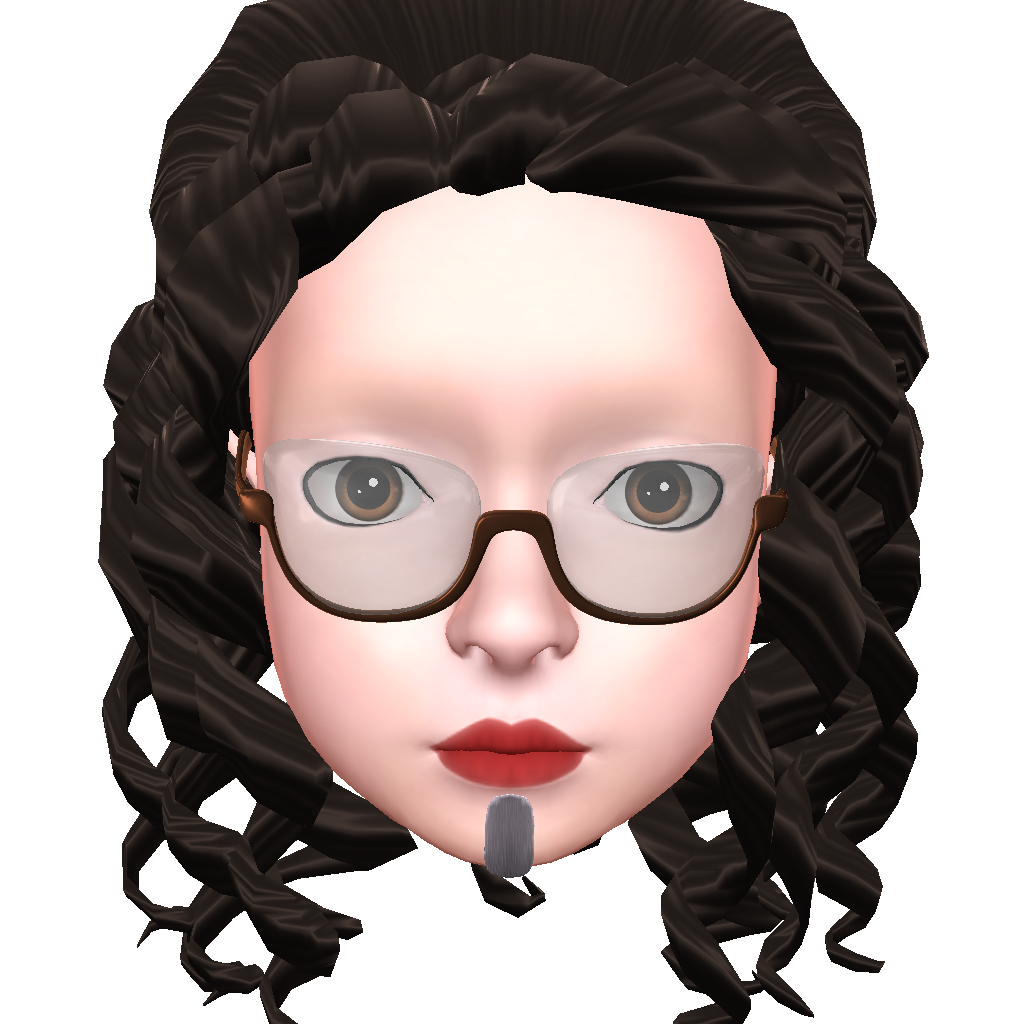}}
\subfloat[(d) Stylization + CNN]{
\includegraphics[width=0.155\linewidth]{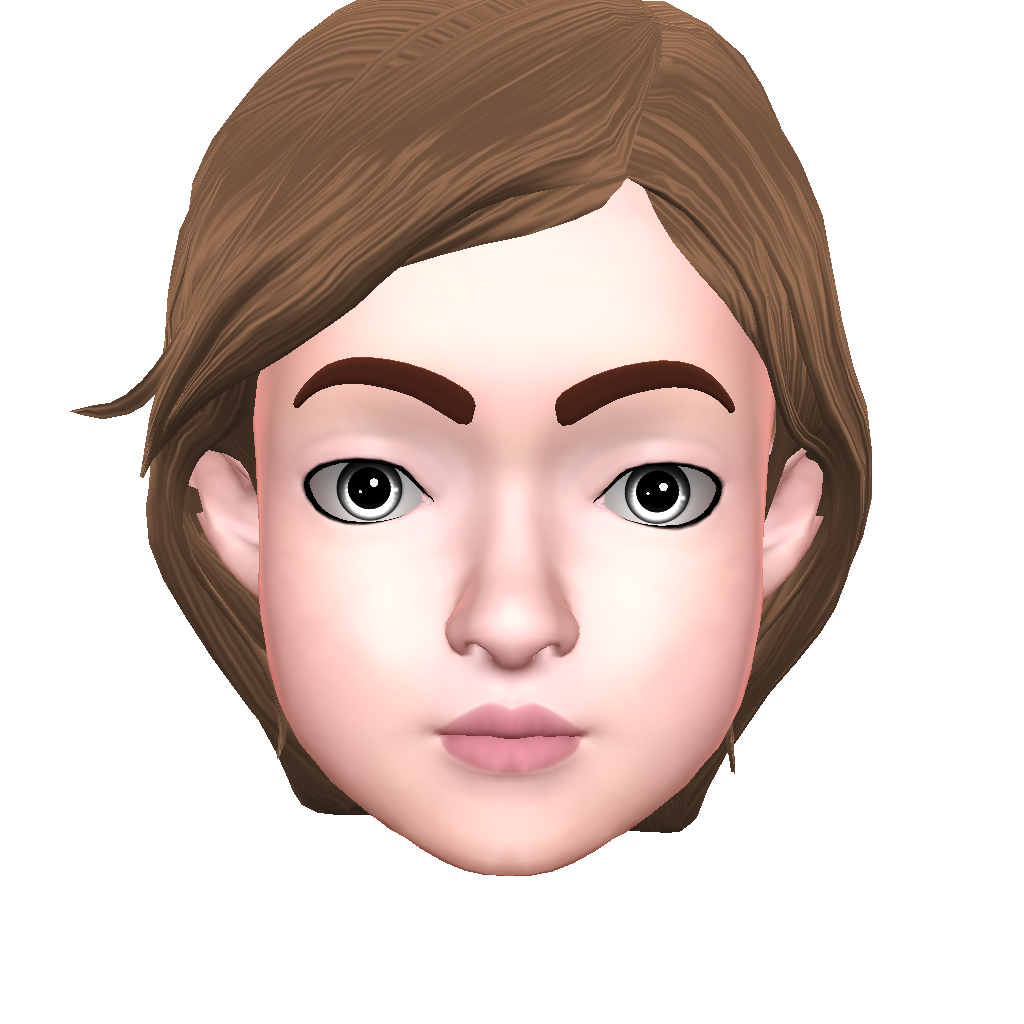}}
\subfloat[(e) F2P~\shortcite{shi2019face}]{
\includegraphics[width=0.155\linewidth]{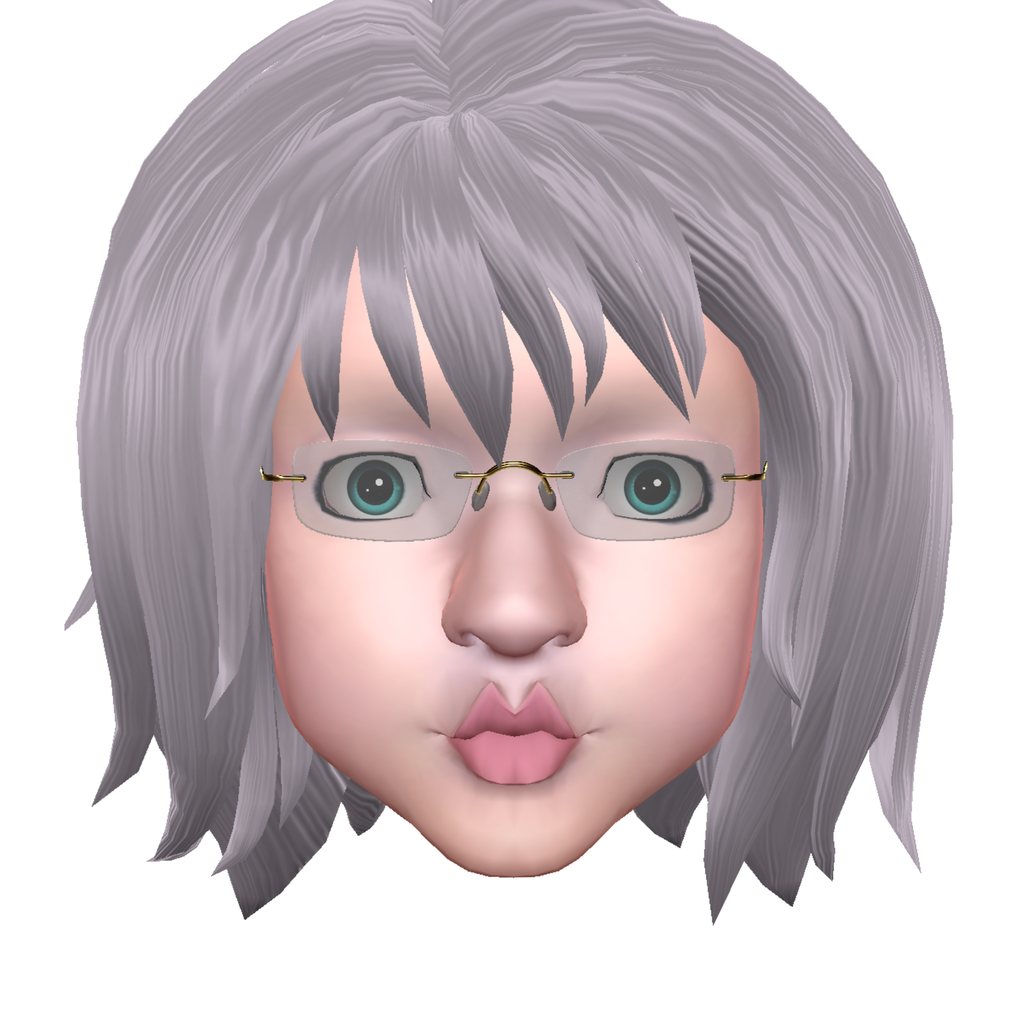}}
\subfloat[(f) Manual]{
\includegraphics[width=0.155\linewidth]{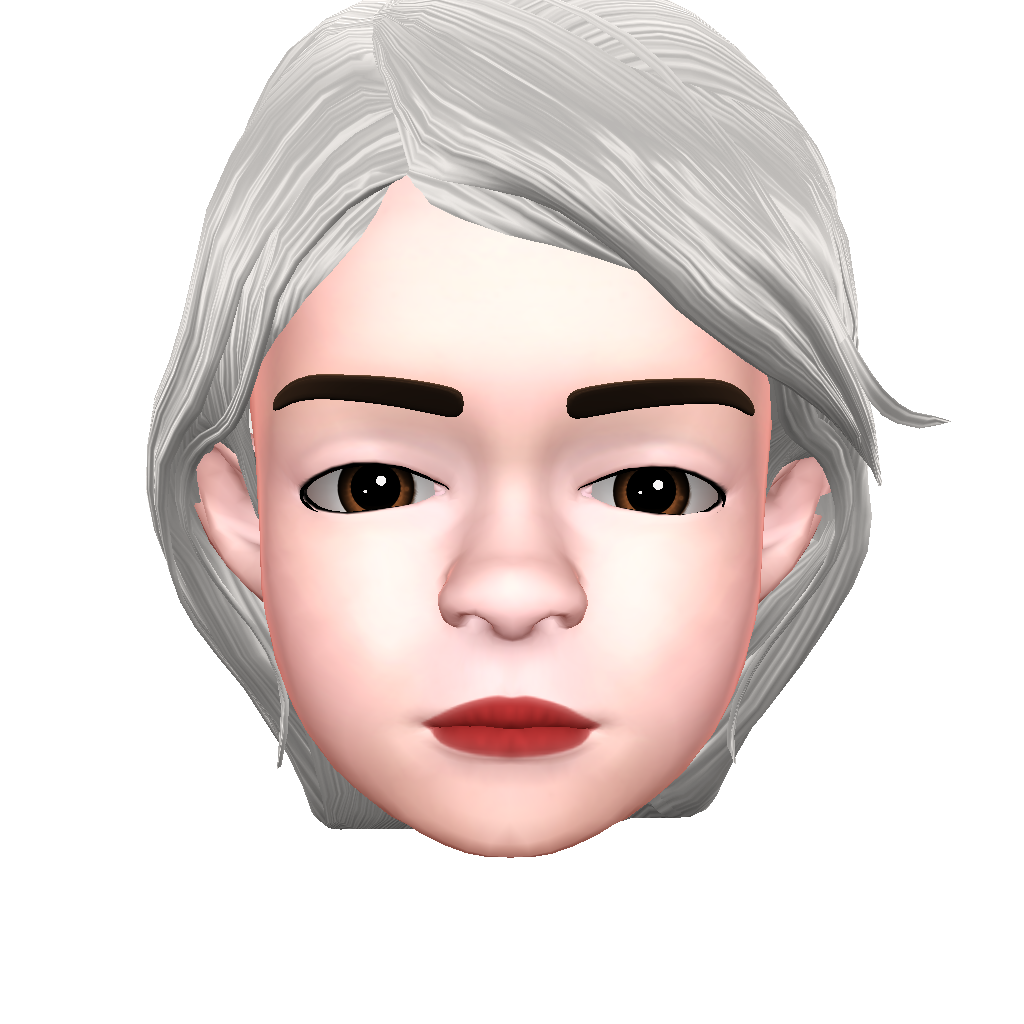}}

  \caption{Results comparison. (a) Given an input image, (b) our method produces an avatar in the target cartoon style that looks similar to the user.  (c) A CNN trained on synthetic data produces incorrect beard, hair style, and glasses on real image inputs due to the significant domain gap. (d) Applying the CNN instead to the results of stylization reduces the domain gap and thus improves results, however significant errors remain. (e) F2P, a baseline method intended to produce semi-realistic avatars does not consider the domain gap and thus produces poor results when used with stylized avatars~\cite{shi2019face}. (f) Manual results were created by expert-trained users. Our results approximate the quality obtainable through manual creation. \textcopyright Sebastiaan ter Burg, NIGP, YayA Lee and S Pakhrin.}
  \label{fig:compare_methods}
\end{figure*}

\section{Experimental Analysis}

\paragraph{Cascaded Domain Bridging:}
To illustrate the effect of each stage in the proposed three-stage pipeline, the intermediate results are visualized in Fig.~\ref{fig:domain}.
Notice how the three stages progressively bridge the domain gap between real images and stylized avatars.  To measure how close the intermediate results are in comparison to the target avatar domain, we use the perceptual metric FID~\cite{kilgour2019frechet}. Notice that the FID becomes lower after each stage, demonstrating the gradual reduction of domain gap.

\paragraph{Visual Comparison with Baseline Methods:}
We compare the proposed method against a number of baselines, shown in Fig.~\ref{fig:compare_methods}. \textit{CNN} is a naive supervised method using rendered avatar images to train a CNN~\cite{sandler2018mobilenetv2} to fit ground truth parameters. The CNN is then applied on the segmented head region of the input image. The domain gap causes the CNN to make poor predictions. \textit{Our stylization + CNN} narrows the domain gap by applying the CNN to our stylized results. This noticeably improves predictions, however errors in hair and face coloration remain. Since the CNN is only trained on synthetic data, it cannot regress the parameters properly due to the domain gap between training and test data even for stylized images. \textit{F2P~\shortcite{shi2019face}} is a self-supervised optimization-based method designed for semi-realistic avatars. This method fails to do well, likely because it naively aligns the segmentation of real faces and the avatar faces, without considering the domain gap. \textit{Manual} results were created by expert-trained users. Given a real face, the users were asked to build an avatar that preserves personal identity while demonstrating high attractiveness based on their own judgement.  Visually, our method shows a quality similar to manual creation, demonstrating the utility of our method.

\paragraph{Numerical Comparison with Baseline Methods:}
To evaluate results numerically we rely on judgements made by human observers recruited through Amazon Mechanical Turk~\footnote{https://www.mturk.com/}. We conduct two user studies for quantitative evaluation: \textit{Attribute Evaluation} and \textit{Matching}. We perform attribute evaluation to evaluate whether users believe that specific identity attributes such as hair color and style match the source photograph using a yes/no selection. 330 opinions were collected for each of 6 attributes.  Table~\ref{tab:user} shows results, indicating that our method retains photograph identity better than the baseline. In the matching task, we evaluate whether an avatar retains personal identity overall. Four random and diverse images were used to create avatars, and the subject must choose which is the correct match to a specific photograph.  A total of 990 judgements were collected. Avatars created with our method were identified correctly significantly more often than baseline methods, approaching the level of manually created avatars. 






\begin{table}
    \caption{Numerical results from two user studies. Our method is judged to produce better avatars than the baseline methods, approaching the quality of manual work. Attribute evaluation: judge whether a specific attribute of the created avatar matches the human image. Matching: choose the correct one out of four avatars which matches the human image. }
    \centering
    \scalebox{0.87}{
    \begin{tabular}{lccccccc}
    \toprule
    & \multicolumn{6}{c}{Attribute Evaluation} &\multirow{3}*{\tabincell{c}{Match \\Task}}\\
    \cmidrule{2-7}
           &  \multirow{2}*{\tabincell{c}{beard \\type}} & \multirow{2}*{\tabincell{c}{face \\shape}}&
           \multirow{2}*{\tabincell{c}{ brow\\type}} &\multirow{2}*{\tabincell{c}{hair \\color}}& \multirow{2}*{\tabincell{c}{hair \\style}}& \multirow{2}*{\tabincell{c}{skin \\tone}}\\
           \\
           \midrule

    F2P~\shortcite{shi2019face}  &0.36 & 0.46 & 0.22 & 0.21 & 0.12 & 0.36 & 0.67   \\ 
    CNN                &0.17& 0.54   & 0.22 & 0.46  & 0.30 & 0.50  & 0.57   \\
    Stylization+CNN  &0.45&0.69  & 0.38  &0.57 & 0.43  & 0.66  & 0.82 \\
    Ours  & \textbf{0.82}& \textbf{0.94}  & \textbf{0.88} & \textbf{0.82}&\textbf{0.72}& \textbf{0.82}  & \textbf{0.92} \\
    \midrule
    Manual & 0.94& 0.97 & 0.85  & 0.90  & 0.86  &0.94   & 0.96 \\
    \bottomrule
    
    \end{tabular}}

    \label{tab:user}
\end{table}

\begin{table}
        \caption{Ablation study for mapper training losses. Users picked the best matching avatar from the six candidates produced by loss combinations. The scores show the fraction of each combination picked. $\mathcal{L}_{LPIPS}$ is the most significant component, while $\mathcal{L}_{id}$ and $\mathcal{L}_{color}$ also improve the results.}
    \centering
    \scalebox{0.87}{
    \begin{tabular}{cccccc}
    \toprule
    ID    & LPIPS  & ID+LPISP & ID+Color & LPIPS+Color & ID+LPIPS+Color \\
    \midrule
    9.3\% & 17.8\% & 18.7\%   & 14.8\%   & 19.1\%      & \textbf{20.3\%}        \\
    \bottomrule
    \end{tabular}}

    \label{tab:loss}
\end{table}

\begin{figure}
\subfloat[]{
\includegraphics[width=0.245\linewidth]{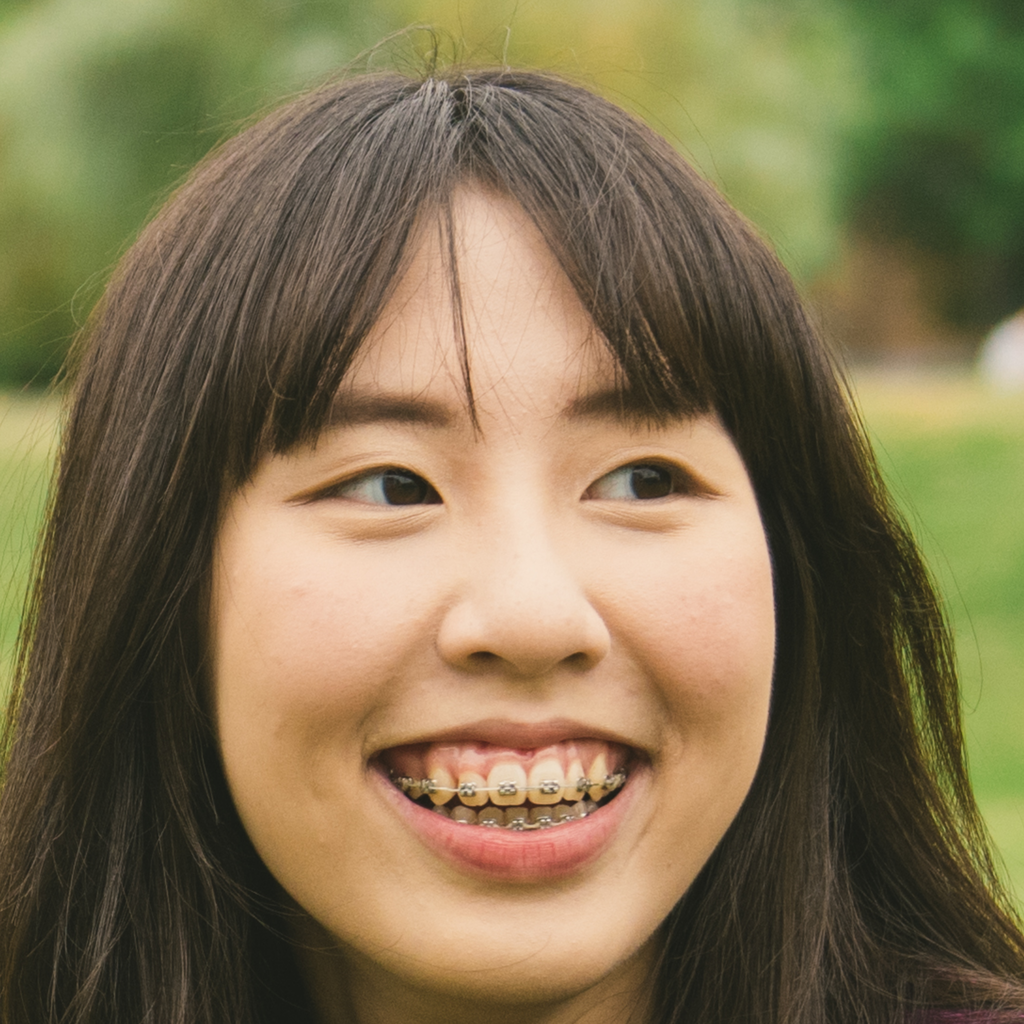}}
\subfloat[]{
\includegraphics[width=0.245\linewidth]{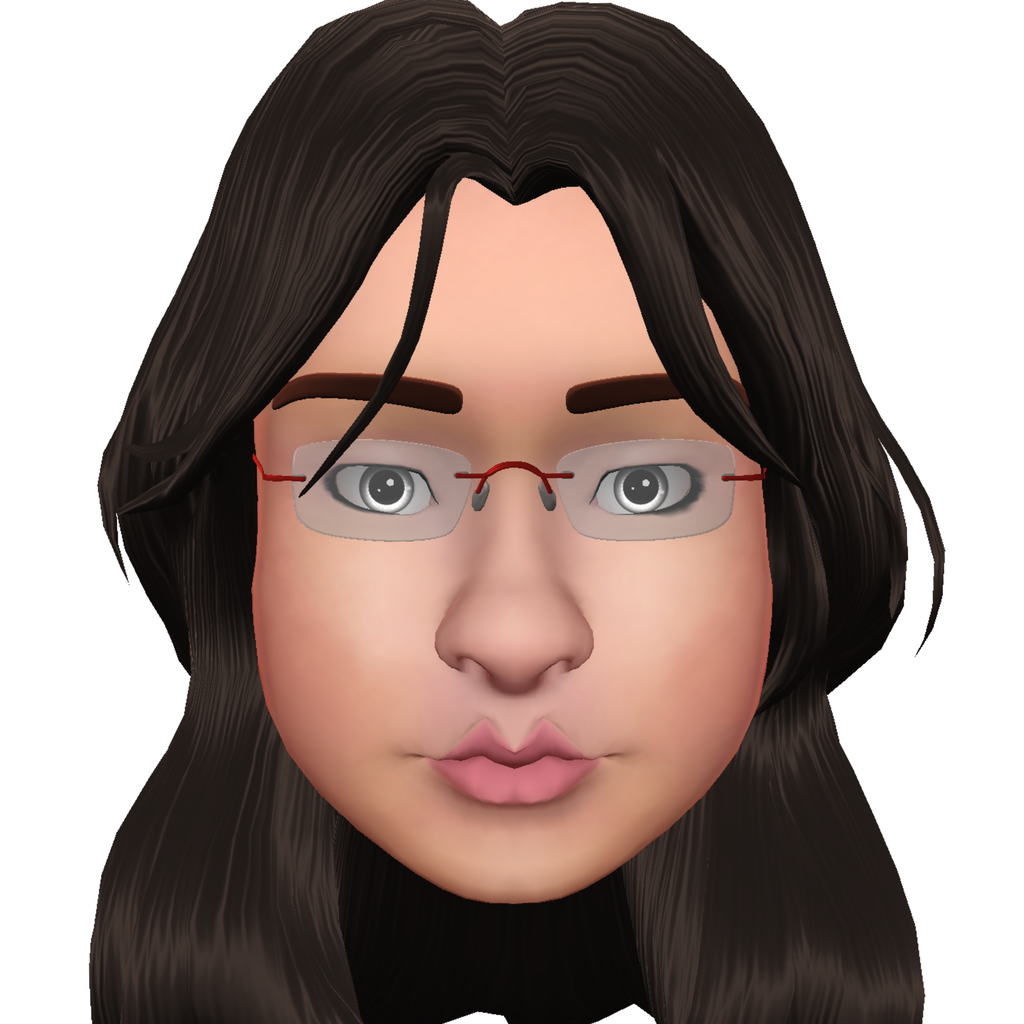}}
\subfloat[]{
\includegraphics[width=0.245\linewidth]{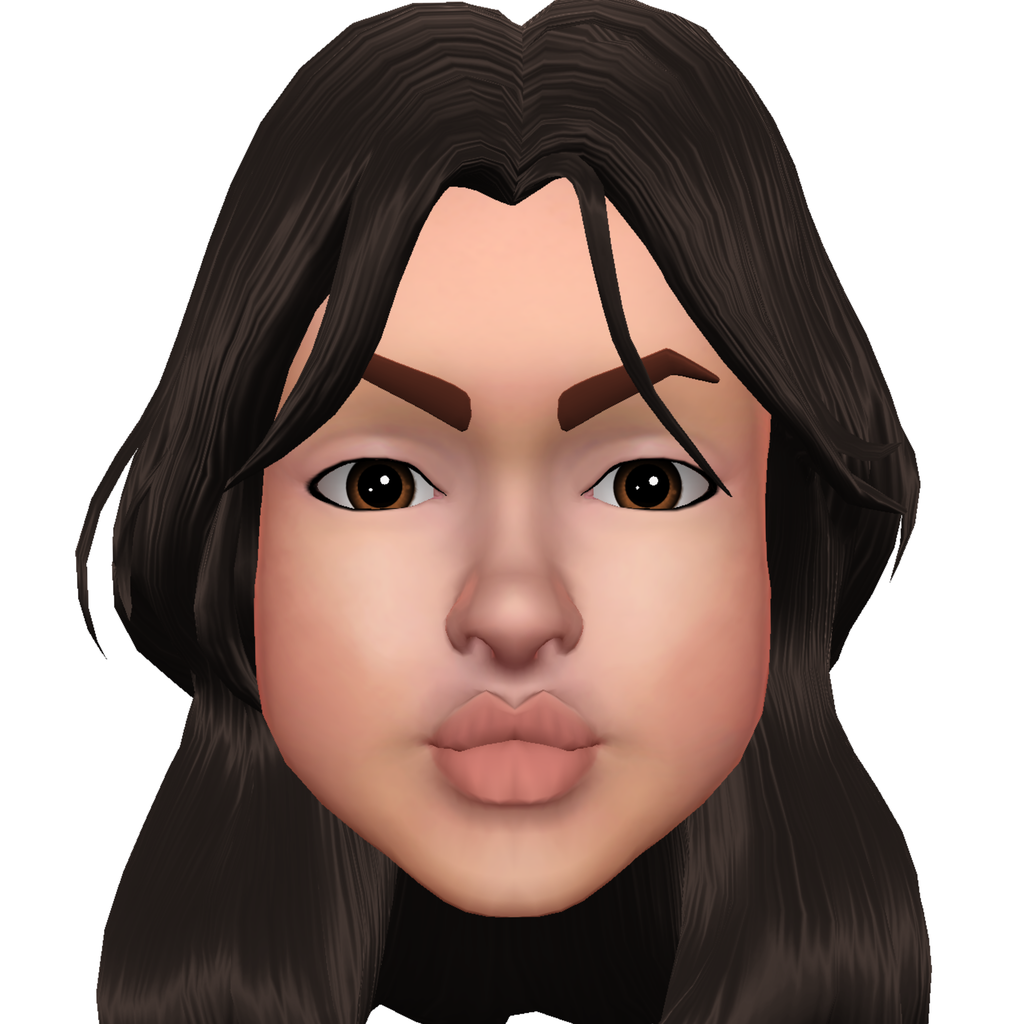}}
\subfloat[]{
\includegraphics[width=0.245\linewidth]{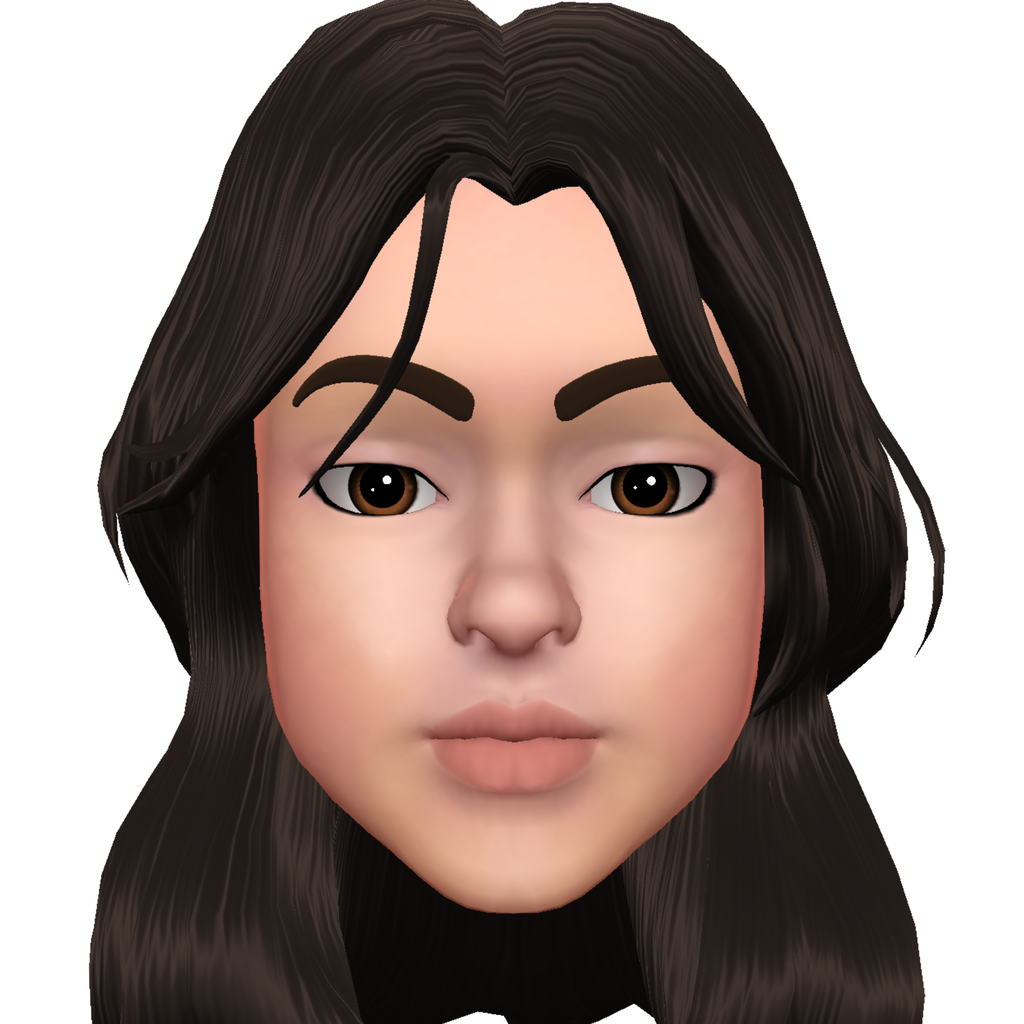}}
\vspace{-22pt}

\subfloat[Input]{
\includegraphics[width=0.245\linewidth]{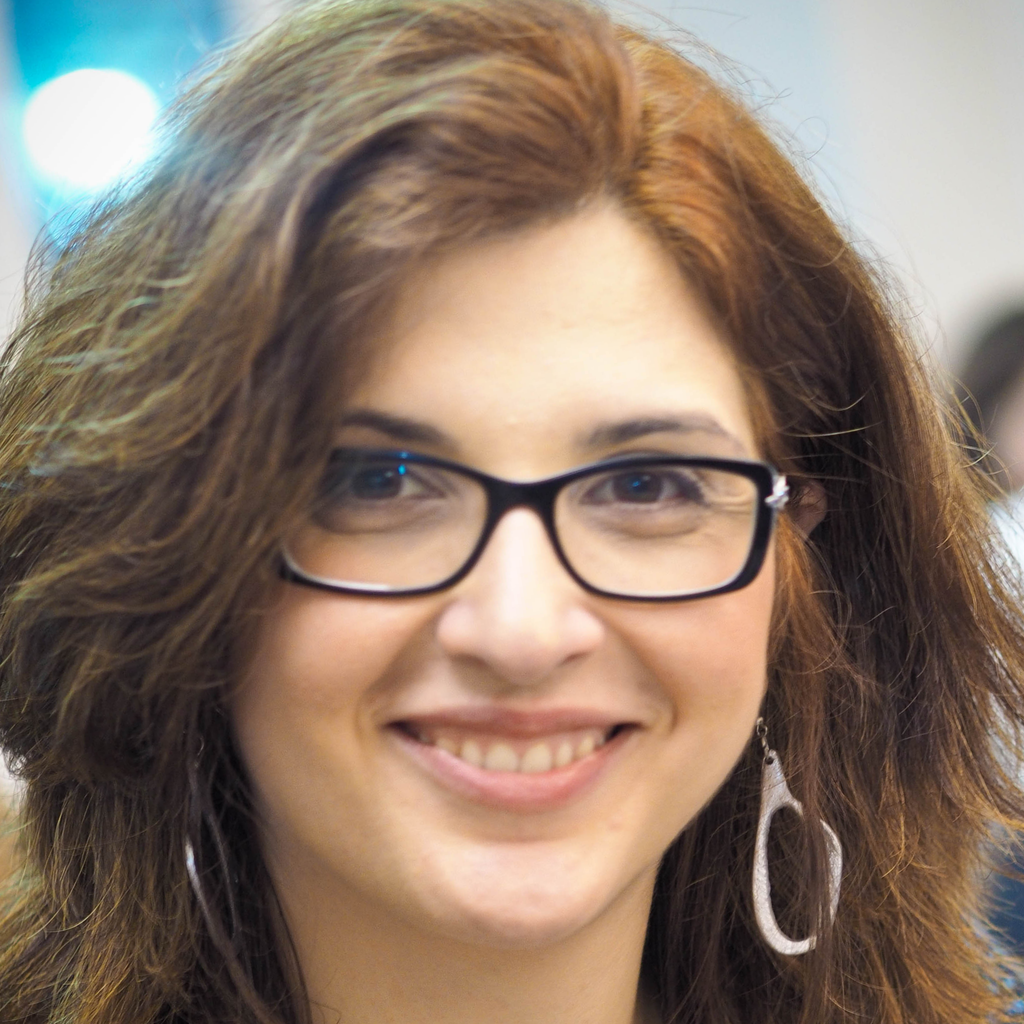}}
\subfloat[(a) No stylization]{
\includegraphics[width=0.245\linewidth]{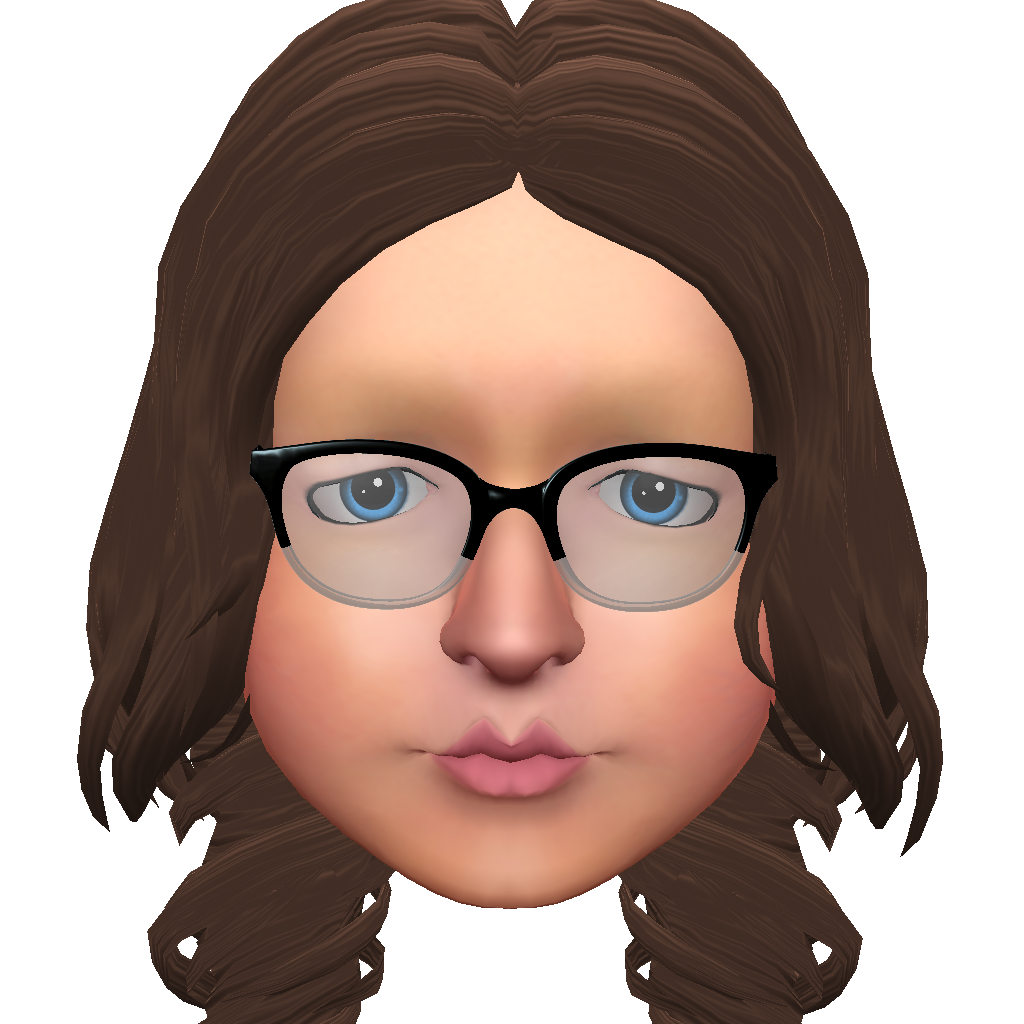}}
\subfloat[(b) AgileGAN]{
\includegraphics[width=0.245\linewidth]{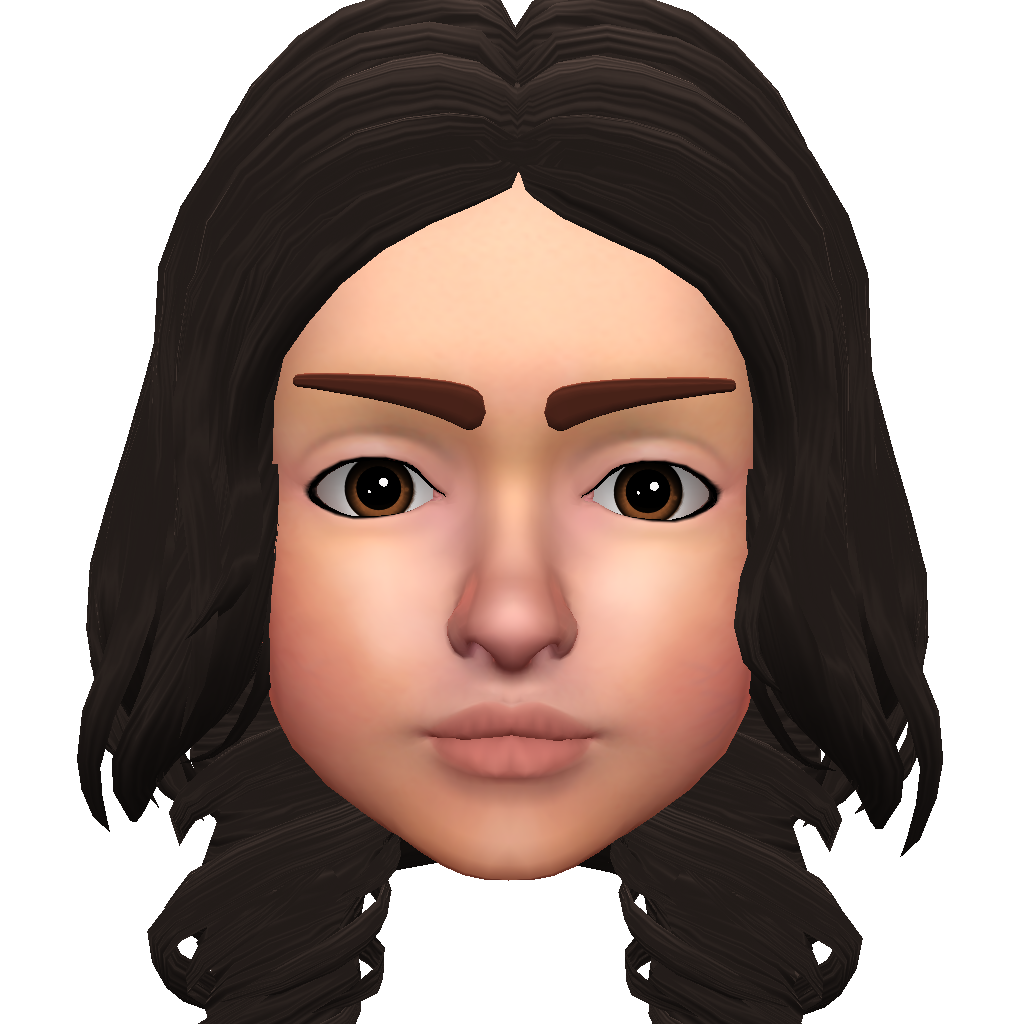}}
\subfloat[(c) Ours]{
\includegraphics[width=0.245\linewidth]{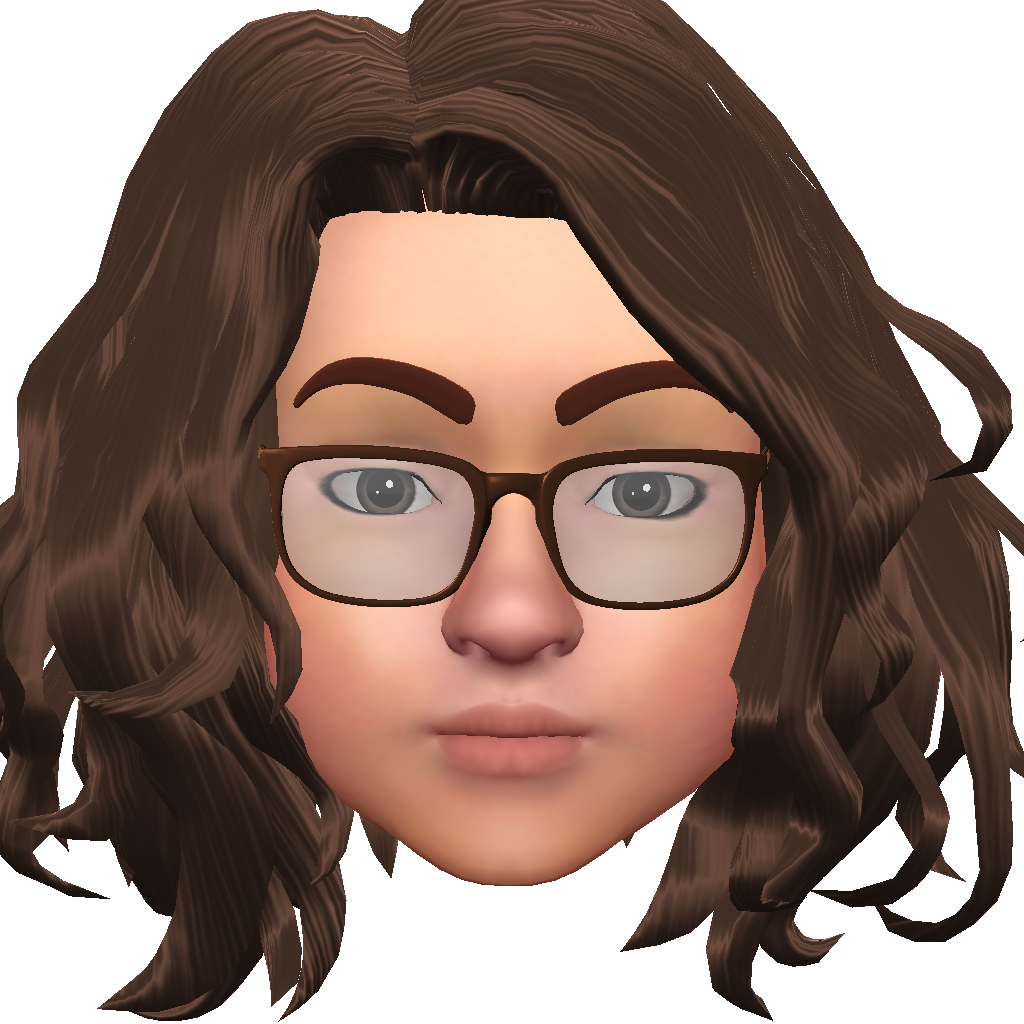}}
   \caption{We ablate by removing the stylization stage, as well as replacing our stylization with a state-of-the-art method. In each case the final renderings from the graphics engine are shown. (a) Fitting directly on a user image results in an avatar that lacks attractiveness. (b) Replacing our stylization with AgileGAN~\shortcite{song2021agilegan} suffers from missing personal information such as glasses and artifacts where smiles are misinterpreted as heavy lips or mustache. (c) Our stylization retains personal features like glasses, and generate visually appealing results in spite of expressions.  \textcopyright Chang-Ching Su and Luca Boldrini.}
  	\label{fig:stylization_ablation}
  	\vspace{6pt}
\end{figure}

\paragraph{Portrait Stylization Ablation:}
To study the impact of Portrait Stylization on the complete avatar creation pipeline we compare three options, shown in Fig.~\ref{fig:stylization_ablation}.  \textit{No stylization} removes this stage entirely and uses the real image as input to parameterization loss calculation. Without stylization, the parameterization module tries to match the real image with the target stylized avatar, leading to poor visual quality. \textit{AgileGAN}~\cite{song2021agilegan} is a state-of-the-art stylization method. It provides stylization and thus improves the final avatar attractiveness compared to no stylization. However, it cannot remove the impact of expressions and does not handle glasses well. In Row 1 (b), the smile expression is explained as a big mouth in the fitting stage, and personal information like glasses is not preserved in Row 2 (b). Our method addresses these issues and achieves better results in both visual quality and personal identity.

\paragraph{Mapper Losses Ablation:}
To study the importance of including all losses while training the Mapper, we generate results using different permutations of loss terms (identity, LPIPS, color). We then collected 990 user judgements from Amazon Mechanical Turk, to select the best matching results to the input image among six permutation results.  Table~\ref{tab:loss} shows the fraction of each option selected.  The full set of losses achieves the best score by a small margin, matching our observations that the overall method is robust to the precise selection of loss, but that the additional terms help in some cases.




\begin{figure}
\vspace{-18pt}
\subfloat[(a) Limitation - Small areas]{
\includegraphics[width=0.24\linewidth]{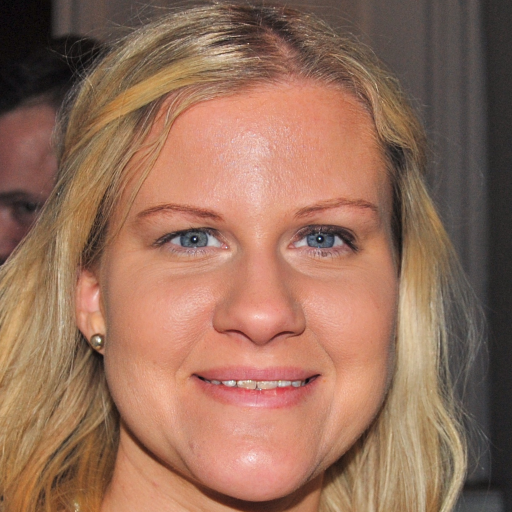}
\includegraphics[width=0.24\linewidth]{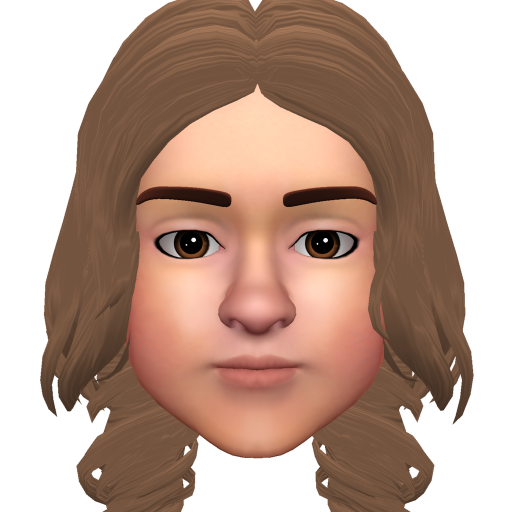}}
\subfloat[(b) Limitation - Shadows]{
\includegraphics[width=0.24\linewidth]{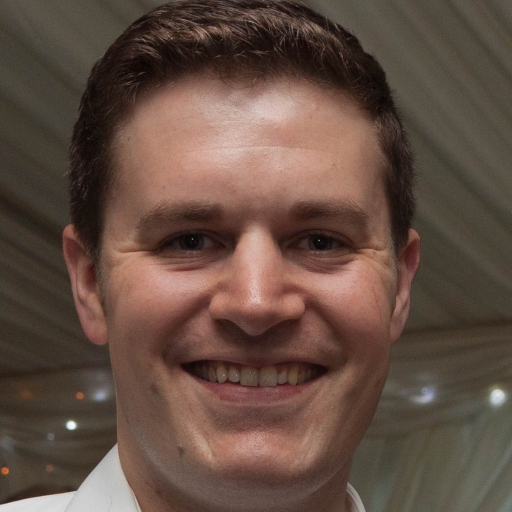}
\includegraphics[width=0.24\linewidth]{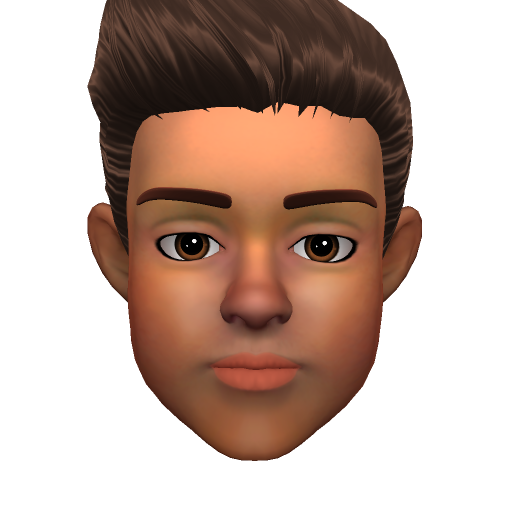}}
  \caption{Limitations: (a) failure on a parameter (eye color) affecting a small number of pixels. (b) incorrect skin tone prediction caused by shadows.  \textcopyright Daniel Åberg and Peter Bright.}
  \label{fig:fail}
\vspace{10pt}
\end{figure}


\section{Limitations}
We observe two main limitations to our method. First, our method occasionally produces wrong predictions on assets covering a small area, because their contribution to the loss is small and gets ignored. The eye color in Fig.~\ref{fig:fail} (a) is an example of this difficulty. Redesigning the loss function might resolve this problem. Second, lighting is not fully normalized in the stylization stage, leading to incorrect skin tone estimates when there are strong shadows, shown in Fig.~\ref{fig:fail} (b). This problem could potentially be addressed by incorporating intrinsic decomposition into the pipeline. 
In addition to the limitations of our method, we experience a loss of ethnicity in the final results, which is mainly introduced by the graphics engine, as also evidenced by the manually-created results. This issue could be addressed by improving the diversity of the avatar system.

\section{Conclusion}
In summary, we present a self-supervised stylized avatar auto-creation method with cascaded domain crossing. Our method demonstrates that the gap between the real images domain and the target avatar domain can be progressively bridged with a three-stage pipeline: portrait stylization, self-supervised avatar parameterization, and avatar vector conversion. Each stage is carefully designed and cannot be simply removed. Experimental results show that our approach produces high quality attractive 3D avatars with personal identities preserved.
In the future, we will extend the proposed pipeline to other domains, such as cubism and caricature avatars.


%
%
%
%

\bibliographystyle{ACM-Reference-Format}
\bibliography{mybib}


\appendix

\section{Portrait Stylization Details}
\label{appendix:agilegan}
\paragraph{Segmentation Models:} The avatar segmentation model is trained using 20k randomly sampled avatar vectors with neural pose, expression and illumination. For real image segmentation, we used an open-source pre-trained BiSeNet module\footnote{https://github.com/zllrunning/face-parsing.PyTorch}~\cite{yu2018bisenet}.

\paragraph{Distribution Prior $\mathcal{W}$:} To sample W+ distribution prior, we inverse CelebA dataset ~\cite{liu2015deep} into W+ space using a pre-trained e4e encoder~\cite{tov2021designing}.

\paragraph{Normalized Style Exemplar Set $\mathcal{Y}$:} For training stylized generator $\mathcal{G}_{\phi_t}$, we synthetically rendered a diverse set of 150 avatar imageries with normalized facial expressions.

\section{Avatar Parameterization Details}
\subsection{Imitator}
\label{appendix:imitator}
To train our module in a self-supervised way, we plug-in a differentiable neural renderer (i.e. imitator) in our learning framework. As we mentioned in the main paper, the imitator can take a relaxed avatar vector as input, although the imitator itself is trained with strict avatar vector. No matter the input is a relaxed or strict avatar vector, it can produce a valid rendering. In this way, we can supervise the training in image space without any ground-truth for the parameters. Due to the differentiability of the imitator, the parameterization stage can be trained with gradient descent. To achieve high fidelity rendering quality, we leverage the StyleGAN2 generator \cite{karras2019style} as our backbone, which is capable of generating high quality renderings matching the graphics engine. The imitator consists of an encoder $\mathcal{E}_{i}$ implemented using MLP and a generator $\mathcal{G}_{i}$ adopted from StyleGAN2. The encoder translates an input avatar vector to a latent code $w+$. The generator then produces a high-quality image given the latent code.

\paragraph{Training:} 
In order to fully utilize the image generation capability of StyleGAN2, we propose to train the imitator in two steps: 1) we first train a StyleGAN2 from scratch with random rendering samples generated by our graphics engine to obtain a high-quality image generator, without any label or conditions; then 2) we train the encoder and the generator together with images and corresponding labels, result in a conditional generator. Given an avatar vector $v$, a target image $\mathcal{I}_{gt}$ and the generated image $\mathcal{I}_{gen} = \mathcal{G}_{i}(\mathcal{E}_{i}(v)) $, we use the following loss function combination to perform the second step training:

\begin{equation}\label{eq:loss_imitator}
\begin{split}
\mathcal{L}_{imitator} = \lambda_{1} \| \mathcal{I}_{gen} - \mathcal{I}_{gt} \|_{1}  + \lambda_{2} \mathcal{L}_{lpips} 
 + \lambda_{3} \mathcal{L}_{id}
\end{split}
\end{equation}
where the first term is an L1 loss, which encourages less blurring than L2. In addition, $\mathcal{L}_{lpips}$ is the LPIPS loss adopted from \cite{zhang2018perceptual},

\begin{equation}\label{eq:loss_lpips}
\mathcal{L}_{lpips} = \| \mathcal{F}(I_{1}) - \mathcal{F}(I_{2}) \|_{2}
\end{equation}

\noindent where $\mathcal{F}$ denotes the perceptual feature extractor. $\mathcal{L}_{id}$ is the identity loss which measures the cosine similarity between two faces built upon a pretrained ArcFace \cite{Deng_2019_CVPR} face recognition network $\mathcal{R}$, 

\begin{equation}\label{eq:loss_identity}
\mathcal{L}_{id} = 1 - cos(\mathcal{R}(I_{1}), \mathcal{R}(I_{2}))
\end{equation}

We set $\lambda_1=1.0$, $\lambda_2=0.8$, $\lambda_3=1.0$, empirically.

\begin{figure}
\subfloat[]{
\includegraphics[width=0.24\linewidth]{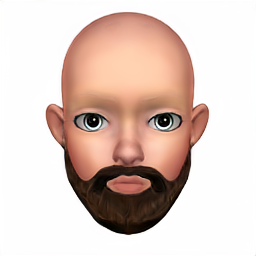}}
\subfloat[]{
\includegraphics[width=0.24\linewidth]{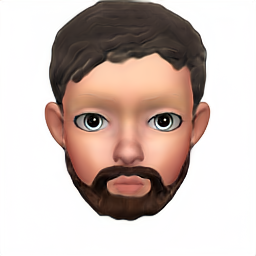}}
\subfloat[]{
\includegraphics[width=0.24\linewidth]{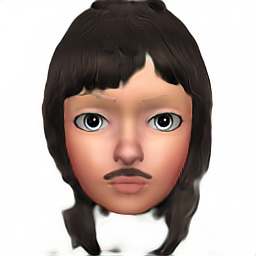}}
\subfloat[]{
\includegraphics[width=0.24\linewidth]{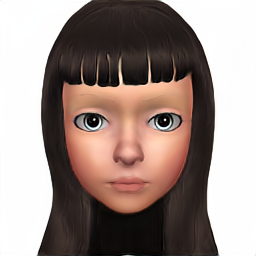}}
\vspace{-10pt}
  \caption{Interpolation of avatar vectors. The neural rendering imitator which temporarily replaces the traditional graphics engine is differentiable, allowing the relaxation of the strict constraint on discrete types. Linear interpolation between two  avatar vectors results in the gradual disappearance of the beard and the gradual growth of the hair.
  }
  	\label{fig:imitator_interpolation}
\end{figure}

\paragraph{Interpolation property:} 
Fig.~\ref{fig:imitator_interpolation} provides an example of the interpolation property of the imitator which enables relaxed optimization over the discrete parameters.

\paragraph{Implementation:} 
To train the imitator, we randomly generate 100,000 images and corresponding parameters. Note that although random sampling leads to strange avatars, our imitator can generate images matching the graphics engine well by seeing plenty of samples in the parameter space. Please refer to our supplementary video for a side-by-side comparison.

We train StyleGAN2 using the official source code\footnote{https://github.com/NVlabs/stylegan2-ada-pytorch} with images of size $256 \times 256 \times 3$, thus the latent code $w+$ has a shape of $14 \times 512$. We build the encoder $\mathcal{E}_{i}$ with 14 individual small MLPs, each is responsible for mapping from the input vector to one latent style. Given the pretrained generator, we train the encoder and simultaneously finetune the generator with Adam~\cite{kingma2014adam}. We set the initial learning as 0.01 and decay it by 0.5 each two epochs. In our experiments, it takes around 20 epochs to converge.

\subsection{Mapper}
We use CelebA-HQ \cite{CelebAMask-HQ} and FFHQ \cite{karras2019style} as our training data. To collect a high quality dataset for training, we use the Azure Face API \footnote{https://azure.microsoft.com/en-us/services/cognitive-services/face} to analyze the facial attributes and keep only facial images that meet our requirements:

1) within a limited pose range ($yaw<8^\circ, pitch<8^\circ, roll<5^\circ$)

2) without headwears

3) without extreme expressions

4) without any occlusions

\noindent Finally, we collect 21,522 images in total for mapper training.

The input is an $18 \times 512$ latent code taken from the Stylization module. Each one of the 18 layers latent code is passed to an individual MLP. The output features are then concatenated together. After that, we apply two MLP heads to generate continuous and discrete parameters separately. 

We apply a scaling before the softmax function for discrete parameters:
\begin{equation}\label{eq:softmax}
\mathcal{S}(x) = \frac{e^{\beta x_{k}}}{\Sigma_{i=1}^{N} e^{\beta x_{i}}}, k = 1, ... N
\end{equation}
where $\beta > 1$ is a coefficient that performs non-maximum suppression over some types that contribute less than the dominant ones, and $N$ is the number of discrete types. 
During training, we gradually increase the coefficient $\beta$ to perform an easy-to-hard training by decreasing the smoothness. Empirically, we increase $\beta$ by 1 for each epoch. We train the mapper for 20 epochs.

\end{document}